\documentclass[journal]{IEEEtai}

\usepackage{bm}
\usepackage{amsmath}
\usepackage{epsf}
\usepackage{graphics}
\usepackage{ amssymb }
\usepackage[dvips]{graphicx}
\usepackage{mathtools}% Loads amsmath
\usepackage{epsfig}
\usepackage{cite}
\usepackage[linesnumbered,ruled,vlined]{algorithm2e}
\usepackage{colortbl}
\usepackage{color}
\usepackage{enumitem}
\usepackage{makecell}
\usepackage{tcolorbox}

\usepackage{bm}
\usepackage{amsmath}
\usepackage{epsf}
\usepackage{graphics}
\usepackage{ amssymb }
\usepackage[dvips]{graphicx}
\usepackage{epsfig}
\usepackage{cite}
\usepackage[linesnumbered,ruled,vlined]{algorithm2e}
\usepackage{graphicx}
\usepackage{epsfig}
\usepackage{latexsym}
\usepackage{amsfonts}
\usepackage{here}
\usepackage{rawfonts}

\usepackage[utf8]{inputenc}
\usepackage[english]{babel}
\usepackage{amsmath}
\usepackage{amsfonts}
\usepackage{amssymb}
\usepackage{color} %May be necessary if you want to color links
\usepackage{bm}
\usepackage{listings}
\usepackage{caption}
    \usepackage{multicol}

\usepackage{tabularx,booktabs}
\newcolumntype{Y}{>{\centering\arraybackslash}X}
\makeatletter
\newcommand\notsotiny{\@setfontsize\notsotiny{6.31415}{7.1828}}
\makeatother
\usepackage{amssymb}
\usepackage{amsthm}
\usepackage{graphicx}
\usepackage{epstopdf}
\usepackage{listings}
\usepackage{float}
\usepackage{amsmath}
\usepackage{amssymb}
\usepackage{amsfonts}
\usepackage{epstopdf}

\usepackage{multirow}
\usepackage{amscd}
\usepackage{mathrsfs}
\usepackage{graphicx}
\usepackage{color}
\usepackage{url}
\usepackage{bm}
\usepackage{ulem} %for editing purposes only
\newcolumntype{?}{!{\vrule width 1pt}}
\usepackage{setspace}
\usepackage{footnote}
\DeclareMathOperator*{\argmin}{arg\,min}

\newtheorem{remark}{Remark}

\usepackage[]{algorithm2e}

\renewcommand{\figurename}{Fig.}
\addto\captionsenglish{\renewcommand{\figurename}{Fig.}}
\usepackage{bbm}

\usepackage{multicol}
\usepackage{mathtools}

\usepackage{lipsum,graphicx}
\usepackage{subcaption}

\usepackage{diagbox}
\usepackage{hyperref}
\usepackage[protrusion=true,expansion=true]{microtype}
\pdfoutput=1
\usepackage[font=footnotesize]{caption}
\allowdisplaybreaks[4]

\usepackage{graphicx}
\usepackage{grffile}
\usepackage{tabularx}
\newcounter{term}[section]

\renewcommand\theterm{\alph{term}}
\makeatletter
\newcommand{\vast}{\bBigg@{4}}

\newcommand{\Vast}{\bBigg@{5}}
\makeatother
\newcommand\semiHuge{\fontsize{22.7}{31.38}\selectfont}
\usepackage{soul}

\usepackage{tabularray}

\definecolor{Gray}{gray}{0.9}

\definecolor{BoxFrame}{HTML}{0F7173}  % Carribean Current
\definecolor{BoxBg}{HTML}{C3E5F5}   % Pale Azure
\definecolor{FedProxBg}{HTML}{C3E5F5} % Emerald
\definecolor{FedProxFrame}{HTML}{0F7173} % Cal Poly Green
\definecolor{SCAFFOLDBg}{HTML}{C3E5F5} % Emerald
\definecolor{SCAFFOLDFrame}{HTML}{0F7173} % Cal Poly Green
\definecolor{DittoBg}{HTML}{C3E5F5} % Emerald
\definecolor{DittoFrame}{HTML}{0F7173} % Cal Poly Green
\definecolor{FedRepBg}{HTML}{C3E5F5} % Emerald
\definecolor{FedRepFrame}{HTML}{0F7173} % Cal Poly Green
\definecolor{FedAmpBg}{HTML}{C3E5F5} % Emerald
\definecolor{FedAmpFrame}{HTML}{0F7173} % Cal Poly Green

\definecolor{LightBlue}{rgb}{0.8,0.89,1}
\definecolor{DarkBlue}{rgb}{0.47,0.784,1}
\definecolor{DarkerBlue}{HTML}{015792}
\definecolor{MagLight}{rgb}{1, 0.89, 0.8}
\newtagform{blue}{\color{blue}(}{)}
\usepackage{hhline}
\newcolumntype{?}{!{\vrule width 1.5pt}}
\newlength{\rowhgtspace} 

\begin{document}

\title{Redefining non-IID Data in Federated Learning for Computer Vision Tasks: Migrating from Labels to Embeddings for Task-Specific Data Distributions
\vspace{-1mm}}
 % Task-Specific %Federated %Data Heterogeneity %Novel %Gap-Inducing
 
 % Task-Specific Data Heterogeneity in Federated Learning for Computer Vision
 % Task-Specific Data Heterogeneity in Federated Learning for Computer Vision
 % Task-Specific Data Heterogeneity in Federated Learning: Towards Embedding-based Heterogeneous Data
 % 
\author{Kasra Borazjani,~\IEEEmembership{Student Member,~IEEE,} Payam Abdisarabshali,~\IEEEmembership{Student Member,~IEEE,}\\ Naji Khosravan,~\IEEEmembership{Member,~IEEE,} and 
Seyyedali Hosseinalipour,~\IEEEmembership{Senior Member,~IEEE}
        % <-this % stops a space
\thanks{K. Borazjani, P. Abdisarabshali, and S. Hosseinalipour are with the Department of Electrical Engineering, University at Buffalo--SUNY, Buffalo, NY, 14260 USA (emails:\{kasrabor,payamabd,alipour\}@buffalo.edu). N. Khosravan is with 
Adobe Firefly, Seattle, WA, 98103 USA (email: naji.khosravan@gmail.com).\\
This work was in part supported by the U.S. National Science Foundation (NSF) under Grant No. SaTC-2513164 and  ECCS-2512911. \\
Corresponding Author: Seyyedali Hosseinalipour}
\vspace{-1mm}}% <-this % stops a space

% \markboth{Journal of Transactions in Neural Networks and Learning Systems,~Vol.~14, No.~8, August~2021}%


\maketitle

\begin{abstract}
Federated Learning (FL) has emerged as one of the prominent paradigms for distributed machine learning (ML). However, it is well-established that its performance can degrade significantly under non-IID (non-independent and identically distributed) data distributions across clients. To study this effect, the existing works predominantly emulate data heterogeneity by imposing label distribution skew across clients. In this paper, we show that label distribution skew fails to fully capture the data heterogeneity in computer vision tasks beyond classification, \textit{exposing an overlooked gap in the literature}. Motivated by this, by utilizing pre-trained deep neural networks to extract task-specific data embeddings, we define \textit{task-specific data heterogeneity} through the lens of each vision task and introduce a new level of data heterogeneity called \textit{embedding-based data heterogeneity}. Our methodology involves clustering data points based on embeddings and distributing them among clients using the Dirichlet distribution. Through extensive experiments, we evaluate the performance of different FL methods under our revamped notion of data heterogeneity, introducing new benchmark performance measures to the literature. For instance, across seven representative computer vision tasks, our embedding-based heterogeneity formulation leads to up to around 60\% increase in the observed loss under FedAvg, indicating that it more accurately exposes the performance degradation caused by data heterogeneity. We further unveil a series of open research directions that can be pursued. (Code: \href{https://github.com/KasraBorazjani/task-perspective-het.git}{https://github.com/KasraBorazjani/task-perspective-het.git})
\end{abstract}

\begin{IEEEImpStatement}
Federated Learning (FL) is a cornerstone of privacy-preserving distributed machine learning (ML), yet its evaluation has been fundamentally shaped (and, as we show, misled) by the prevalent reliance on label-based data heterogeneity. Existing benchmarks overwhelmingly emulate non-IID (non-independent and identically distributed) data settings by skewing label distributions, an approach suitable for classification but inadequate for vision tasks beyond classification (e.g., depth estimation, edge detection, segmentation). 
Our work makes a conceptual and methodological leap by redefining data heterogeneity through task-specific embeddings. We introduce embedding-based data heterogeneity, extracted from the penultimate layer of task-trained deep neural networks, enabling the emulation of data heterogeneity from each vision task’s perspective.
Through extensive experiments across multiple FL algorithms and various representative computer vision tasks, we show that our embedding-based data heterogeneity produces significantly different performance trends, which reveals performance degradations invisible to label-based data heterogeneity models. 
As a result, this work establishes one of the first benchmarks for task-specific data heterogeneity, offers a new lens for studying task similarity, and opens fertile research directions in single- and multi-task FL.
\end{IEEEImpStatement}


\vspace{-3mm}
\section{Introduction}
\noindent 
{\semiHuge \textbf{F}}EDERATED learning (FL) has emerged as a pioneering paradigm for privacy-preserving distributed learning and has been applied across diverse domains such as vehicular networks~\cite{hakeem2025advancing}, medical imaging~\cite{borazjani2024multi}, and sensing networks~\cite{kapoor2024federated}. Compared to centralized training, a defining challenge in FL is that model performance is highly sensitive to data heterogeneity across participating devices/clients, often resulting in biased local model updates and degraded global convergence~\cite{kairouz2021advances}. Consequently, understanding and accurately characterizing the impact of data heterogeneity has become a central research theme in FL~\cite{pei2024review}.
In this paper, we shed light on an underexplored dimension of data heterogeneity in FL by moving beyond the conventional \textit{class-based formulations} (which rely on label distributions in classification tasks) to computer vision tasks such as ``Reshading," where labels are inherently absent. This shift raises two fundamental questions: How does data heterogeneity impact the performance of FL in such vision tasks? And are current methodologies adequately capturing this impact? Our short answer is \textit{No}. Specifically, we reveal a notable oversight in the literature -- namely, that the performance of existing FL strategies in computer vision has been overestimated in the presence of data heterogeneity among the participating clients.

\vspace{-3mm}
\subsection{Overview and Literature Review}
FL has emerged as a promising paradigm for distributed machine learning (ML), which enables multiple clients to collaboratively train ML models without sharing raw data, thus preserving user privacy\cite{kairouz2021advances,fed-avg,fedprox}. This approach offers an alternative to traditional centralized ML techniques, where data must be transferred to a central server, often resulting in significant communication overhead and privacy concerns. The training process in FL takes place over several global rounds. In each round, clients first train ML models locally on their datasets, typically using several iterations of stochastic gradient descent (SGD). They then send the updated local models to a central server, which aggregates the models (usually through averaging) to form a new global model. This global model is subsequently broadcast back to the clients to start the next round of local training.
Implementing FL in practice faces unique challenges, largely due to the \textit{inherent non-IID (non-independent and identically distributed) or heterogeneous nature of data across clients}~\cite{fed-avg,fedprox,pieri2023handling,mendieta2022local,NEURIPS2021_2f2b2656,darzi2024tackling,zhang2022splitavg}. This 
data heterogeneity has a significant impact on FL performance.  This challenge has motivated a range of FL methods designed to mitigate the adverse effects of data heterogeneity~\cite{fedprox,scaffold,fedamp,fedrep}.

% Several approaches have been proposed to partition data and distribute them among clients, each imposing different types of data heterogeneity~\cite{kairouz2021advances,li2022federated}. 
% The most prominent ones are \textit{feature distribution skew} and \textit{label distribution skew}. Feature distribution skew refers to scenarios where clients' data exhibit varying feature distributions. Although the labels may be similar across clients, the underlying feature characteristics might differ significantly. For instance, in a speech recognition task, clients from different regions may have varying accents or background noise levels. Conversely, the label distribution skew approach, primarily found in computer vision tasks, involves partitioning data in such a way that clients receive data with imbalanced label distributions~\cite{fed-avg, fedprox,qu2022rethinking, fl-dirichlet, darzi2024tackling, he2021fedcv,li2022federated,wang2020tackling, mendieta2022local}. For example, in an image classification task involving street scenes, some clients may have an overrepresentation of images featuring urban environments with skyscrapers, while others may only have images of suburban neighborhoods or rural areas with houses and trees.
 Effective evaluation of the robustness of FL algorithms thus necessitates examining how data heterogeneity influences FL performance. 
 To this end, state-of-the-art methods in FL that concern modeling the data heterogeneity have mainly emulated data heterogeneity across clients through imposing \textit{label distribution skew\footnote{The label distribution skew approach involves partitioning data in such a way that clients receive data with imbalanced label distributions~\cite{kairouz2021advances,li2022federated}.}}, largely applied to computer vision \textit{classification tasks}~\cite{fed-avg, fedprox,qu2022rethinking, fl-dirichlet, darzi2024tackling, he2021fedcv,li2022federated,wang2020tackling, mendieta2022local}. 
In particular, when FL first emerged as a decentralized ML paradigm, researchers
emulated data heterogeneity through  distributing data points based on their labels across clients to impose label distribution skew. This approach became a widely used method to simulate data heterogeneity in subsequent FL research~\cite{fed-avg,qu2022rethinking, darzi2024tackling, he2021fedcv}. For example, in a handwritten digits classification task (MNIST dataset~\cite{mnist}), one client might have a local dataset mostly consisting of the digits ``0" and ``1", while another client might only have images of the digits ``7" and ``8"~\cite{fed-avg}. As the number of labels per clients decreases, each client's dataset becomes less representative of the global dataset, emulating a more non-IID data distribution. While this fundamental approach introduces some degree of data heterogeneity, there was a pressing need for mathematical and unified approaches to introduce and measure the level of data heterogeneity in FL. A major step in this direction was taken by utilizing the \textit{Dirichlet distribution} to impose label distribution skew among clients in FL~\cite{fedprox,he2021fedcv, fl-dirichlet,li2022federated,wang2020tackling, mendieta2022local, darzi2024tackling}, enabling gradual heterogeneity in label/class distribution across the clients by fine-tuning Dirichlet concentration parameter.
% This methodology inherently induces label-based heterogeneity.

Nevertheless, the above-mentioned class-based (equivalently, label-based) data heterogeneity may not be a perfect match for other types of tasks, specifically computer vision tasks beyond classification, such as surface normal estimation, depth estimation, edge detection, and object detection~\cite{taskonomy,he2021fedcv} (see Fig.~\ref{fig:taskonomy} for a list of these tasks). Intuitively, the ineffectiveness of class-based data heterogeneity for generic computer vision tasks stems from the fact that non-classification vision tasks do not necessarily \textit{perceive} data distribution in the same way as classification tasks. In fact, \textit{their perception of data heterogeneity can be label-independent and majorly different from each other: an aspect largely overlooked in the FL literature.} For instance, the “2D Edges” task may be more sensitive to the configuration of edges within an image rather than its label. In this context, two datapoints with the same label (e.g., both with “Living Room” label) may exhibit highly different edge structures, rendering them heterogeneous from the perspective of the “2D Edges” task. Conversely, two datapoints with different labels (e.g., “Living Room” and “Kitchen”) might have similar edge structures, making them homogeneous for this task. This highlights the need for methodologies in FL that emulate data heterogeneity based on the unique perspectives of vision tasks to data.

Building on the above intuition, in this paper, we propose the following fundamental hypothesis:
\begin{quote}
\textit{As we demonstrate in this paper, the performance of FL for generic computer vision tasks has been \textit{overestimated} in prior studies~\cite{zhuang2023mas, lu2024fedhca2, mortaheb2022personalized,cai2023many}. These studies largely depend on class-based heterogeneity, which fails to accurately reflect the data distribution in non-classification computer vision tasks. Consequently, the actual performance and robustness of FL, and its variants, in handling data heterogeneity for generic computer vision tasks remain largely unexplored, highlighting a significant gap in the current literature.}
\end{quote}
 
To bridge this gap, we redefine the notion of data heterogeneity in FL for computer vision tasks as follows:

\begin{quote}
    \textit{Instead of inducing heterogeneity solely from a raw data perspective, we propose a migration towards assessing data heterogeneity through the unique perspective of each task to the data after applying task-specific transformations on data.}
\end{quote}

To achieve the above objective, we introduce a new layer of data heterogeneity, which we call embedding-based data heterogeneity. To model \textit{embedding-based data heterogeneity} (i.e., embedding distribution skew), we leverage a pre-trained deep neural network (DNN) for each task and feed forward data points through this network (see Step 1 in Fig.~\ref{fig:nn-model}). We then exploit the fact that the initial layers of the DNN of each task extract low-level features, and as the data progresses through deeper layers, the network extracts higher-level representations that are increasingly specific to the vision task of interest. Particularly, as we reach the layer just prior to the last layer (i.e, penultimate layer), \textit{embeddings} that encapsulate a comprehensive view of the data from the perspective of the task are obtained (see Step 2 in Fig.~\ref{fig:nn-model}), integrating both low-level and high-level features. We then leverage K-means to cluster datapoints based on these embeddings (see Step 3 in Fig.~\ref{fig:nn-model}). Subsequently, we utilize a Dirichlet distribution to assign data points of each cluster to different clients to emulate task-specific data heterogeneity (see Step 4 in Fig.~\ref{fig:nn-model}).

We note that, similar to the widely used class-based Dirichlet distribution approach, our embedding-based method is intended as a \textit{data heterogeneity emulation mechanism} for the simulation phase of FL.  In this setting, the method operates in a centralized manner on publicly available datasets prior to the start of FL training simulation to accommodate an offline generation of data partitions for clients in simulations. This preprocessing step offers a task-aware alternative to traditional class-based data splits and enables the study of how FL methods behave under the types of emulated data heterogeneity that can arise in non-classification computer vision tasks.

% To achieve the above objective, for the first time in the literature, we introduce a new layer of data heterogeneity, which we call \textit{embedding-based data heterogeneity}. To model embedding-based data heterogeneity (i.e., embedding distribution skew), we leverage a pre-trained deep neural network (DNN) for each task and feed forward data points through this network. We then exploit the fact that the initial layers of the DNN of each task extract low-level features, and as the data progresses through deeper layers, the network extracts higher-level representations that are increasingly specific to the vision task of interest. Particularly, as we reach the layer just prior to the last layer (i.e, penultimate layer), \textit{embeddings} that encapsulate a comprehensive view of the data from the perspective of the task are obtained, integrating both low-level and high-level features. We then leverage K-means to cluster datapoints based on these embeddings. Subsequently, we utilize a Dirichlet distribution to assign data points of each cluster to different clients to emulate task-specific data heterogeneity. 

% In this paper, we demonstrate that the performance of the current state-of-the-art methodologies, which rely on imposing label distribution skew to model data heterogeneity, drop significantly under embedding distribution skew. 
\begin{figure*}[!ht]
\vspace{-5mm}
    \centering
    \includegraphics[width=0.21\textwidth]{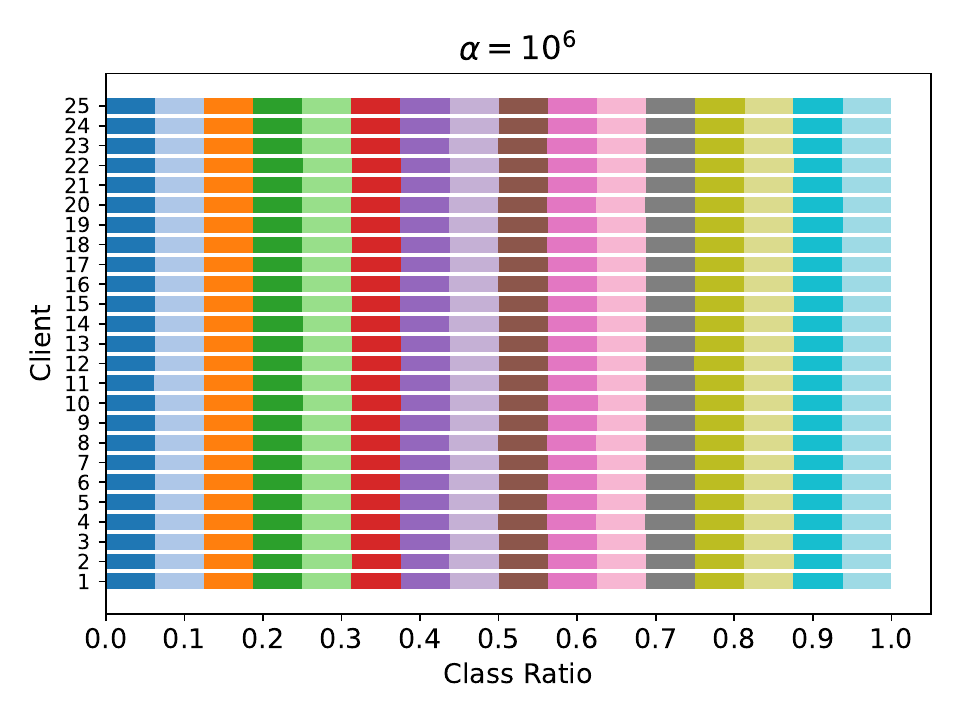}
    \includegraphics[width=0.21\textwidth]{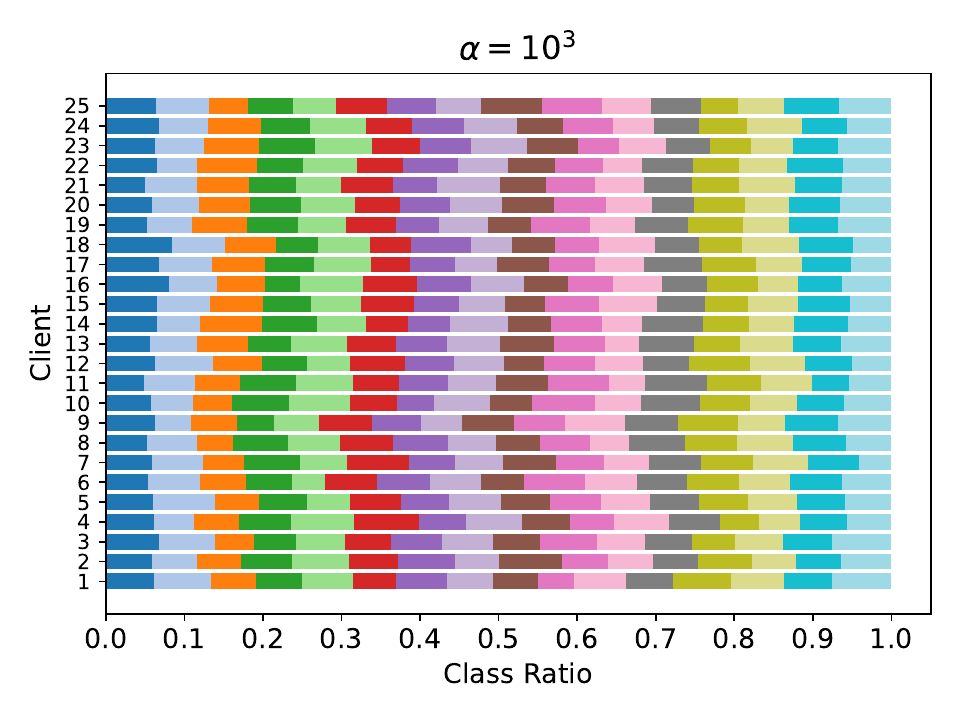}
    \includegraphics[width=0.21\textwidth]{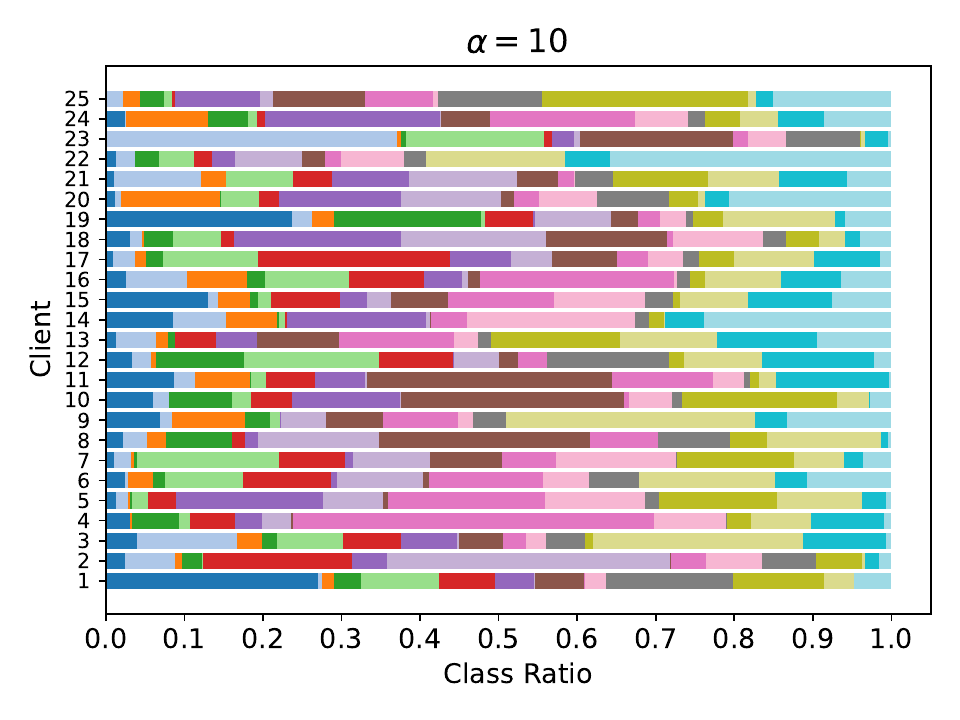}
    \includegraphics[width=0.21\textwidth]{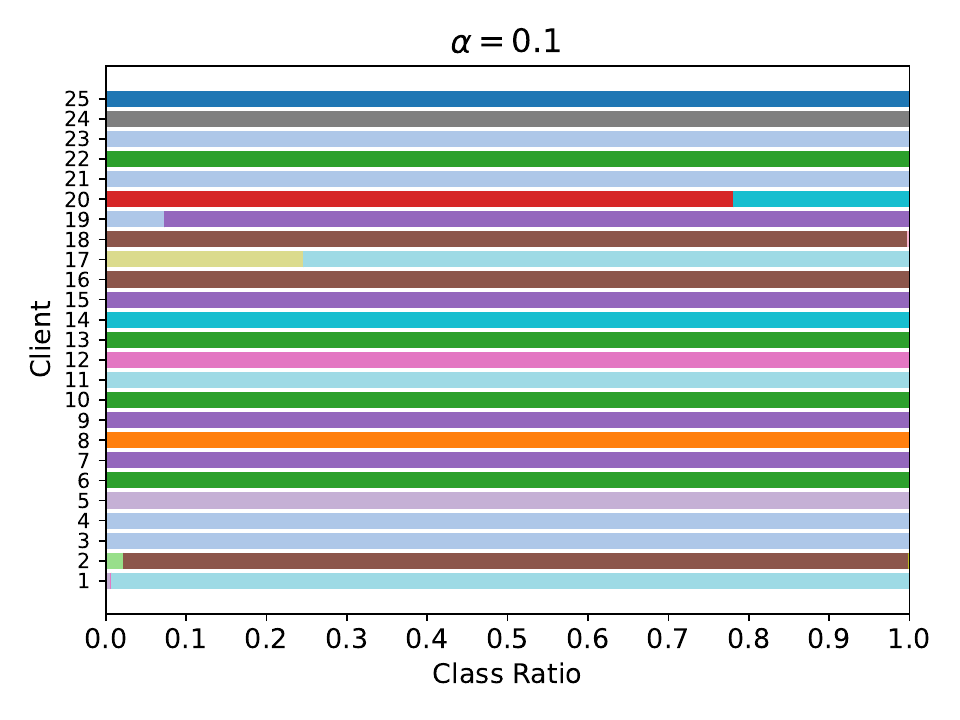}
    \caption{Dirichlet-generated data distribution for $N = 25$ clients and $16$ classes with homogeneous prior probabilities (i.e., $p_i = 1/|\mathcal{C}|,~ \forall 1 \leq i \leq |\mathcal{C}|$). In the case of $\alpha=10^6$, the data is homogeneously distributed across clients, whereas in the case of $\alpha=0.1$, the majority of each client's data comprises two labels at most. The decrease in the homogeneity of the clients' data is also observed by comparing the data distributions generated by $\alpha=10^3$ and $\alpha=10$.}
    \label{fig:alpha-values}
    \vspace{-4.5mm}
\end{figure*}

\vspace{-1.75mm}

\subsection{Summary of Contributions}\label{sec:summary_of_contribution}
\vspace{-0.5mm}

Our main contributions can be summarized as follows:
\begin{description}[font=$\bullet$~\normalfont,leftmargin=4.3mm]
 \itemsep-0.03em 
    \item  We show through extensive experiments that prior FL studies relying on class/label distribution skew fail to impactfully model data heterogeneity for FL simulation in generic (non-classification) computer vision tasks. 
    \item We demonstrate that the notion of data heterogeneity is inherently task-dependent, motivating a shift away from the prevailing ``one-size-fits-all" label-based formulation toward \textit{task-specific data heterogeneity} models in FL.
    \item We introduce \textit{embedding-based data heterogeneity}, a new formulation that induces data heterogeneity from the perspective of each vision task, and show that FL performance can degrade under this revamped  data heterogeneity model.
    \item We outline a set of new research directions for handling task-specific data heterogeneity in single-task and multi-task FL settings involving computer vision tasks.
\end{description}

\vspace{-1mm}
\section{Preliminaries and Motivations} \label{sec:background}
\noindent In this section, we first model an FL system in Sec. \ref{sec:fed-sys-model}, describing the operations performed at both the client and server levels. We then discuss the current approaches for modeling data heterogeneity in FL in Sec. \ref{sec:dirichlet-approach}, where we propose an experiment to evaluate these methods for computer vision tasks and highlight their limitations.
% In this section, we first describe the modeling of an FL system  in Sec. \ref{sec:fed-sys-model}, which contains the operations done in both client and server levels in training a model in a federated fashion.
% We then discuss the current approaches on modeling the data heterogeneity in FL in Sec. \ref{sec:dirichlet-approach}, in which we propose an experiment to test the capabilities of these approaches for computer vision tasks and discuss their limitations.

% \ali{Let us stick to "clients" instead of "devices". I have done so already! Make sure to do the same when drafting the simulation results.}

\vspace{-2.5mm}
\subsection{FL System Modeling} \label{sec:fed-sys-model}
We consider executing FL over a set of clients denoted by $\mathcal{N}$. Each client $n \in \mathcal{N}$ is assumed to possess a local dataset  $\mathcal{D}_n$ with size $D_n = |\mathcal{D}_n|$. We also let  $\mathcal{D} = \bigcup_{n \in \mathcal{N}} \mathcal{D}_n$ with size $D = |\mathcal{D}|=\sum_{n\in\mathcal{N}}D_n$ denote the global dataset, encompassing the datasets distributed across the clients. For an ML task of interest, we let $\ell(\bm{\omega};d)$ denote the loss of the task under arbitrary ML model parameter $\bm{\omega}$ and datapoint $d$. Subsequently, the local loss function of each client $n$ is given by $\mathcal{L}_n(\bm{\omega})=\frac{1}{D_n}\sum_{d\in\mathcal{D}_n} \ell(\bm{\omega};d)$. The goal of FL is to find the optimal model parameter $\bm{\omega}^\star$ to minimize the global loss function $\mathcal{L}(\bm{\omega})=\frac{1}{D}\sum_{d\in\mathcal{D}} \ell(\bm{\omega};d)$ for the ML task of interest, which can be mathematically represented as follows:
\begin{equation}
    \bm{\omega}^\star = \argmin_{\bm{\omega}} \mathcal{L}(\bm{\omega}) \equiv \argmin_{\bm{\omega}} \sum_{n \in \mathcal{N}}{D_n \mathcal{L}_n(\bm{\omega})}\big/{D}.
\end{equation}

To achieve $\bm{\omega}^\star$, FL follows an iterative approach consisting of a series of \textit{local model training rounds} at the clients and \textit{global model aggregation rounds} at the server. During each FL global aggregation round $t \in \{0,\cdots,T-1\}$, the central server first broadcasts the global model, $\bm{\omega}^{(t)}$, to all clients (with $\bm{\omega}^{(0)}$ typically initialized randomly).
% In particular, during each FL global aggregation round $t \in \{0,\cdots,T-1\}$, the global model, $\bm{\omega}^{(t)}$, preserved at the central server, is first broadcasted to all clients ($\bm{\omega}^{(0)}$ is often randomly initialized).
Each client $n$ will then initialize its local model  as
$\bm{\omega}_n^{(t),0}~=~\bm{\omega}^{(t)}$ and
performs $K$ rounds of local ML model training via mini-batch stochastic gradient descent (SGD), each indexed by $k$, where $1\leq k\leq K$. In particular, the local model of client $n$ after the $k$-th SGD iteration  during the global aggregation round $t$ is given by: 
\begin{equation}
    \bm{\omega}_n^{(t), k} = \bm{\omega}_n^{(t),k-1} - \eta^{(t), k}_n\widetilde{\nabla}\mathcal{L}_n(\bm{\omega}_n^{(t),k-1}),
\end{equation}
where $\eta^{(t), k}_n$ is the instantaneous learning-rate. Also, $\widetilde{\nabla}$ denotes the stochastic gradient approximation defined as  $\widetilde{\nabla}\mathcal{L}_n(\bm{\omega}_n^{(t),k-1}) =  \frac{1}{|\mathcal{B}_n^{(t), k-1}|} \sum_{d \in \mathcal{B}_n^{(t), k-1}} \nabla \ell(\bm{\omega}_n^{(t),k-1};d)$, where $\mathcal{B}_n^{(t), k-1}\subseteq \mathcal{D}_n$ represents the set of randomly sampled datapoints contained in the mini-batch.

\begin{figure*}[!ht]
\vspace{-5mm}
    \centering
    \includegraphics[width=.8\textwidth]{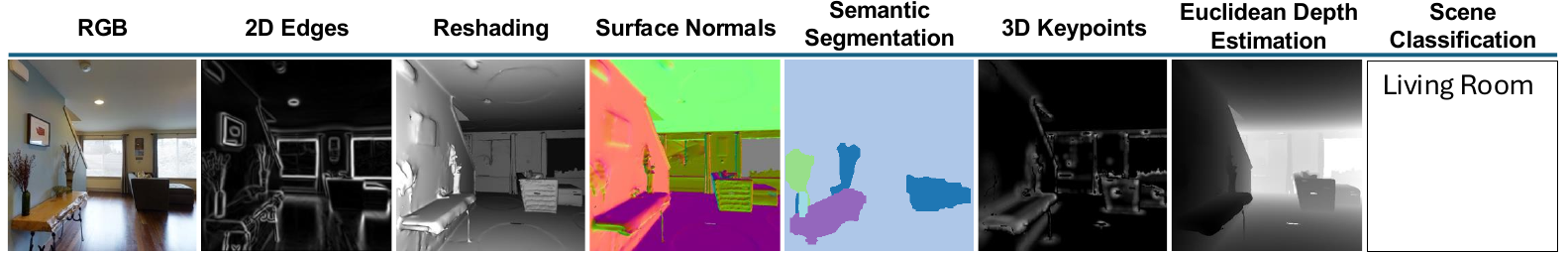}
    \caption{Visualization of the tasks from the Taskonomy dataset that have been used in our experiments. Unless otherwise stated, the loss functions used for the tasks are as follows: $\ell_1$ loss for Euclidean Depth Estimation, 2D Edges, Surface Normals, and 3D Keypoints; mean squared error for Reshading; and cross-entropy for Scene Classification and Semantic Segmentation.}
    \label{fig:taskonomy}
    \vspace{-5mm}
\end{figure*}

After the conclusion of local model training, clients send their latest trained local ML models (i.e.,  $\bm{\omega}_n^{(t), K}$ for each client $n$) to the server, where the received models are aggregated to create a new global model as follows:
\begin{equation}
    \bm{\omega}^{(t+1)} = \sum_{n\in\mathcal{N}} {D_n \bm{\omega}_n^{(t),K}}\big/{D}.
\end{equation}
This new global model is then broadcast back to the clients to start the next local model training round and this iterative procedure is repeated until a desired convergence criterion is achieved.  
Note that the procedure formulated above describes the vanilla FedAvg method~\cite{fed-avg}. Other variations of FL, which are studied later in Sec.~\ref{sec:benchmarks}, involve modifications to the above-described local training and global aggregation processes.

\vspace{-3mm}
\subsection{Analyzing Current Approaches on Modeling Heterogeneity} \label{sec:dirichlet-approach}

\subsubsection{Importance and Impact of Data Heterogeneity in FL}
In FL, data is collected and stored locally by the clients, never being transferred over the network. Consequently, the server has often no control over the composition or distribution of the clients' local datasets. This leads to \textit{data heterogeneity} (commonly referred to as non-IID data\footnote{``non-IID" stands for ``non-independent and identically distributed."}) across clients, a fundamental challenge in FL. Specifically, when the datasets $\mathcal{D}_n$ and $\mathcal{D}_{n'}$ differ significantly across clients ($n \neq n'$), variations in local data distributions introduce \textit{local model bias} during the local SGD iterations at the clients. This bias can degrade the overall performance of the FL global model. In particular, non-IID datasets result in local models that capture client-specific characteristics, which creates inconsistencies during model aggregation and exacerbates issues such as divergence and slow convergence. \textit{Modeling} and \textit{addressing} data heterogeneity are therefore critical for advancing FL.  In this work, we aim to introduce a novel \textit{modeling} approach for non-IID data in FL, leaving \textit{addressing} it to future research. Specifically, we focus on highlighting the inefficiencies of existing methods for modeling non-IID data in FL, particularly for computer vision tasks. Our proposed method provides a generic framework for data heterogeneity that not only extends to computer vision tasks but also encompasses existing methods as special cases.

\subsubsection{State-of-the-art (SoTA) Methods in Modeling Data Heterogeneity}

% \kasra{Since we're talking about classification tasks, shouldn't we change this into \textit{classified data heterogeneity}?}
SoTA methods in FL that concern modeling the data heterogeneity \cite{qu2022rethinking, fl-dirichlet, darzi2024tackling, he2021fedcv} have heavily relied on classification tasks. In this setting, the global dataset often has a limited set of class labels, which streamlines the emulation of non-IID data across the clients through imposing imbalanced class label distribution across their local datasets. More trivial approaches \cite{li2022federated, chen2024feddat} include ensembling datasets to create a heterogeneous data gathering setting \cite{chen2024feddat} and partially distributing labels between clients to emulate non-IID data (i.e., each client's local dataset contains data with a predetermined ratio from each class label)\cite{li2022federated}. 
However, the most common approach  is a theoretically-oriented method, which utilizes the \textit{Dirichlet distribution} \cite{he2021fedcv, fl-dirichlet,li2022federated,wang2020tackling, mendieta2022local, darzi2024tackling} to emulate non-IID data. This SoTA  method considers the ratio of each available class label in each client's local dataset as a random variable drawn from Dirichlet distribution which can be tuned with specific parameters. In the following, we first explain this SoTA method, which induces class-based data heterogeneity across the clients, 
and then study its effectiveness in more generic non-classification computer vision tasks.

\subsubsection{Creating non-IID Data via Dirichlet Distribution}\label{sec:specDric}
To model the class-based data heterogeneity using the Dirichlet distribution in FL, a set of classification labels $\mathcal{C}$ with each member $c\in\mathcal{C}$ representing a class label in the global dataset is considered (e.g., each digit in MNIST \cite{mnist} dataset). 
Then, parameter
$\mathbf{p} = \{p_c\}_{c\in\mathcal{C}}$ is used for the set of ratios of each class $c$ in the global dataset $\mathcal{D}$ which is being dispersed across clients.
Afterwards, a label distribution vector is created for each client $n$ as: $\mathbf{q_n} = \mathsf{Dir}(\alpha\mathbf{p}) = \{q_{n,c}\}_{c\in\mathcal{C}}$ where $q_{n,c}$ denotes the ratio of client $n$'s dataset containing data from class label $c$. Here, $\alpha \geq 0$ is the \textit{concentration parameter} which determines how closely each $\mathbf{q_n}$ will resemble $\mathbf{p}$, with higher $\alpha$ leading to a more similar $\mathbf{q_n}$ to $\mathbf{p}$. We can therefore interpret decreasing in the concentration parameter $\alpha$ as increasing the data heterogeneity across the clients. This phenomenon is visualized in Fig. \ref{fig:alpha-values} for an FL system with $25$ clients containing $16$ labels (similar results apply to other datasets such as MNIST~\cite{mnist}, Fashion MNIST~\cite{xiao2017} and CIFAR-10~\cite{cifar10}). In the figure, data points with the same class label are shown in similar color tones. As illustrated, reducing $\alpha$ from $10^6$ to $0.1$ corresponds to increasing data heterogeneity, with $\alpha = 0.1$ representing an extreme case of non-IID data where most clients' datasets contain fewer than two distinct labels. As a result, it can be construed that smaller values of $\alpha$ represent emulation of scenarios where 
local datasets of clients induce severe local model bias during the local training rounds and negatively affect the convergence of FL.

% Due to its fundamental reliance on class label distribution, 
% emulation of non-iid data using Dirichlet distribution is a powerful method when classification tasks are considered in FL, which has also led to its popularity as the SoTA method in the literature~\cite{li2022federated,wang2020tackling}. Nevertheless, we next demonstrate that this method loses its effectiveness when FL is considered for more complex computer vision tasks.

Due to its reliance on class-label distributions (which are readily available for classification tasks) and its tunable concentration parameter $\alpha$, the Dirichlet distribution has become the standard and widely adopted method for emulating non-IID data in FL~\cite{li2022federated,wang2020tackling}. Specifically, its mathematical structure allows practitioners to precisely control data imbalance across clients, which explains its prominence as the de facto data heterogeneity model in the FL literature. Nevertheless, we next show that this method loses its effectiveness when FL is considered for more complex computer vision tasks.

\subsubsection{FL for Non-Classification Vision Tasks under Class-based Data Heterogeneity} \label{sec:class-based-experiments}
We next take one of the well-known computer vision datasets, called Taskonomy~\cite{taskonomy}, which has been previously used in both centralized \cite {sax2018mid} and federated \cite{zhuang2023mas} multi-task learning. We utilize the above-described Dirichlet distribution to create a heterogeneous allocation of data across  $25$ clients' local datasets in FL using the available class labels, referred to as ``class\_scene" in this dataset.\footnote{The classes are provided in the dataset as probabilities for each datapoint belonging to each available class, from which we choose the class with the highest probability as the determined class for each datapoint. More details on the data processing are provided in Sec. \ref{sec:app-data-preprocessing}.}
In our experiments, we select a representative set of vision tasks in this dataset, which are visualized in Fig.~\ref{fig:taskonomy} and include ``2D Edges", ``Reshading", ``Surface Normals", ``Semantic Segmentation", ``3D Keypoints", ``Euclidean Depth Estimation", and ``Scene Classification" (unless otherwise stated, the loss functions used for the tasks are as follows: $\ell_1$ loss for Euclidean Depth Estimation, 2D Edges, Surface Normals, and 3D Keypoints; mean squared error for Reshading; and cross-entropy for Scene Classification and Semantic Segmentation). 

\begin{figure*}[!h]
\vspace{-2mm}
    \centering
    \begin{tcolorbox}[colback=BoxBg, colframe=BoxFrame, width=0.49\textwidth, left=0pt, right=0pt, top=-2pt, bottom=-2pt, after=\hspace{2mm}, title=2D Edges --- \textbf{FedAvg}, halign title=flush center, fonttitle=\small,     toptitle=-3pt,
    bottomtitle=-3pt]
    \includegraphics[width=0.48\textwidth]{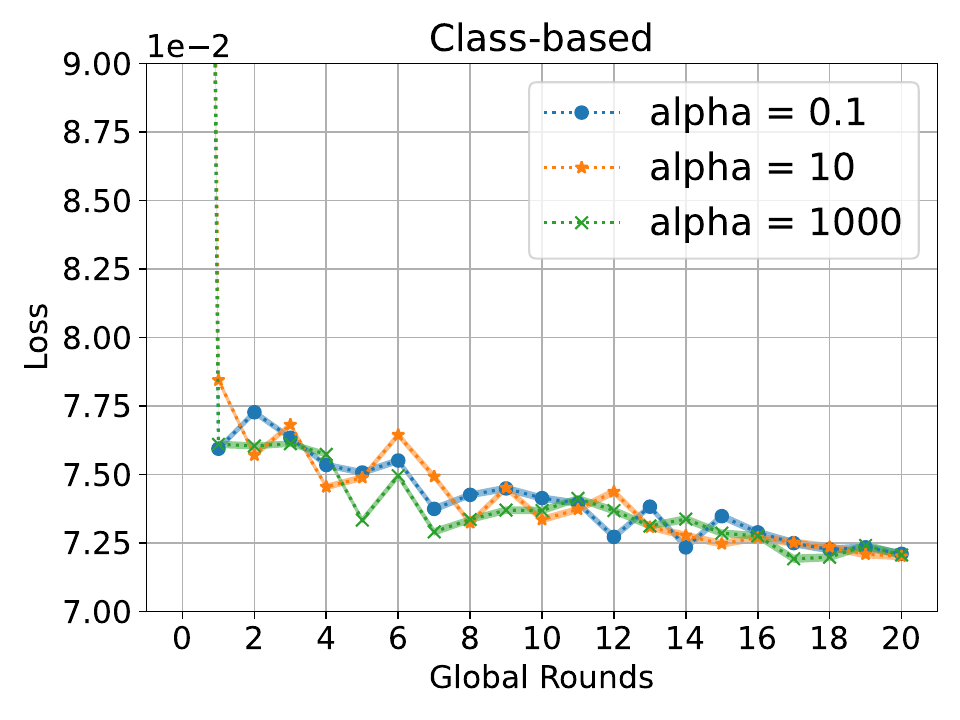}
    \includegraphics[width=0.48\textwidth]{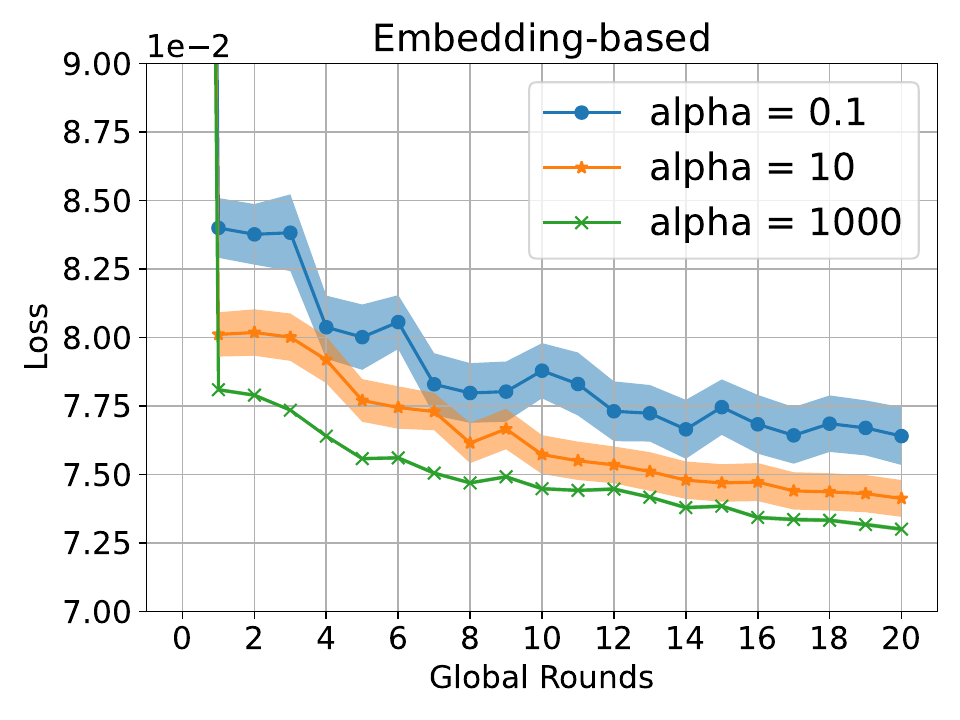}
    \end{tcolorbox}
    \hfill
    \begin{tcolorbox}[colback=BoxBg, colframe=BoxFrame, width=0.49\textwidth, left=0pt, right=0pt, before=,  top=-2pt, bottom=-2pt, title=Reshading --- \textbf{FedAvg}, halign title=flush center,toptitle=-3pt,
    bottomtitle=-3pt]
    \includegraphics[width=0.48\textwidth]{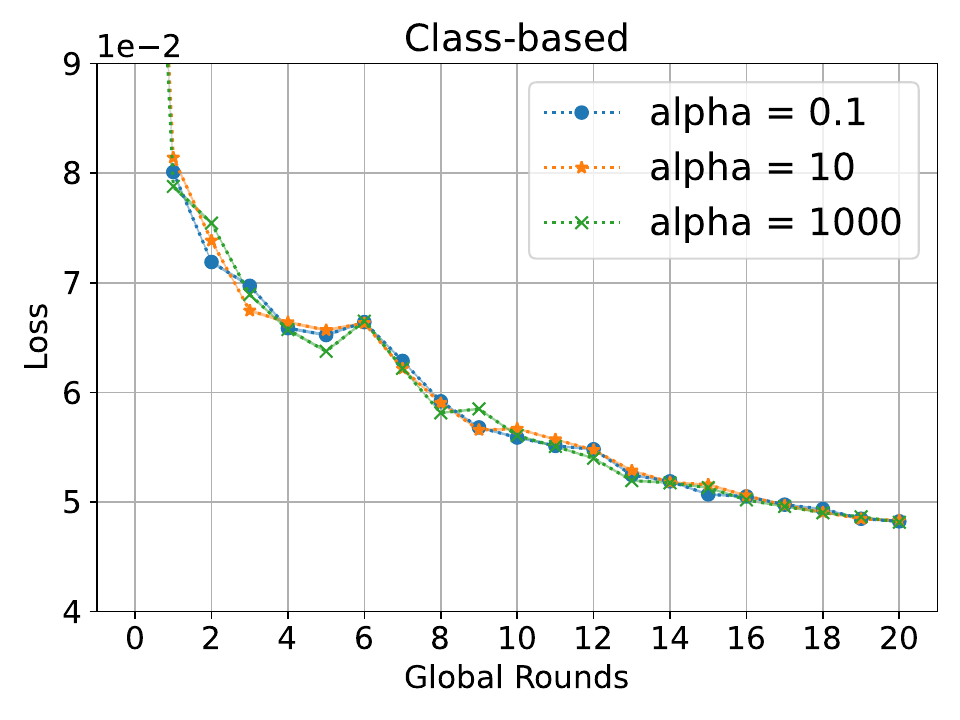}
    \includegraphics[width=0.48\textwidth]{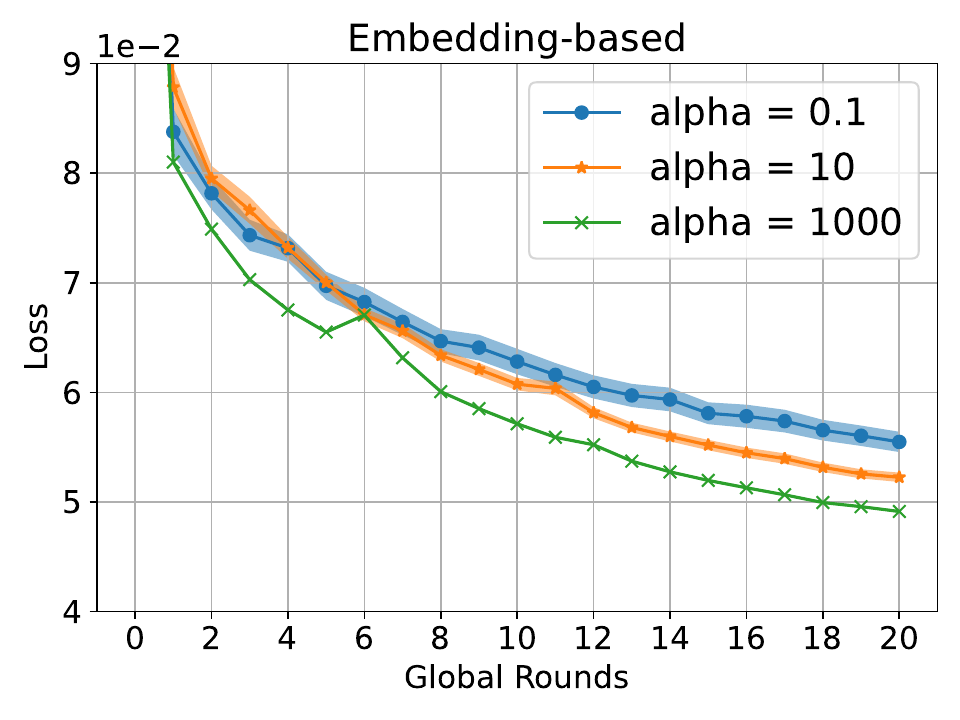}
    \end{tcolorbox}
     \vspace{-3.600mm}
     
    \begin{tcolorbox}[colback=BoxBg, colframe=BoxFrame, width=0.49\textwidth, left=0pt, right=0pt, top=-2pt, bottom=-2pt, after=\hspace{2mm}, title=Surface Normals --- \textbf{FedAvg}, halign title=flush center,toptitle=-3pt,
    bottomtitle=-3pt]
    \includegraphics[width=0.49\textwidth]{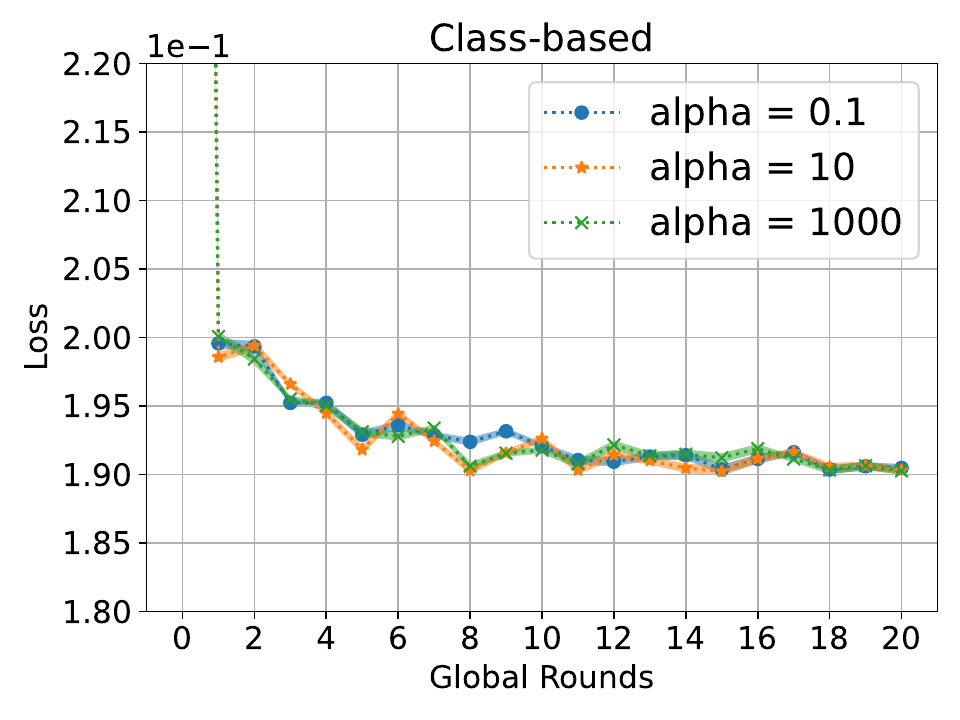}
    \includegraphics[width=0.49\textwidth]{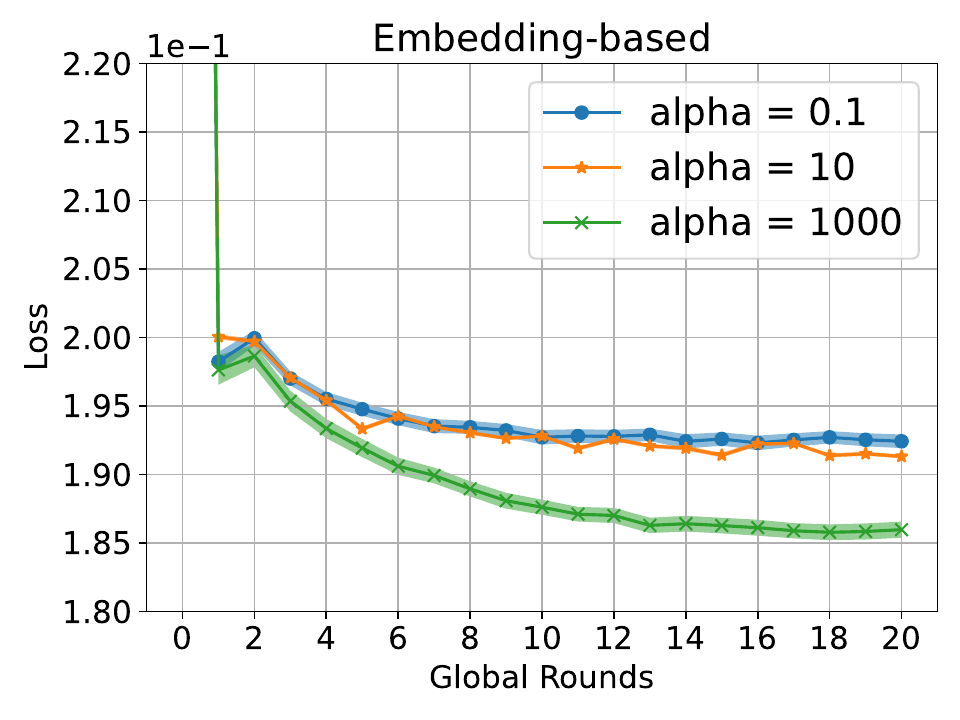}
    \end{tcolorbox}
    \hfill
    \begin{tcolorbox}[colback=BoxBg, colframe=BoxFrame, width=0.49\textwidth, left=0pt, right=0pt, top=-2pt, bottom=-2pt, before=, title=Semantic Segmentation --- \textbf{FedAvg}, halign title=flush center,toptitle=-3pt,
    bottomtitle=-3pt]
    \includegraphics[width=0.49\textwidth]{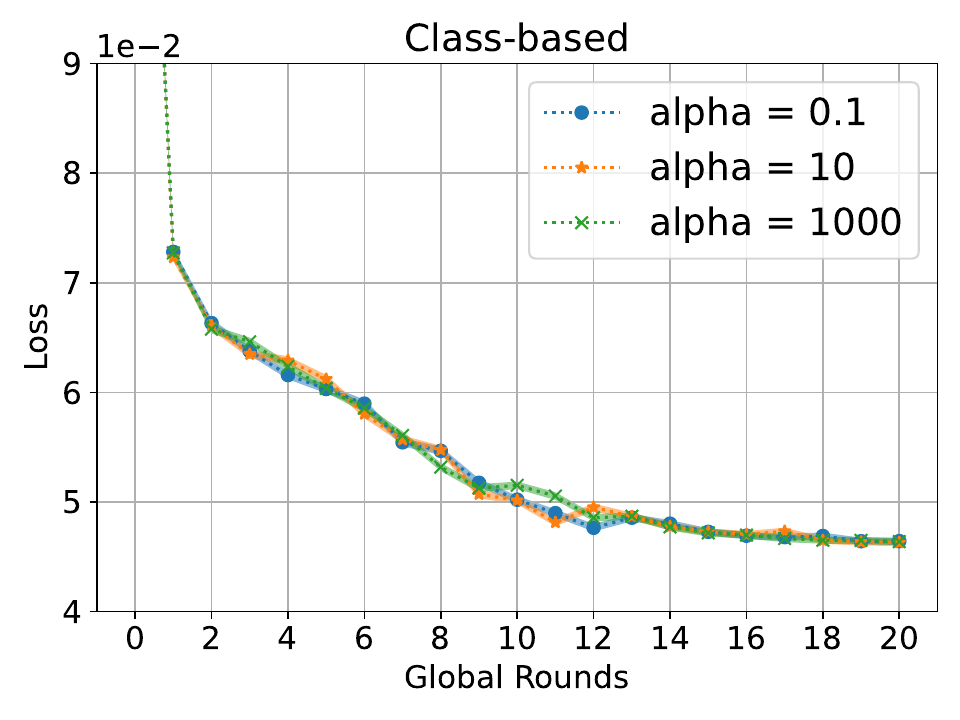}
    \includegraphics[width=0.49\textwidth]{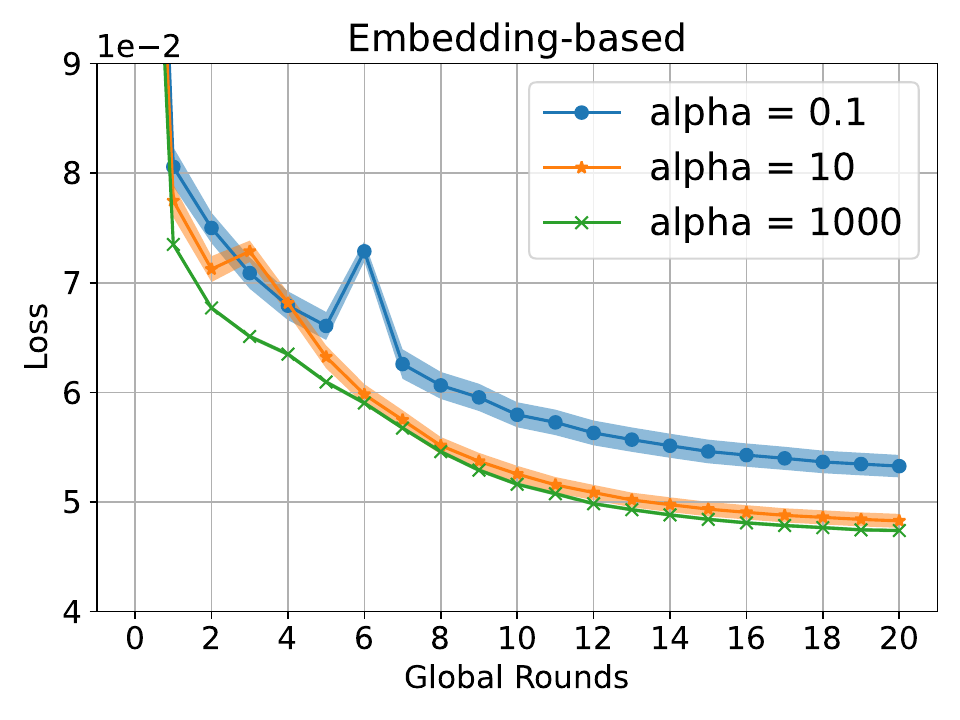}
    \end{tcolorbox}
 \vspace{-3.600mm}
 
    \begin{tcolorbox}[colback=BoxBg, colframe=BoxFrame, width=0.49\textwidth, left=0pt, right=0pt, top=-2pt, bottom=-2pt, after=\hspace{2mm}, title=Euclidean Depth Estimation --- \textbf{FedAvg}, halign title=flush center,toptitle=-3pt,
    bottomtitle=-3pt]
    \includegraphics[width=0.49\textwidth]{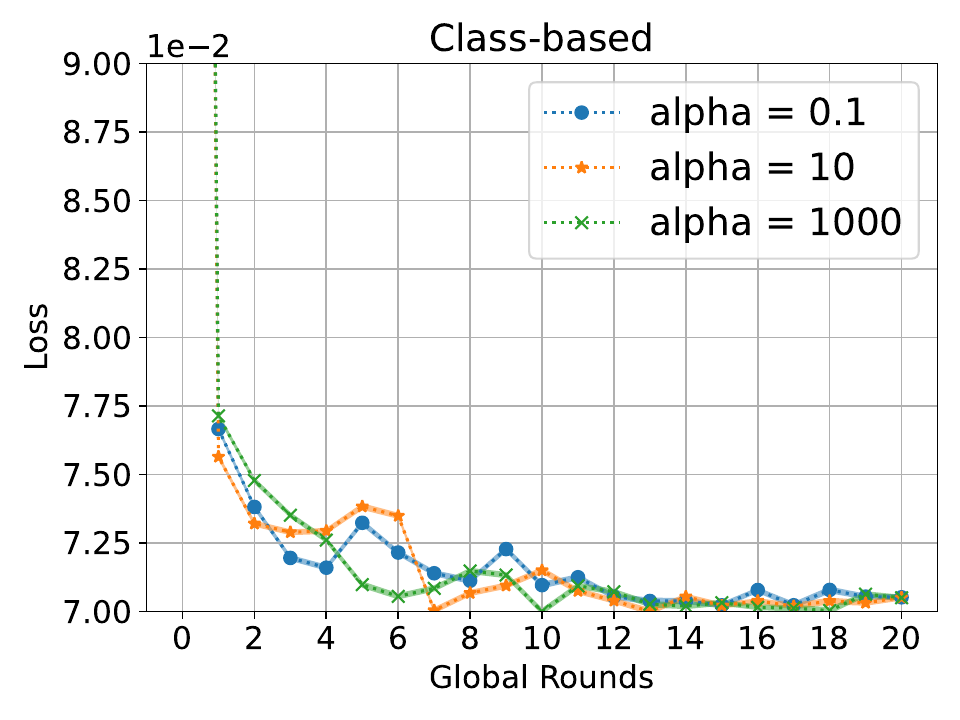}
    \includegraphics[width=0.49\textwidth]{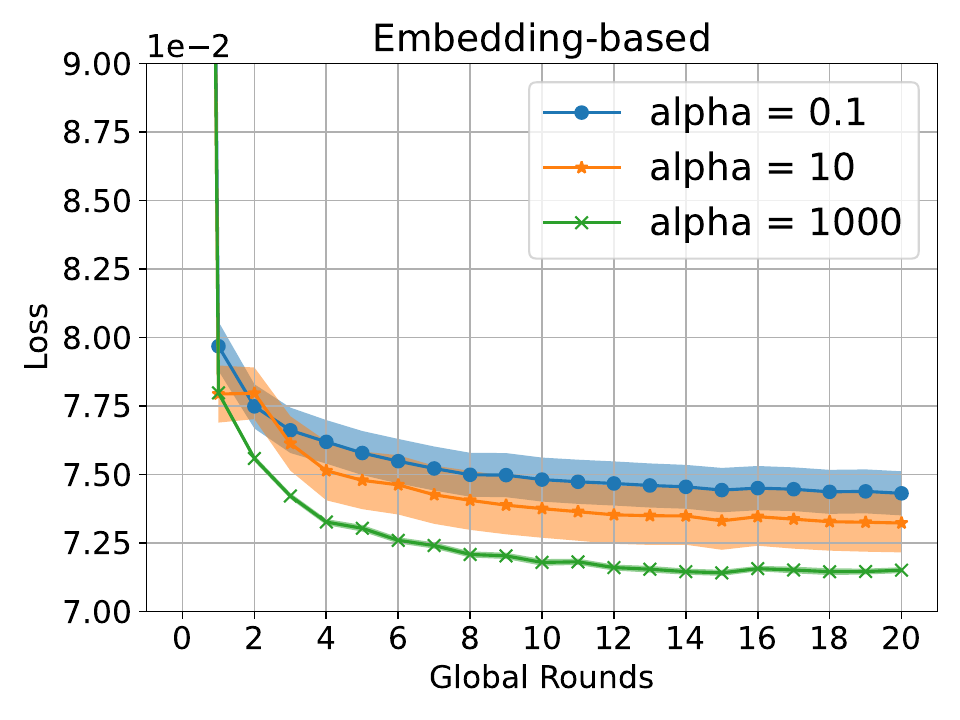}
    \end{tcolorbox}
    \hfill
    \begin{tcolorbox}[colback=BoxBg, colframe=BoxFrame, width=0.49\textwidth, left=0pt, right=0pt, top=-2pt, bottom=-2pt, title=3D Keypoints --- \textbf{FedAvg}, halign title=flush center, before=, toptitle=-3pt,
    bottomtitle=-3pt]
    \includegraphics[width=0.49\textwidth]{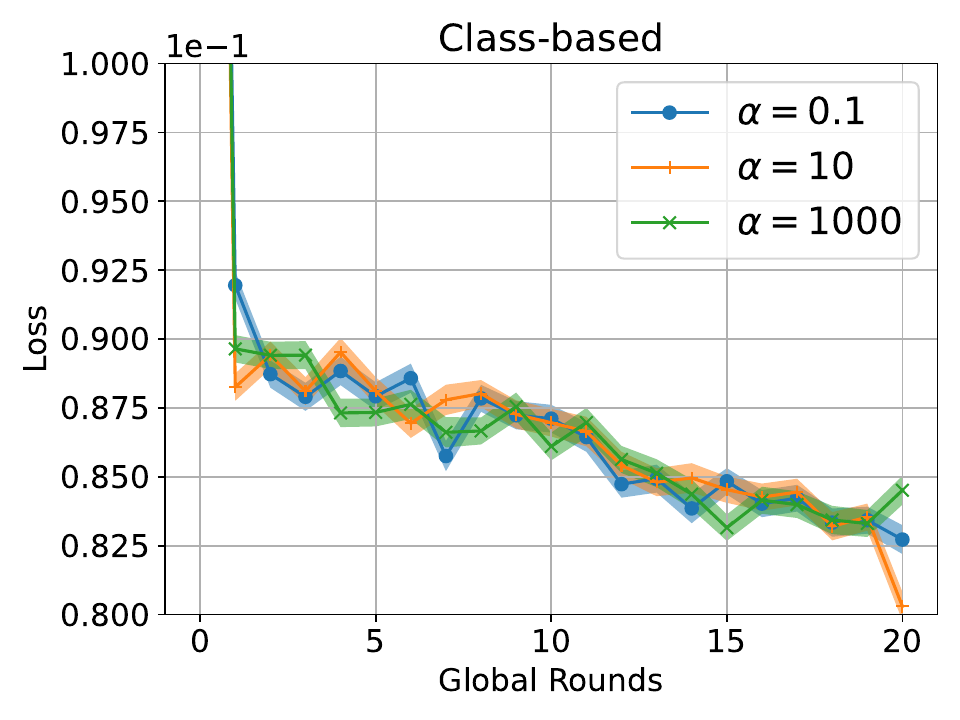}
    \includegraphics[width=0.49\textwidth]{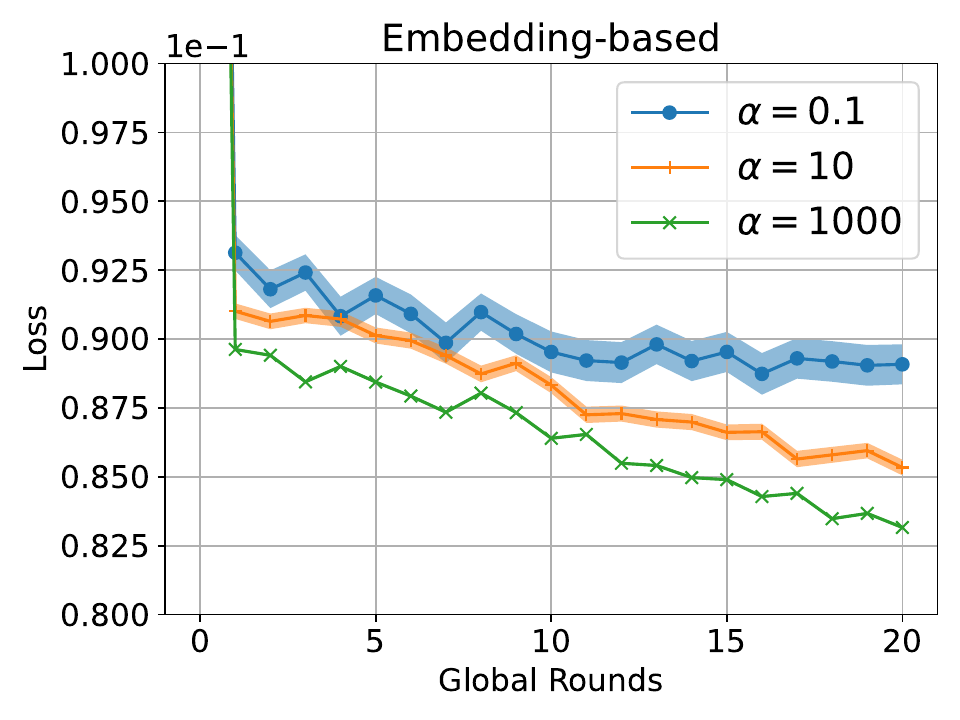}
    \end{tcolorbox}
     \vspace{-3.600mm}
     
    \begin{tcolorbox}[colback=BoxBg, colframe=BoxFrame, width=0.49\textwidth, left=0pt, right=0pt,  top=-2pt, bottom=-2pt, after=\hspace{2mm}, title=Scene Classification --- \textbf{FedAvg}, halign title=flush center,toptitle=-3pt,
    bottomtitle=-3pt]
    \includegraphics[width=0.49\textwidth]{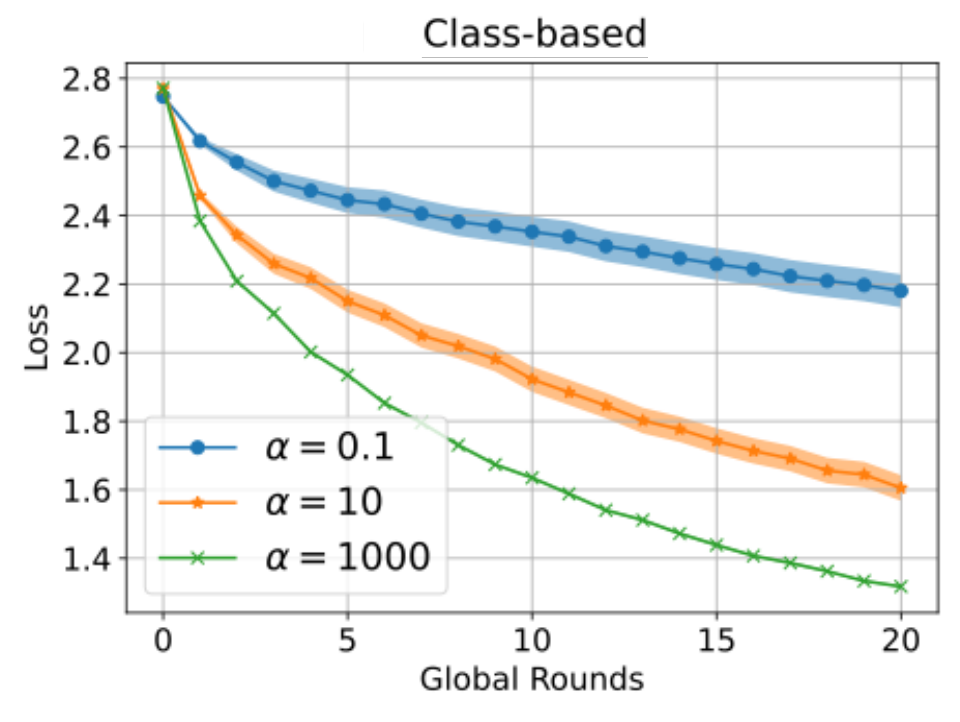}
    \includegraphics[width=0.49\textwidth]{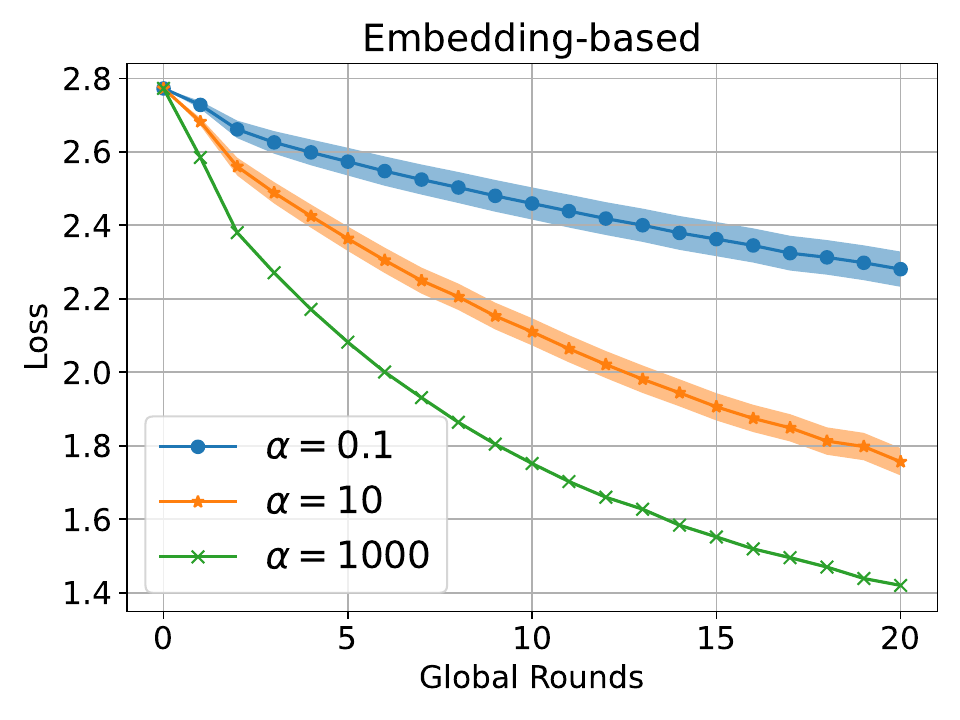}
    \end{tcolorbox}    
    \vspace{-1mm}
    \caption{\textbf{Class-based vs. Embedding-based distribution:} Comparison of how performing Dirichlet distribution over the datapoints' labels (equivalently, the scene class feature) and the extracted embeddings affect the performance of FL.  Solid lines show the mean loss across clients; the shaded region denotes client-level dispersion (standard deviation of the loss across clients) at each global round.  As it is shown, distributing the datapoints based on the class/label of the scene they correspond to (the left plot in each box) has not resulted in a significant change in the performance across various values of $\alpha$. However, distributing the datapoints based on the clusters formed by the extracted embeddings (the right plot in each box) with the Dirichlet distributions of various $\alpha$ parameters has created a performance gap across the possible scenarios with  $\alpha=0.1$ (i.e., the most heterogeneous case) resulting in the worst performance and $\alpha=1000$ (i.e., the least heterogeneous case) resulting in the best performance.  The results further show that class-based data heterogeneity overestimates FL performance: it yields seemingly small loss values because it does not perturb the task-relevant feature space. In contrast, embedding-based heterogeneity increases the loss,  revealing the sensitivity of FL methods to task-relevant feature variations that are not captured by class-based simulation splits in computer vision tasks. Also, it can be seen that only the class-based experiments on the Scene Classification task (i.e., the bottom-most box) exhibit a notable performance change with varying $\alpha$, mirroring the trend seen in the embedding-based experiments. This indicates that the most salient feature for the Scene Classification task (i.e., the semantic class label) is naturally captured in the embeddings as well, leading to consistent changes in loss across both class-based and embedding-based settings.}
    \label{fig:embedding-vs-class-p1}
        \vspace{-4mm}
\end{figure*}

These vision tasks, except Scene Classification, can be construed as \textit{non-classification tasks} since they are not associated with a class label. Following the non-IID data induced by Dirichlet distribution across the clients, we study the performance of the FL global model under different levels of heterogeneity (i.e., various $\alpha$ values) for the aforementioned vision tasks. The results of this experiment are shown in Fig.~\ref{fig:embedding-vs-class-p1} (the left plot in each box, titled ``Class-based").

% . Based on the effects of non-IID data distribution observed on visual classification tasks in \cite{fl-dirichlet}, we expect the global loss function ($\mathcal{L}(\bm{\omega}^{(t)})$) to converge to a higher value as $\alpha$ decreases. We consider the tasks ``2D Edges," ``Reshading," ``Surface Normals," ``Semantic Segmentation," ``3D Keypoints," ``Euclidean Depth Estimation," ``segment\_unsup2d," "Curvature Estimation," and ``class\_scene."  We then plot the results for each task's average global model loss across various $\alpha$ values to account for different levels of heterogeneity. The results of this experiment are shown in Fig. \ref{fig:embedding-vs-class}.

Focusing on the \textit{only} classification task, i.e., Scene Classification (bottom box), we can see that decreasing $\alpha$ has a notable impact on the performance of the task. However, when focusing on all the other non-classification vision tasks, \textit{we can see no significant degeneration in the performance of the global model as the value of $\alpha$ decreases.} This key observation signals the ineffectiveness of the class-based (equivalently, label-based) data heterogeneity emulation methods for generic computer vision tasks and calls for revamping the approaches used to induce data heterogeneity for such tasks in the FL setting. 

% Intuitively, the ineffectiveness of label-based data heterogeneity emulation methods for generic computer vision tasks stems from the fact that non-classification vision tasks do not necessarily \textit{perceive} data distribution in the same way as classification tasks. In fact, \textit{their perception of data heterogeneity can be label-independent and majorly different from each other: an aspect largely overlooked in the existing FL literature.} For instance, consider the “2D Edges” task illustrated in Fig.~\ref{fig:taskonomy}. This task is more sensitive to the configuration of edges within an image rather than the image’s label. In this context, two datapoints with the same label (e.g., both with “Living Room” label) may exhibit highly different edge structures, rendering them heterogeneous from the perspective of the “2D Edges” task. Conversely, two datapoints with different labels (e.g., “Living Room” and “Kitchen”) might have very similar edge structures, making them homogeneous for this specific task despite having different labels. This highlights the need for methodologies in FL that emulate data heterogeneity based on the unique perspectives of vision tasks to data.

Intuitively, class-based data heterogeneity emulation methods often fall short for generic computer vision tasks because non-classification vision tasks \textit{perceive} data distribution differently from classification tasks. In fact, their perception of data heterogeneity can be \textit{label-independent} and vary significantly across tasks, an aspect that has largely been overlooked in FL literature. 
This happens because labels provide only coarse semantic groupings that do not necessarily reflect the underlying feature distributions relevant to each vision task. 
For example, consider the Euclidean Depth Estimation task, which focuses on estimating the depth of objects in an image. In this context, two images with the same label (e.g., both labeled ``Living Room") may exhibit highly different depth profiles due to variations in camera viewpoint or lighting conditions, making them heterogeneous from the perspective of this task. On the other hand, two images with different labels (e.g., ``Hallway" and ``Living Room") may share similar depth structures, making them homogeneous for this task despite their differing labels. This underscores the need for FL methodologies that account for data heterogeneity from the unique perspectives of each vision task, rather than relying solely on label-based notions of data heterogeneity.

% Intuitively, class-based data heterogeneity emulation methods often fall short for generic computer vision tasks because non-classification vision tasks \textit{perceive} data distribution differently from classification tasks. In fact, their perception of data heterogeneity can be \textit{label-independent} and vary significantly across tasks, an aspect that has largely been overlooked in FL literature. 
% For example, consider the Euclidean Depth Estimation task, which focuses on estimating the depth of objects in an image. In this context, two data points with the same label (e.g., both labeled ``Living Room") could exhibit very different depth structures due to variations in camera angle or lighting conditions, making them heterogeneous from the perspective of this task. On the other hand, two data points with different labels (e.g., ``Hallway" and ``Living Room") may share similar depth structures, making them homogeneous for this task despite their differing labels. This underscores the need for FL methodologies that account for data heterogeneity from the unique perspectives of each vision task.

Motivated by this observation, we propose an approach to emulate data heterogeneity in FL that accommodates both classification and non-classification computer vision tasks.

\begin{figure*}[!h]
\vspace{-4mm}
    \centering
    \includegraphics[width=.85\textwidth, trim=0 0 120 0, clip]{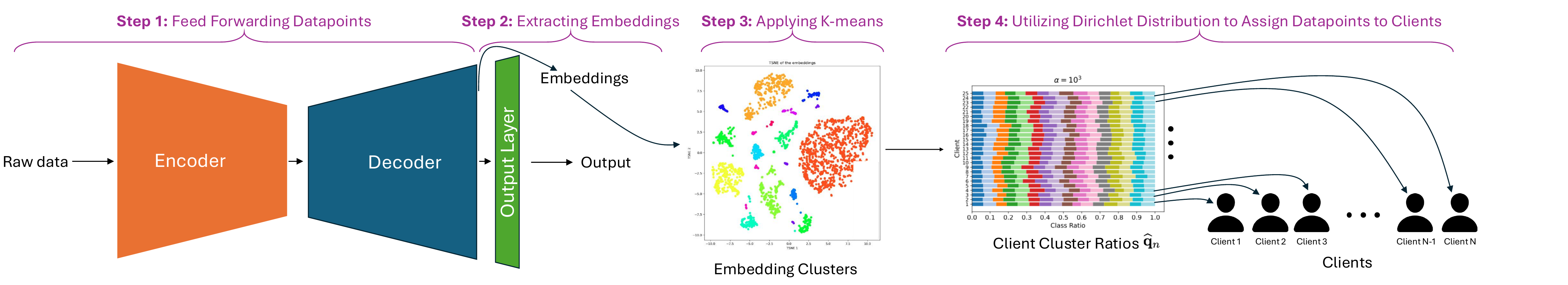}
    \caption{In our approach, data heterogeneity is induced from the unique task perspective. At first (i.e., in Step 1), data points are fed to a pre-trained neural network trained on the task (this can be any off-the-shelf model as long as the model is trained on the task). Afterwards (i.e., in Step 2), 
    the embeddings of data points are extracted from the penultimate layer of the model. Then (i.e., in Step 3), the embeddings are  clustered using clustering algorithms such as K-means, which unveils the similarity/dissimilarity of the data points from the task perspective. Finally (i.e., in Step 4), 
    Dirichlet distribution is  applied on the clustered data to emulate the local datasets of clients, treating the datapoints in the same cluster as having the same \textit{group} (analogous to \textit{labels} in classification tasks).}
    \label{fig:nn-model}
    \vspace{-3.5mm}
\end{figure*}

\section{Emulation of Data Heterogeneity from the Task Perspective}\label{sec:method}
\noindent To address the need for a task-specific modeling of data heterogeneity, we \textit{shifted our perspective} from the raw input domain to the \textit{task output domain} in order to get more task-related information. In particular, instead of using the raw data features, such as its label, we aim to use the processed data features, such as \textit{embeddings}, to emulate data heterogeneity in FL. 
This stems from the unavailability of representative characteristics, such as class labels, in the case of non-classification tasks, motivating us to identify a utility analogous to class labels that can effectively associate or disassociate datapoints based on the unique perspective of a given task. 

In particular, we note that in the case of (deep) neural networks, training the network on different tasks leads to a reconfiguration of its model parameters, where the parameters optimized for each task are often unique to that task. Consequently, feeding a raw data point into a neural network trained for a specific task and observing the outputs of different layers can provide unique representations of how the task perceives the data point. Specifically, the outputs of the initial layers, especially in case of convolutional neural networks (CNNs), often correspond to low-level features or simple patterns, while the outputs of the later layers correspond to high-level, task-specific features  or other abstract representations.
 Following this notion, we aimed to extract the task's perspective of each datapoint by passing each datapoint through a network trained on the task and extracting the output of the penultimate layer as shown in Fig. \ref{fig:nn-model}. We will refer to this extracted output as \textit{embedding} which we use to describe each datapoint from a task's perspective for the remainder of this discussion.

\subsection{Task-Specific Data/Embedding Clustering}
% In a nutshell, in our approach (see Fig.~\ref{fig:nn-model}), data heterogeneity is induced from the unique perspective of the task through pre-processing the data points {\color{blue} to accommodate for an offline and centralized alternative to how the class-based Dirichlet partitions are constructed in prior work}. 

 In a nutshell, in our approach (see Fig.~\ref{fig:nn-model}), data heterogeneity is induced from the unique perspective of the task through an offline pre-processing stage, serving as a centralized client data generation procedure analogous to the class-based Dirichlet partitioning commonly adopted in prior work.  First, data points are fed into a pre-trained neural network trained specifically for the task of interest (any off-the-shelf pre-trained model such as the ResNet\cite{he2016deep} or CLIP \cite{radford2021learning} trained on the target task can be utilized). Next, embeddings for each data point are extracted from the penultimate layer of the model. These embeddings, which encapsulate task-specific features, are then clustered using algorithms such as K-means  \cite{lloyd1982least, macqueen1967some} to reveal the similarity or dissimilarity of data points from the task's perspective  (alternative choices to K-means have been studied in Appendix~J). Finally, a Dirichlet distribution is applied to the clustered data to emulate the local datasets of clients, treating datapoints within the same cluster as belonging to the same \textit{group} (analogous to ``labels" in classification tasks). 

\begin{figure}[!h]
 \vspace{-2mm}
    \centering
    \includegraphics[width=0.32\textwidth]{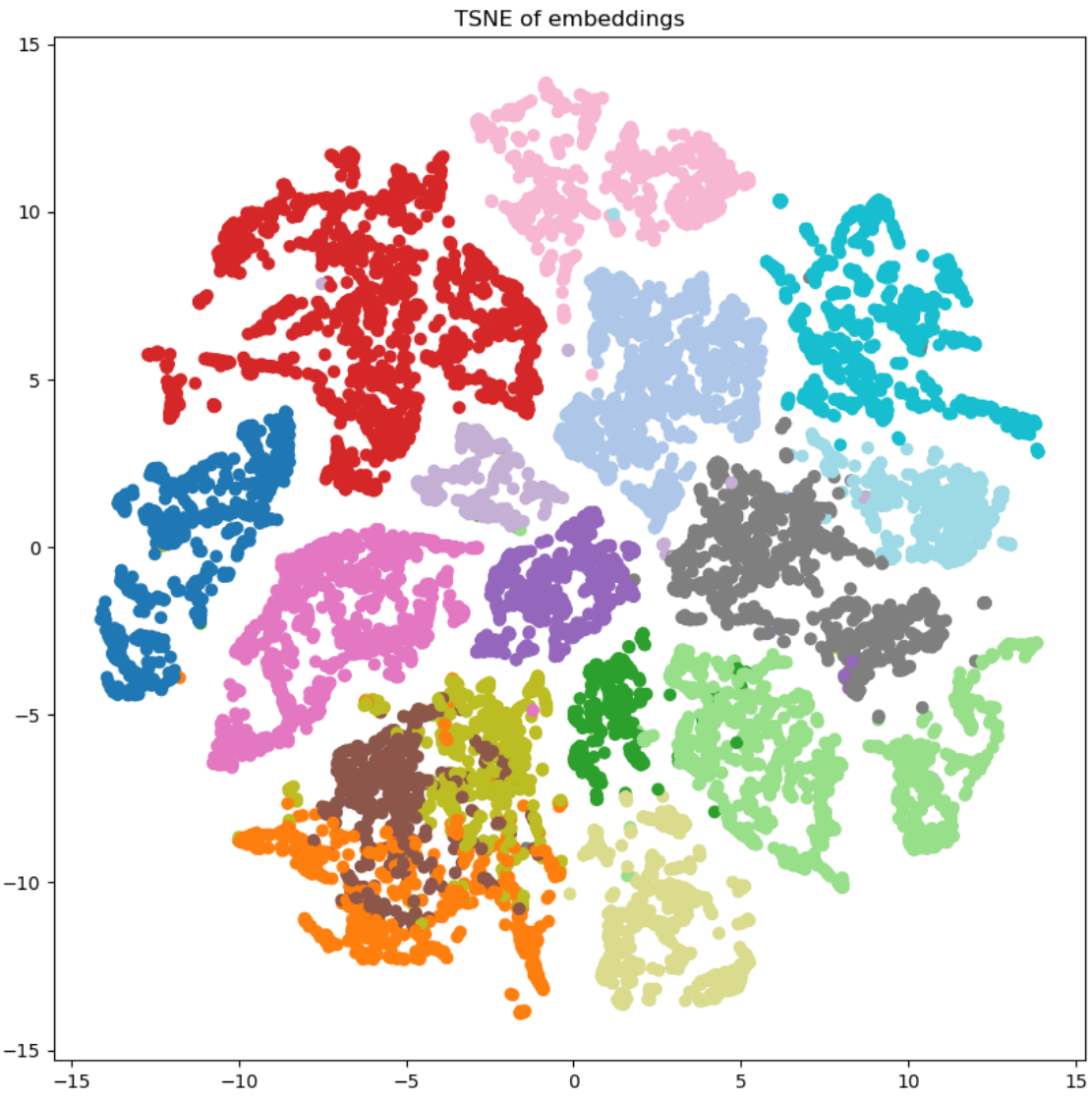}
      \vspace{-2.15mm}
    \caption{The figure illustrates embedding clusters extracted from a \textit{Scene Classification} task model, where each color represents a specific class. It is evident that the embeddings create a separable space corresponding to the scene classification labels. This demonstrates that distributing data points across FL clients using the clusters/groups depicted in the figure is equivalent to distributing data based on the original scene class labels. The observed alignment between the embedding-based clustering and the class labels underscores the effectiveness of using embeddings as a method to induce task-specific data heterogeneity in FL.}
    \label{fig:scene-class-embedding-clusters}
    % \vspace{-1mm}
\end{figure}

Our method enables a task-driven and systematic approach to modeling data heterogeneity in FL, which works both for non-classification as well as classification tasks. In particular, in case of having a classification task, datapoints with the same label are expected to end up in the same cluster/group after conducting the above procedure, which makes class-based data heterogeneity, explained in Sec.~\ref{sec:dirichlet-approach}, a special case of our method.
We demonstrate this via visualizing our method on the Scene Classification task of Taskonomy dataset, which is a classification task, in Fig. \ref{fig:scene-class-embedding-clusters}.
As can be seen, the clustering of the data points with respect to their embeddings has resulted in the datapoints being clustered into separable clusters which possess only datapoints from one label. This mapping between the clustering of data points based on their embeddings and their class labels for classification tasks further affirms that the clusters made from the embeddings can model the data points similarity/dissimilarity accurately from the task perspective and the clusters/groups obtained based on the embeddings can be hypothetically considered as having the same labels when applying the Dirichlet distribution. 

Note that, for non-classification tasks, the clusters formed by the extracted embeddings may vary from those formed based on labels. In particular, we will later study the \textit{overlap} between the clusters formed based on class labels (i.e., for the classification task) and those formed based on data embeddings of non-classification tasks (see Table~\ref{tab:ari_heatmap} in Sec.~\ref{sec:benchmarks}).

\subsection{Emulating Heterogeneous Local Datasets for Clients} \label{sec:embedding-based-experiments}
Formally, once the clustering of datapoints based on their embeddings are obtained, we emulate the local datasets $\mathcal{D}_n$ for each client $n$, where the set of ratios for classes $\mathbf{p}$ detailed in Sec.~\ref{sec:specDric} is now replaced by the distribution of the embeddings in each cluster/group, which we denote by $\widehat{\mathbf{p}}$. 
Also, we let the set $\mathcal{G}$ contain the distinct cluster/group indices (e.g., all $16$ groups in Fig.~\ref{fig:scene-class-embedding-clusters}). In our approach, each client $n$ will be assigned data from each cluster/group based on its own  cluster/group distribution vector $\widehat{\mathbf{q}}_{n} = \mathsf{Dir}(\alpha\widehat{\mathbf{p}})= \{\widehat{q}_{n,g}\}_{g\in\mathcal{G}}$ where $\widehat{q}_{n,g}$ denotes the ratio of client $n$'s dataset containing data from group $g$ (e.g., one of the $16$ groups in Fig.~\ref{fig:scene-class-embedding-clusters}). These local datasets are then used during the local training of FL. 

% We have experimented with the same tasks in Sec. \ref{sec:dirichlet-approach} and included the results in Fig. \ref{fig:embedding-vs-class}.

To demonstrate this revamped notion of data heterogeneity, we revisit the experiments in Sec.~\ref{sec:dirichlet-approach} to evaluate the effect of the induced data heterogeneity, replacing the class-based method with our embedding-based approach, which groups data points based on their embeddings. The results are presented in Fig. \ref{fig:embedding-vs-class-p1} (the right plot in each box, titled ``Embedding-based") next to the class-based results for a better comparison. 
Unlike the findings from Sec.~\ref{sec:dirichlet-approach}, where data was distributed across clients based solely on class labels, we observe a pronounced impact of $\alpha$ (which controls the level of data heterogeneity across clients) on the performance of the trained FL global model. Specifically, as data heterogeneity increases (i.e., as $\alpha$ decreases), the global loss converges to a higher value, indicating a degradation in the model's performance.

These results highlight that task-specific heterogeneity is more effective in modeling a heterogeneous federated system for vision tasks compared to class-based heterogeneity, which has been predominantly used in prior FL studies \cite{qu2022rethinking, mendieta2022local, NEURIPS2021_2f2b2656, fl-dirichlet}.
 It is noteworthy that while we plot loss trajectories to visualize optimization behavior, another primary endpoint can be task-specific evaluation metrics computed at a fixed communication budget (20 global rounds). These standardized metrics are reported in Appendix~D and summarized in Table~\ref{tab:task_metrics_reformatted} for both data partitioning protocols as we will explain later. 
Furthermore, \textit{the results reveal that the performance of FL for non-classification generic computer vision tasks has been overestimated in prior simulations \cite{zhuang2023mas, lu2024fedhca2, mortaheb2022personalized,cai2023many} as these works rely on class-based data heterogeneity}, which does not reflect the true data variations that vision tasks depend on (e.g., geometry, depth, texture, illumination). In particular, when the data heterogeneity model fails to capture these task-relevant attributes, it leads to an overly optimistic assessment of FL performance under heterogeneous data conditions. Moreover, this experiment underscores that the actual performance and robustness of FL, and its variants, in handling data heterogeneity for generic computer vision tasks remain largely unknown -- a critical gap in the literature. We address this gap through the benchmarks and experiments presented in Sec.~\ref{sec:benchmarks}.

% These results highlight that task-specific heterogeneity is more effective in modeling a heterogeneous federated system for vision tasks compared to class-based heterogeneity, which has been predominantly used in prior FL studies \cite{qu2022rethinking, mendieta2022local, NEURIPS2021_2f2b2656, fl-dirichlet}. Furthermore, \textit{the results reveal that the performance of FL for non-classification generic computer vision tasks has been overestimated in previous works \cite{zhuang2023mas, lu2024fedhca2, mortaheb2022personalized,cai2023many} due to the reliance on class-based heterogeneity}, which fails to adequately emulate the local datasets of clients. Moreover, this experiment underscores that the actual performance and robustness of FL, and its variants, in handling data heterogeneity for generic computer vision tasks remain largely unknown -- a critical gap in the literature. We address this gap through the benchmarks and experiments presented in Sec.~\ref{sec:benchmarks}.

In addition to the above-described main takeaways of the results, our approach has broader implications, which we elaborate on through the following remarks.

\begin{remark}
    [Implications on Designing Effective Countermeasures to Address Data Heterogeneity in FL for Vision Tasks] Our work demonstrates that the prior notion of data heterogeneity, primarily based on the ratio of data points belonging to specific classes, is largely applicable only to classification tasks. By contrast, our method introduces a task-specific approach to inducing data heterogeneity, enabling experiments in FL to be extended to non-classification vision tasks. This shift not only broadens the scope of FL research to include a wider variety of vision tasks but also lays the groundwork for designing vision-specific methods to effectively handle and address/combat data heterogeneity in FL.
\end{remark}

\begin{remark}
    [Implications on the Concept of Task Similarity]\label{remark:task-similarity} Following the footsteps of prior work on modeling task similarity in the centralized ML domain \cite{tasksimilarity, le2022fisher}, our approach opens a model-driven perspective to quantify the similarity between two tasks. Specifically, this involves clustering data points based on their embeddings obtained from the models pre-trained for each task and comparing the resulting clusters to assess similarity. Using this straightforward approach, we establish a baseline performance for task similarity analysis, with results discussed in Sec. \ref{subsubsec:heatmap}. While this serves as a preliminary exploration, more sophisticated methods for task similarity measurement in the FL setting remain an open area for future investigation. Furthermore, this analysis provides an avenue for developing new task grouping strategies in FL, akin to prior efforts in centralized ML \cite{zamir-taskgrouping}. For example, by employing the task similarity measures based on the embedding-based clustering of data points, tasks with highly similar clustering patterns can be grouped and trained collaboratively, potentially enhancing FL performance for related tasks. We leave further exploration of this notion as future work.
\end{remark}

\begin{remark}
   [Implications on Multi-Task FL and its Extensions] \label{remark:multi-task}
   A fundamental implication of our method lies in its application to multi-task FL and its extensions.  In conventional multi-task FL \cite{cai2023many,mortaheb2022personalized,smith2017federated}, each client typically has a single local task, with tasks varying across clients. On the other hand, the extended variations of  multi-task FL \cite{zhuang2023mas, lu2024fedhca2} consider scenarios where multiple tasks coexist on a single FL client.
In the context of  conventional multi-task FL, our method enables researchers to explore how varying levels of data heterogeneity, from the perspectives of different tasks, affect the performance of multi-task FL systems. This opens the door to designing countermeasures specifically aimed at addressing task-induced data heterogeneity in multi-task FL. Similarly, for the extensions of multi-task FL, our method allows researchers to investigate how inducing data heterogeneity from the perspective of one task impacts the performance of other tasks trained concurrently on the same client. Moreover, it facilitates the study of grouping and training local tasks together under different levels of data heterogeneity, potentially leading to new strategies for optimizing task interactions in FL. We leave  exploration of these concepts to future work.
\end{remark}

\begin{remark}[Complexity of Data Partitioning in FL]
Configuring data heterogeneity across FL devices is inherently a combinatorial problem with an enormous solution space. Even in a simplified setting with \(50{,}000\) data points and \(25\) devices, the number of possible data splits is given by \(\mathcal{S}(50{,}000, 25)\), where $\mathcal{S}(.,.)$ denotes the Stirling number of the second kind, which exceeds the estimated number of atoms in the universe (\(\approx 10^{80}\)). To address this, in a nutshell, our method introduces an \textit{ML-based pre-processing phase}, which entails fine-tuning a pre-trained model to extract embeddings for all data points (via one epoch of feedforward), and then clustering the embedded data. This process circumvents the brute-force nature of the combinatorial problem while enabling controlled configuration of data heterogeneity. We show that such clustering-based partitioning can produce client data distributions that impact global model performance, revealing a more severe performance gap than naive data splits.  
Importantly, the overhead of this pre-processing is justifiable as it is conducted entirely offline, before any FL training. Moreover, in high-stakes scenarios (e.g., military or medical domains), investing in such a pre-processing step  enables a more faithful stress-testing of FL methods under data heterogeneous conditions, rather than relying on benchmarks that may not fully capture task-relevant data variability.  
\end{remark}

% Importantly, the overhead of this pre-processing is justifiable as it is conducted entirely offline, before any FL training. Moreover, in high-stakes scenarios (e.g., military or medical domains), investing in such a pre-processing step enables a more faithful stress-testing of FL methods under data heterogeneous conditions, rather than relying on benchmarks that may not fully capture task-relevant data variability.

\vspace{-3mm}

\section{Experiments and Benchmarks} \label{sec:benchmarks}
\noindent In this section, we experiment with various FL methods to give a performance benchmark for each method trained on each vision task under various levels of embedding-based heterogeneity. Our purpose in doing so is three-fold. \textit{(i) Universality of the Phenomenon across FL Methods:} We demonstrate that the phenomenon observed in Sec. \ref{sec:embedding-based-experiments} is universally applied across various FL methods and does not depend on the specific FL method employed. \textit{(ii) Consistency of Performance Impact on Vision Tasks:} We  show the performance drop is consistent across all non-classification vision tasks. \textit{(iii) Benchmark for Future Research:} We provide a benchmark of reference for future endeavors to tackle the task-specific data heterogeneity across non-classification vision tasks in FL. \footnote{All of our implementations and results are contained in the public online GitHub repository of this paper available at\\ \href{https://github.com/KasraBorazjani/task-perspective-het.git}{https://github.com/KasraBorazjani/task-perspective-het.git}.}

\subsection{Data Embedding Extraction}
  In our experiments, we use a pre-trained CLIP~\cite{radford2021learning} model which we fine-tune on the Taskonomy dataset~\cite{taskonomy} for the tasks shown in Fig. \ref{fig:taskonomy}. Then, we extract the embeddings from the fine-tuned model's penultimate layer and cluster them using a K-means algorithm with $K=16$ for better comparison with class-based clusters (note that as we will discuss in Sec.~\ref{sec:app-data-preprocessing}, our dataset has $16$ class labels).
We have also provided the code on how we have fine-tuned the CLIP model in the above open-source GitHub repository which enables easy adaptations to other model architectures and datasets.  
% Nevertheless, given the high variety of choice in model architectures and datasets, it is possible to use different pre-trained models trained on various datasets which are publicly available. 

\begin{figure*}[!h]
% \vspace{-3mm}
    \centering
    \begin{tcolorbox}[colback=FedProxBg, colframe=FedProxFrame, width=0.49\textwidth, left=0pt, right=0pt, top=-2pt, bottom=-2pt, after=\hspace{2mm}, title=2D Edges --- \textbf{FedProx}, halign title=flush center,toptitle=-3pt,
    bottomtitle=-3pt]
    \includegraphics[width=0.48\textwidth]{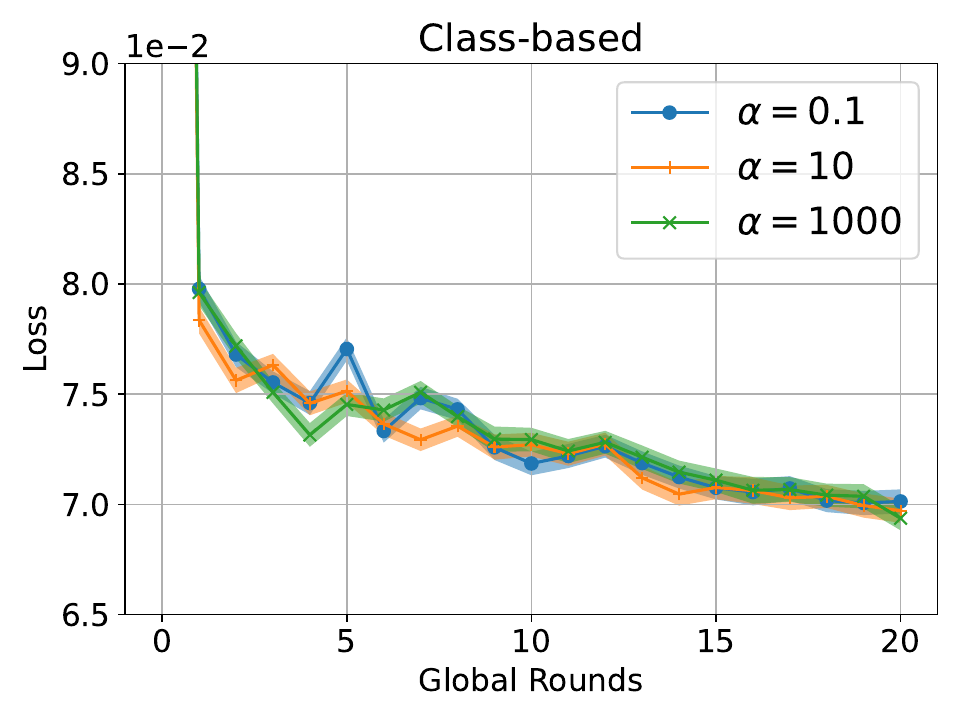}
    \includegraphics[width=0.48\textwidth]{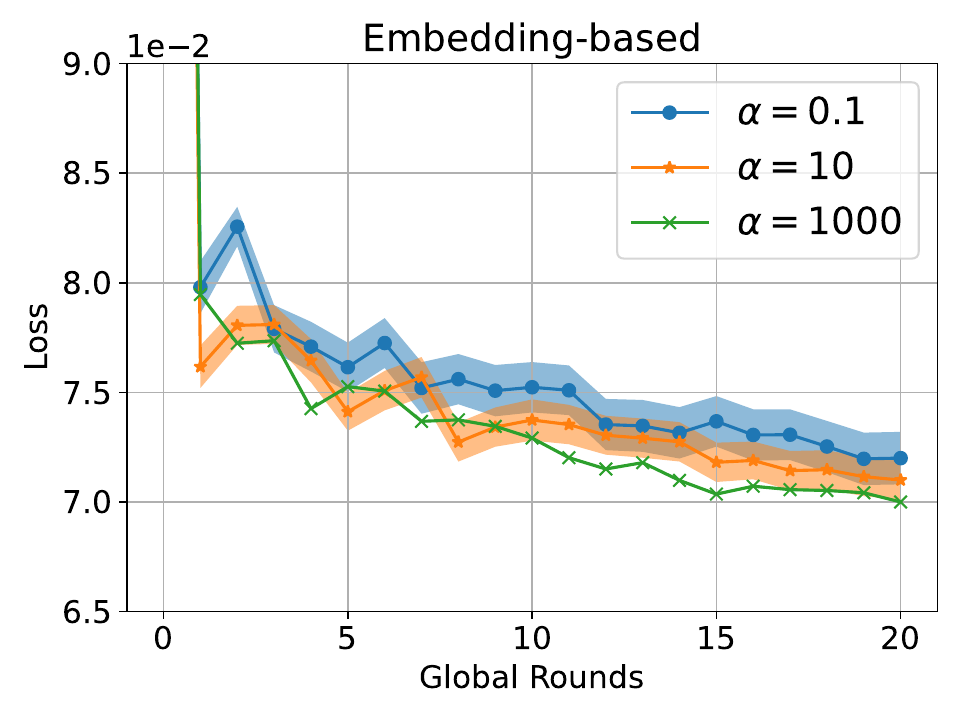}
    \end{tcolorbox}
    \hfill
    \begin{tcolorbox}[colback=FedProxBg, colframe=FedProxFrame, width=0.49\textwidth, left=0pt, right=0pt, before=, top=-2pt, bottom=-2pt, title=Reshading --- \textbf{FedProx}, halign title=flush center,toptitle=-3pt,
    bottomtitle=-3pt]
    \includegraphics[width=0.48\textwidth]{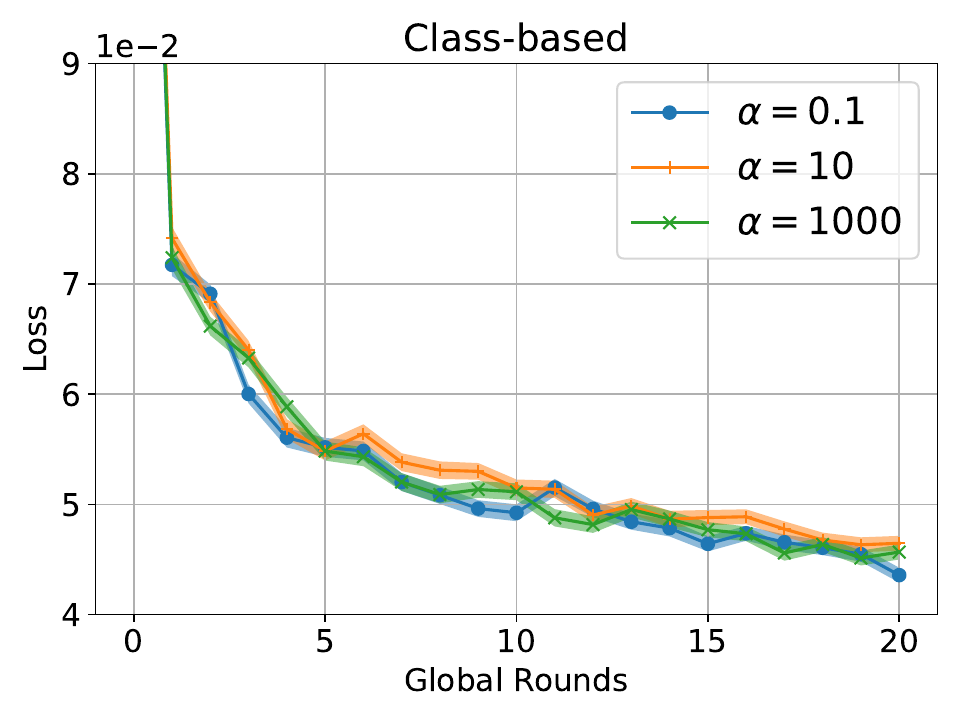}
    \includegraphics[width=0.48\textwidth]{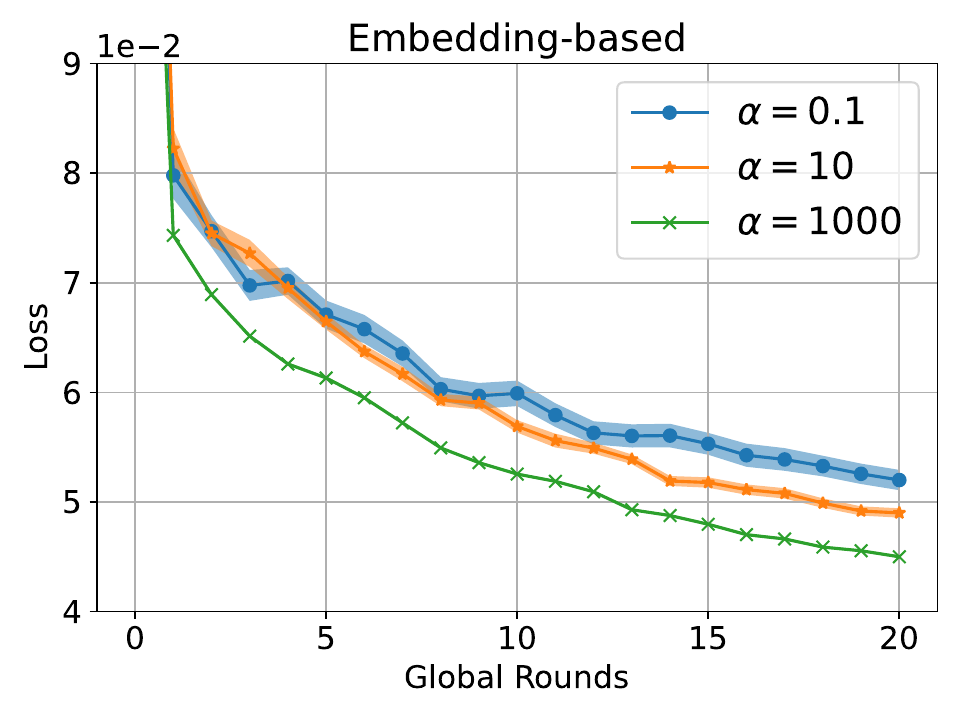}
    \end{tcolorbox}

    \vspace{-3.6mm}
    
    \begin{tcolorbox}[colback=FedProxBg, colframe=FedProxFrame, width=0.49\textwidth, left=0pt, right=0pt, top=-2pt, bottom=-2pt, after=\hspace{2mm}, title=Surface Normals --- \textbf{FedProx}, halign title=flush center,toptitle=-3pt,
    bottomtitle=-3pt]
    \includegraphics[width=0.49\textwidth]{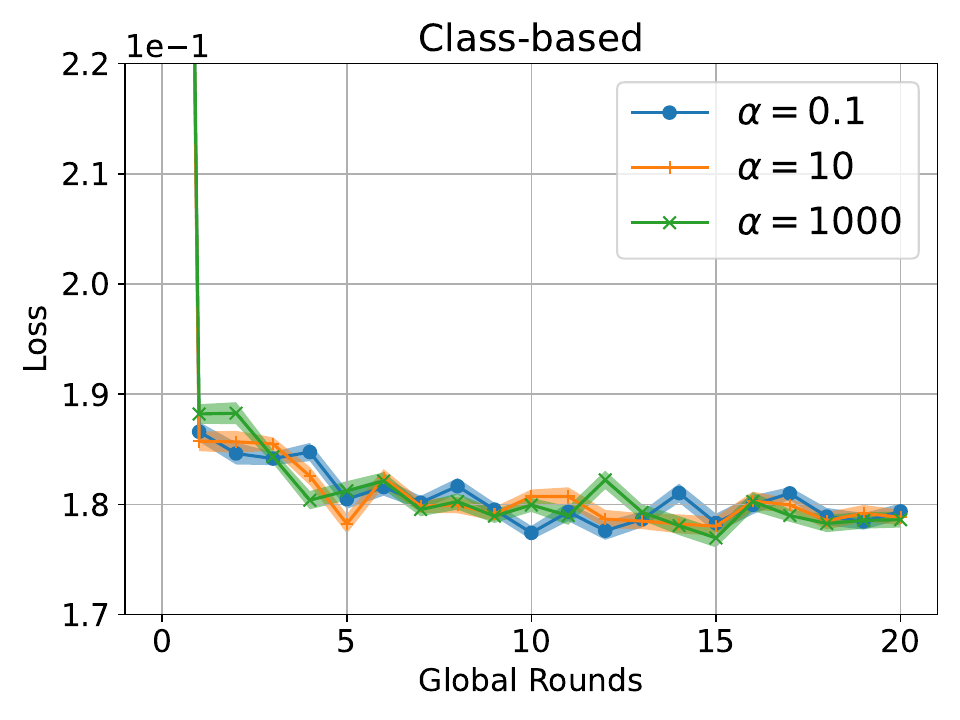}
    \includegraphics[width=0.49\textwidth]{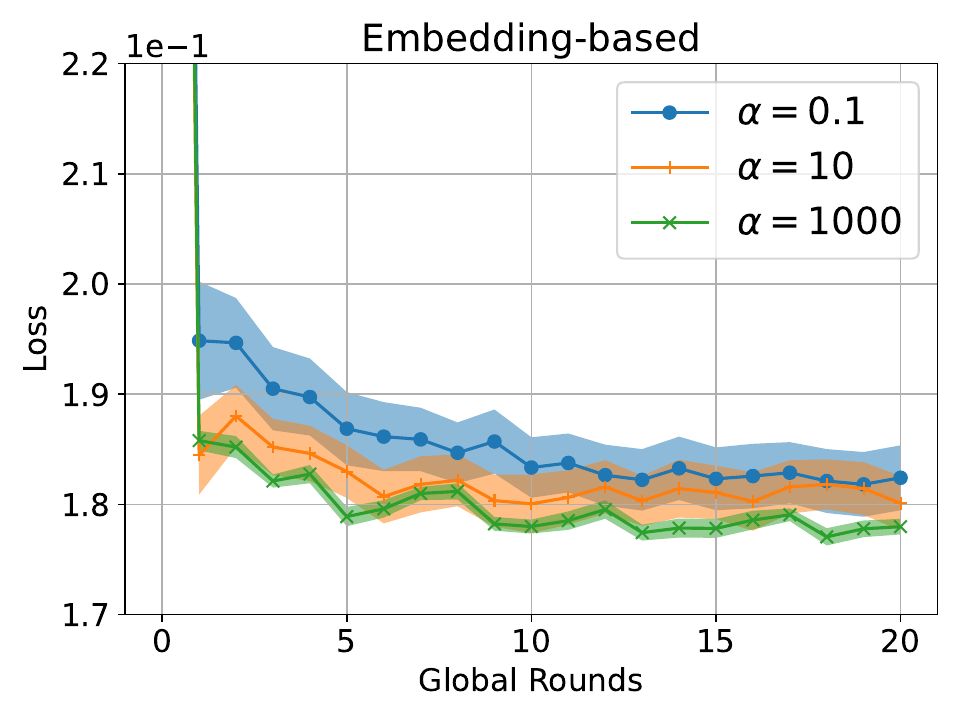}
    \end{tcolorbox}
    \hfill
    \begin{tcolorbox}[colback=FedProxBg, colframe=FedProxFrame, width=0.49\textwidth, left=0pt, right=0pt, top=-2pt, bottom=-2pt, before=, title=Semantic Segmentation --- \textbf{FedProx}, halign title=flush center,toptitle=-3pt,
    bottomtitle=-3pt]
    \includegraphics[width=0.49\textwidth]{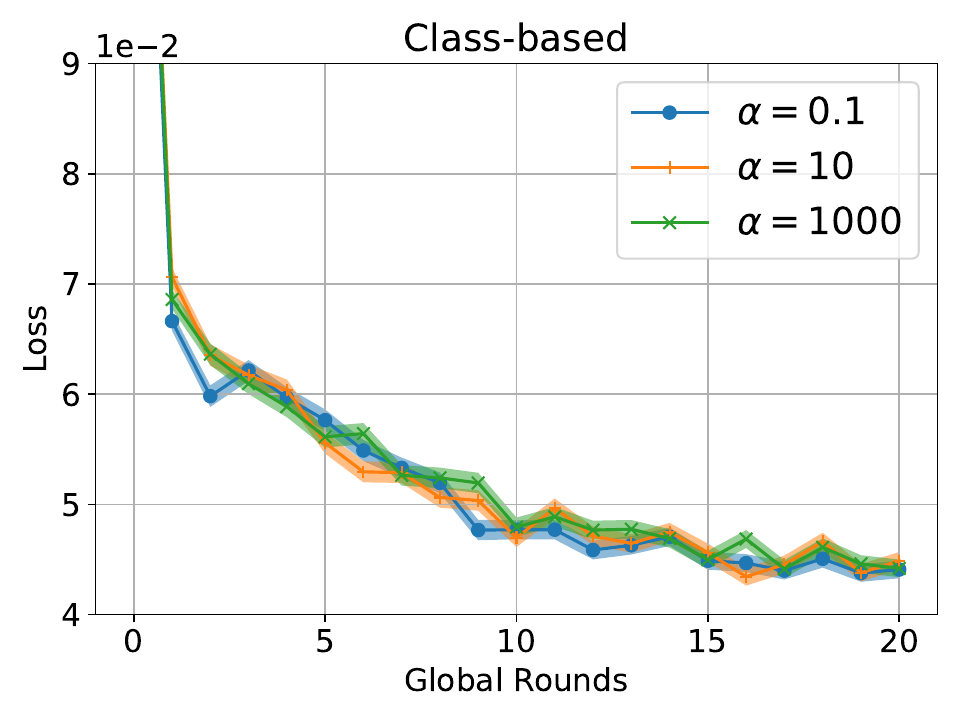}
    \includegraphics[width=0.49\textwidth]{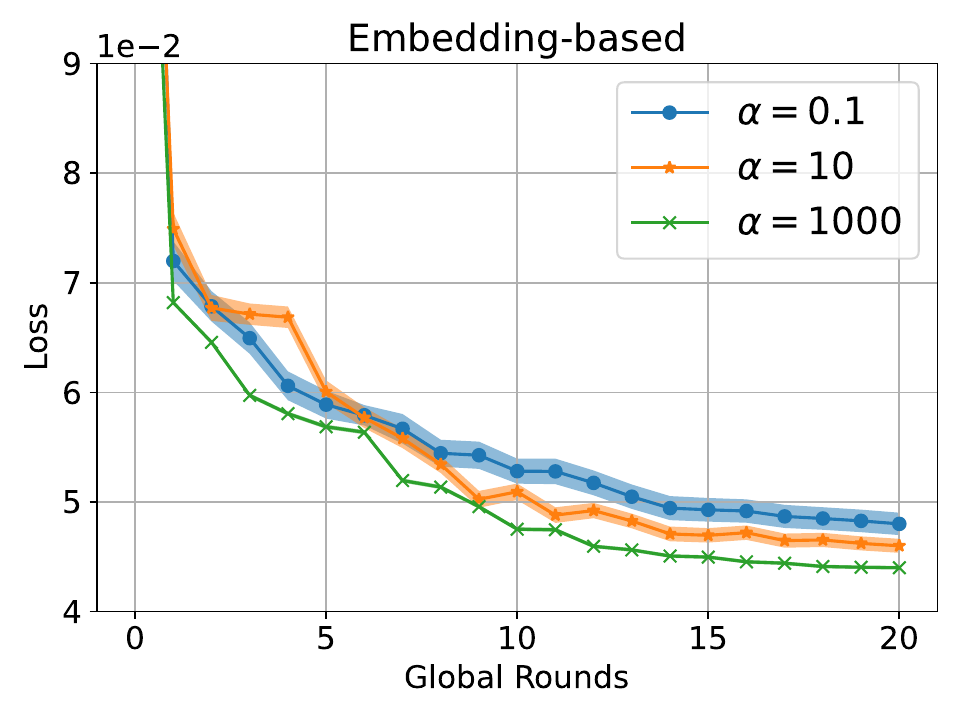}
    \end{tcolorbox}

   \vspace{-3.6mm}
    \begin{tcolorbox}[colback=FedProxBg, colframe=FedProxFrame, width=0.49\textwidth, left=0pt, right=0pt, top=-2pt, bottom=-2pt, after=\hspace{2mm}, title=Euclidean Depth Estimation --- \textbf{FedProx}, halign title=flush center,toptitle=-3pt,
    bottomtitle=-3pt]
    \includegraphics[width=0.49\textwidth]{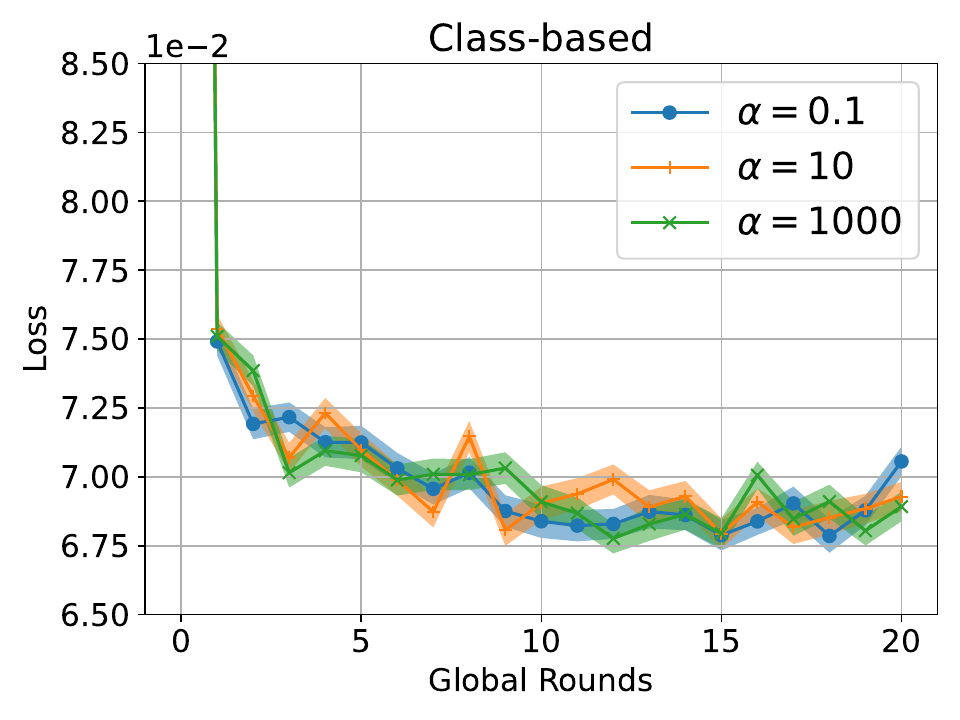}
    \includegraphics[width=0.49\textwidth]{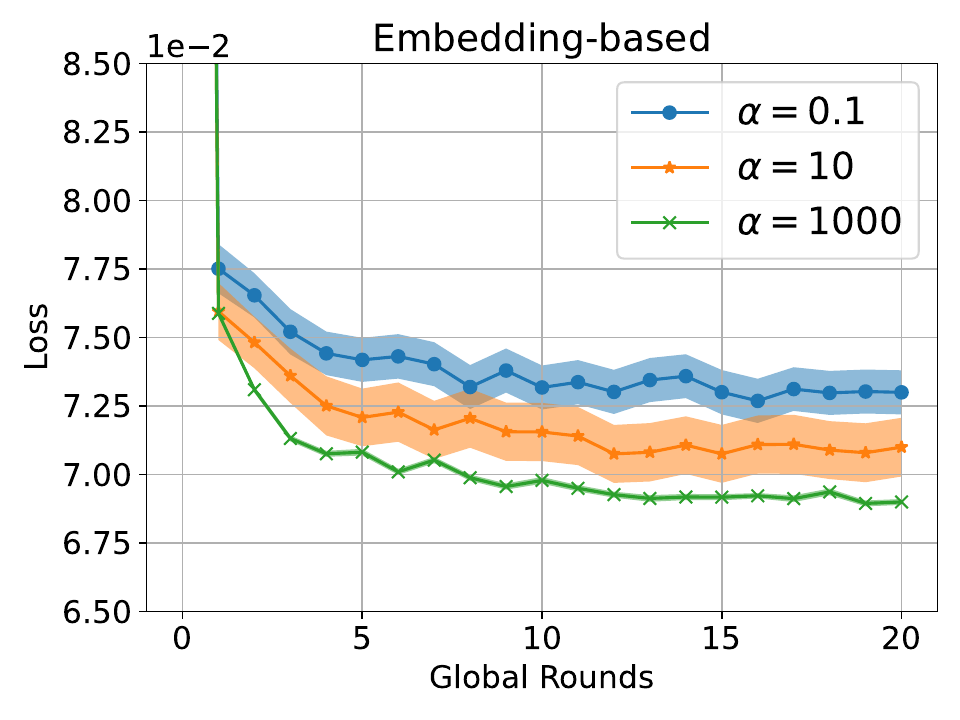}
    \end{tcolorbox}
    \hfill
    \begin{tcolorbox}[colback=FedProxBg, colframe=FedProxFrame, width=0.49\textwidth, left=0pt, right=0pt, top=-2pt, bottom=-2pt, title=3D Keypoints --- \textbf{FedProx}, halign title=flush center, before=, toptitle=-3pt,
    bottomtitle=-3pt]
    \includegraphics[width=0.49\textwidth]{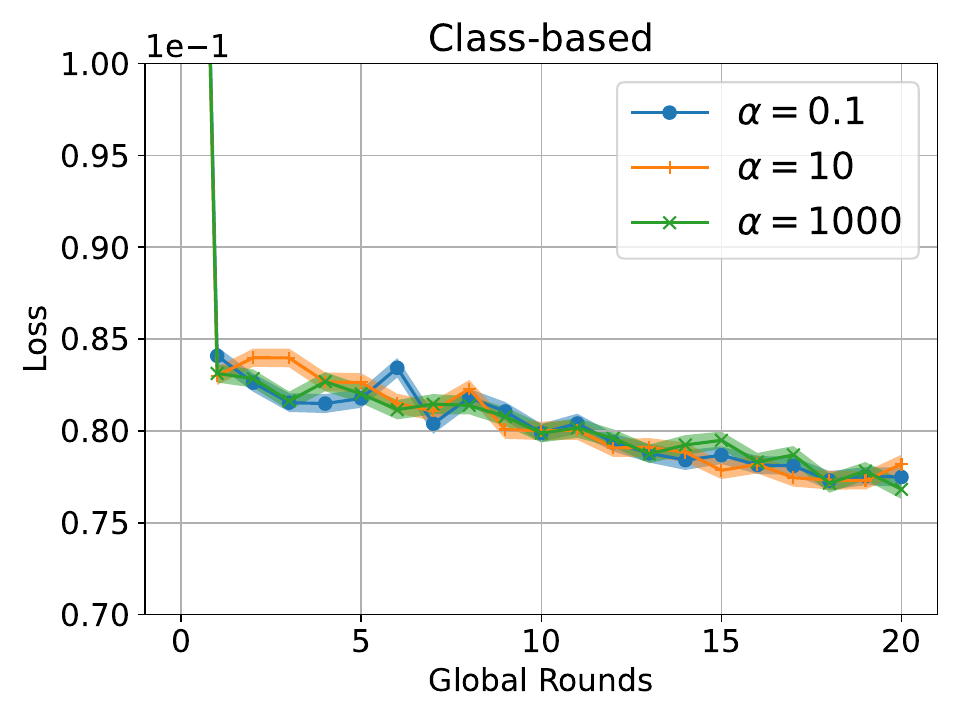}
    \includegraphics[width=0.49\textwidth]{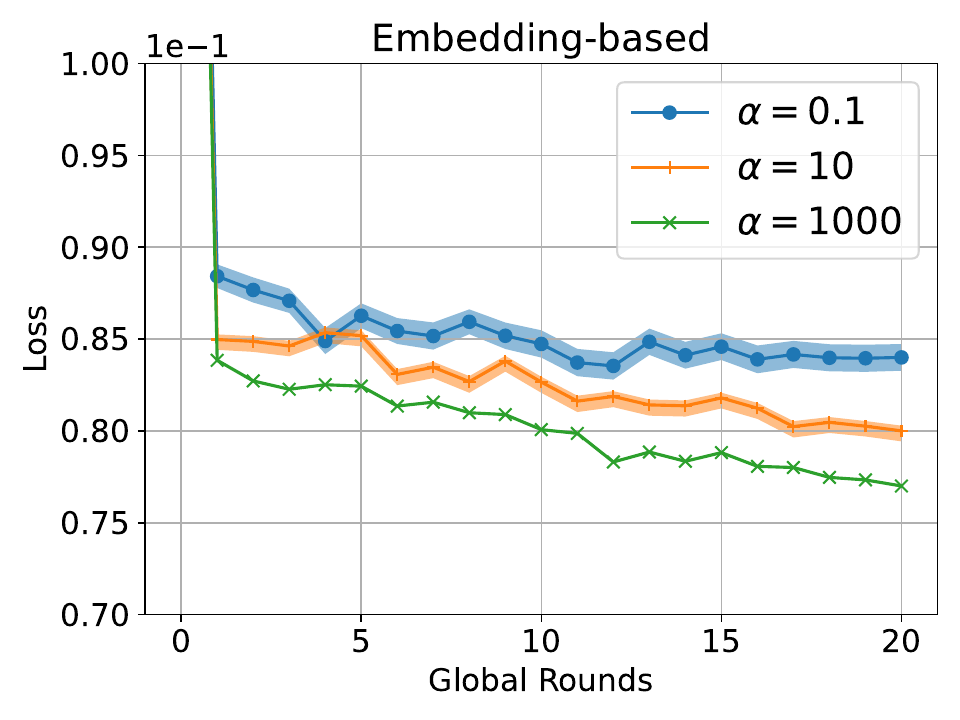}
    \end{tcolorbox}

   \vspace{-3.6mm}
    \begin{tcolorbox}[colback=FedProxBg, colframe=FedProxFrame, width=0.49\textwidth, left=0pt, right=0pt, top=-2pt, bottom=-2pt, after=\hspace{2mm}, title=Scene Classification --- \textbf{FedProx}, halign title=flush center,toptitle=-3pt,
    bottomtitle=-3pt]
    \includegraphics[width=0.49\textwidth]{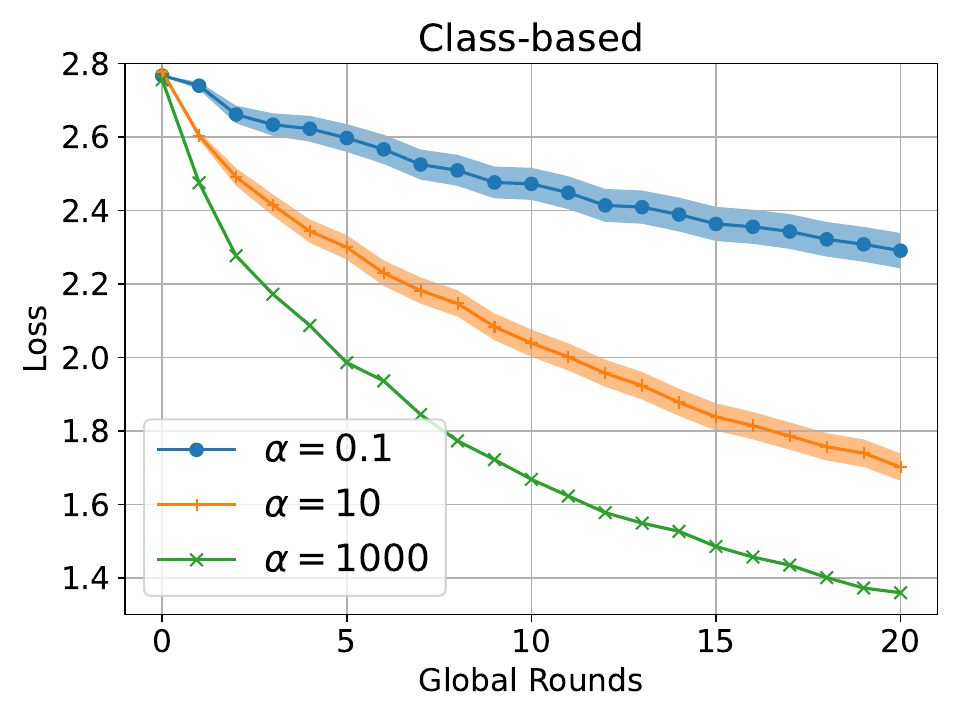}
    \includegraphics[width=0.49\textwidth]{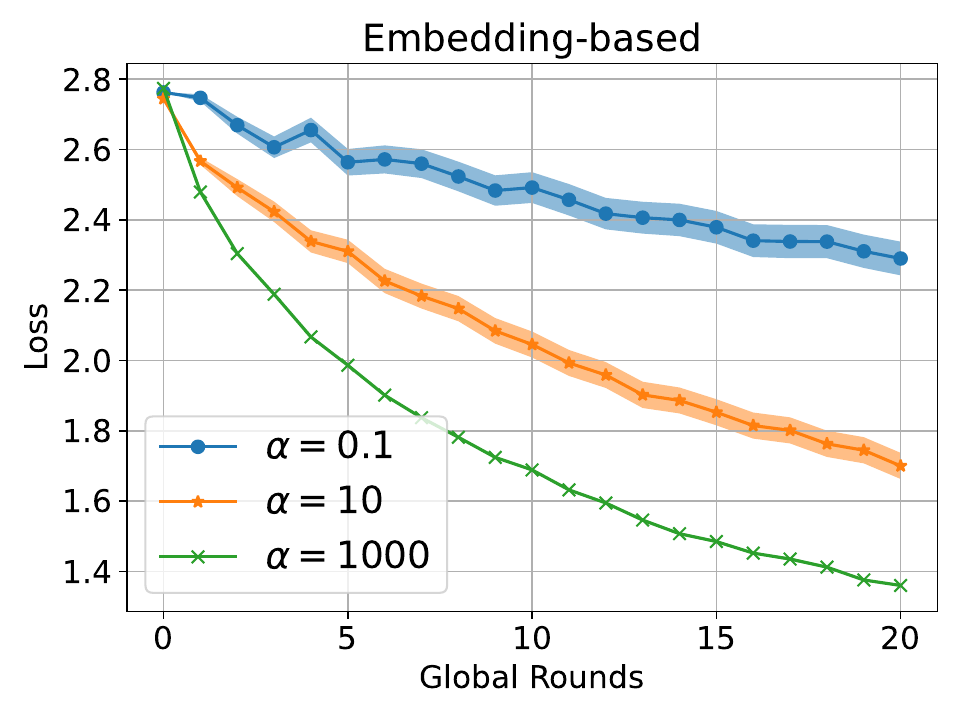}
    \end{tcolorbox}
    
    \caption{\textbf{Class-based vs. Embedding-based distribution:} Comparison of how performing Dirichlet distribution over the datapoints' labels (equivalently, the scene class feature) and the extracted embeddings affect the performance in FL when the FedProx technique is used.  Solid lines show the mean loss across clients; the shaded region denotes client-level dispersion (standard deviation of the loss across clients) at each global round. The same phenomena observed in Fig.~\ref{fig:embedding-vs-class-p1} can be observed in the FedProx results.}
    \label{fig:fedprox-results}
    \vspace{-5mm}
\end{figure*}
\subsection{Data Preprocessing Details} \label{sec:app-data-preprocessing}
\noindent The data gathered in the Taskonomy dataset, pertaining to the scene classification label, is provided in the form of a probability vector $\bm{\beta}_d = \{\beta_{d,1}, \beta_{d,2}, \cdots, \beta_{d,J}\}$ for each datapoint $d$. Each $\beta_{d,j}, 1 \leq j \leq J$ represents the probability of the datapoint $d$ belonging to the scene class $j$ in a set of $J = 365$ available classes. To simplify the process of distributing the data, we first take the $\max$ across the elements of each $\bm{\beta}_d$ for each datapoint and assign it as the class label to that datapoint. However, not all of the $J$ classes have enough datapoints to be distributed across the simulated clients in our experiments: some classes have small numbers of datapoints which may fall as low as $1$ datapoint. Thus, distributing the datapoints belonging to these ill-represented classes across the clients and including them in the classification task may lead to only some clients (as low as one) having data from these classes. This phenomenon will keep us from ever having a homogeneous data distribution across the clients which is one of the settings we have experimented with throughout the paper. Another problem is that training the global model in a federated fashion when one class is very poor in datapoint count will lead to an imbalanced training, where the biased classifying ability towards the ill-represented class will automatically worsen the results, rendering them unreliable for comparison across different hyperparameter values such as the Dirichlet distribution's concentration parameter $\alpha$. To overcome these obstacles, we take the top 16 most-populated classes into consideration in all of our experiments and distribute their data across the clients.
% We use the same set of data for all tasks  to ensure no effect from the less populated classes is induced into the results of the experiments in any of the tasks.

\subsection{Dataset Partitioning and Training Protocol}
  In the case of the Taskonomy dataset, for the centralized pre-training of each task, we use the initial three scenes (``allensville," ``beechwood," and ``benevolence") summing up to an initial count of $14246$ datapoints available.  To track the performance of the centralized model, which is used for embedding extraction, for hyperparameter tuning and early stopping purposes, we use 100 datapoints chosen uniformly at random from the dataset as validation data.  When performing the federated training, we distribute the data across the clients based on the underlying data partitioning method (i.e., embedding- and class-based), where $10\%$ of local data points are used for validation.  The numbers and statistics reported in the manuscript are based on this validation set.  For global model validation, we report the average loss of the global model across the clients on their validation data. For the centralized training, we use the early stopping method to halt training if no improvement in loss was observed after $5$ epochs. For the federated training, we train all methods for $20$ global aggregations, each comprising $100$ SGD rounds with the mini-batch size of $16$. In addition, we sample all $25$ clients during each global aggregation round as our method is not focused on client sampling. Unless stated otherwise, we fix random seeds for the FL training. To quantify uncertainty, we vary the K-means clustering seed and report mean$\pm$std across seeds in Sec.~\ref{subsubsec:fedavg-seed}. For a summary of simulation hyperparameters refer to Table IV in Appendix~E.

\begin{figure*}[!h]
% \vspace{-3mm}
    \centering
    \begin{tcolorbox}[colback=BoxBg, colframe=BoxFrame, width=0.49\textwidth, left=0pt, right=0pt, top=-2pt, bottom=-2pt, after=\hspace{2mm}, title=2D Edges --- \textbf{SCAFFOLD}, halign title=flush center,toptitle=-3pt,
    bottomtitle=-3pt]
    \includegraphics[width=0.48\textwidth]{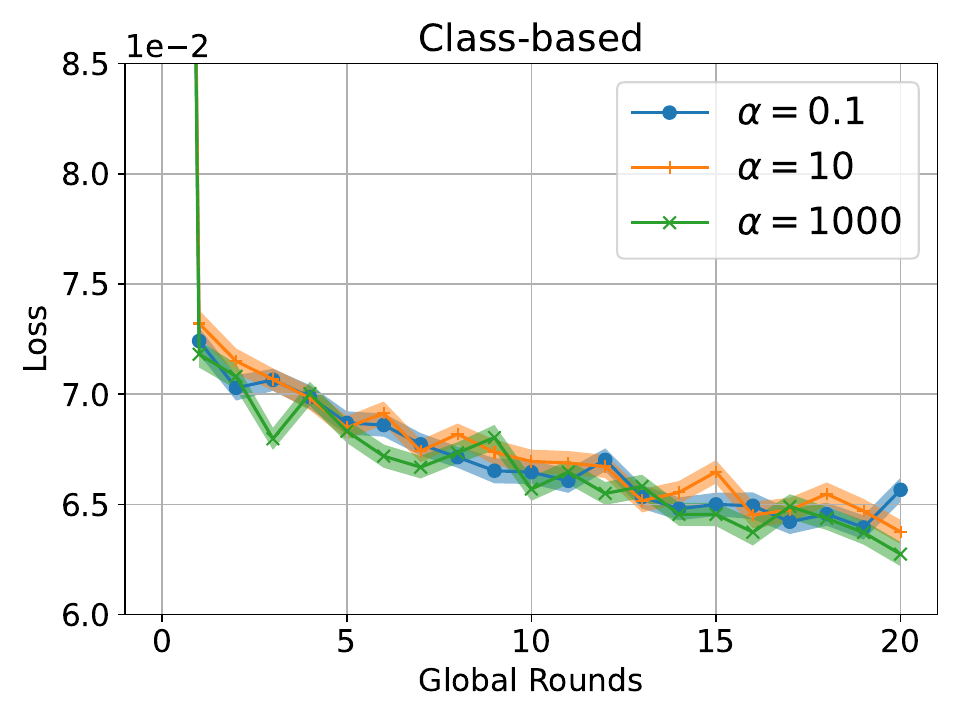}
    \includegraphics[width=0.48\textwidth]{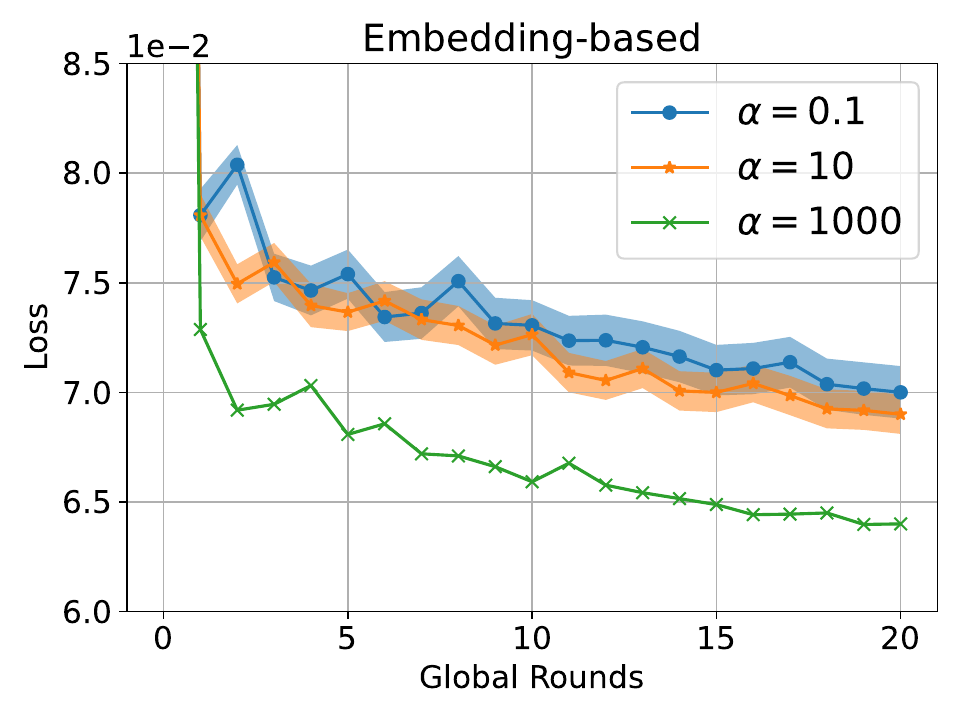}
    \end{tcolorbox}
    \hfill
    \begin{tcolorbox}[colback=BoxBg, colframe=BoxFrame, width=0.49\textwidth, left=0pt, right=0pt, top=-2pt, bottom=-2pt, before=, title=Reshading --- \textbf{SCAFFOLD}, halign title=flush center,toptitle=-3pt,
    bottomtitle=-3pt]
    \includegraphics[width=0.48\textwidth]{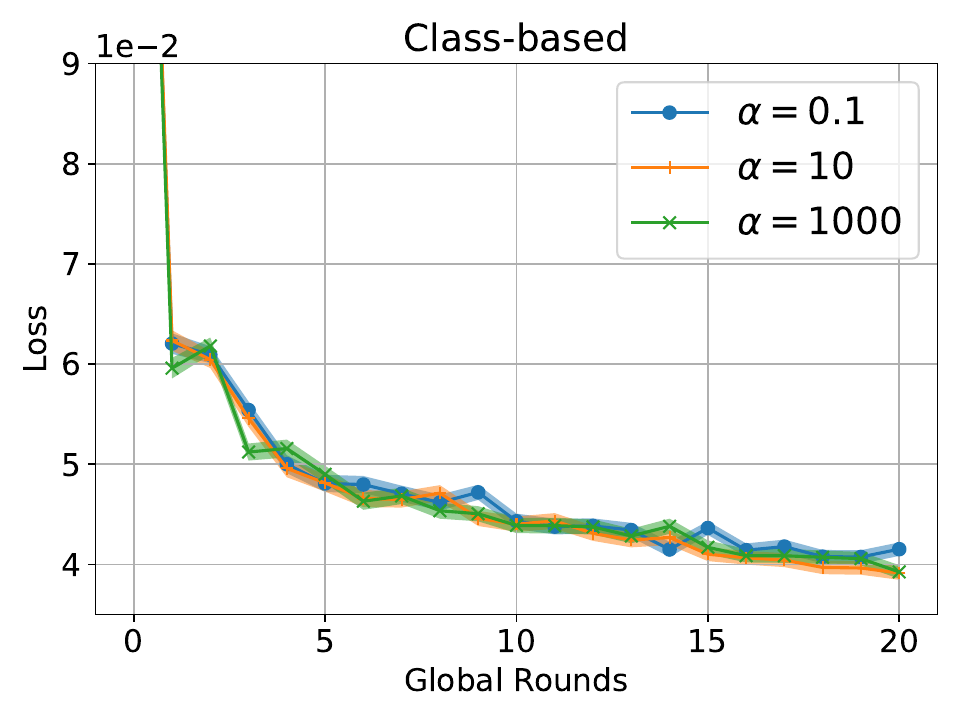}
    \includegraphics[width=0.48\textwidth]{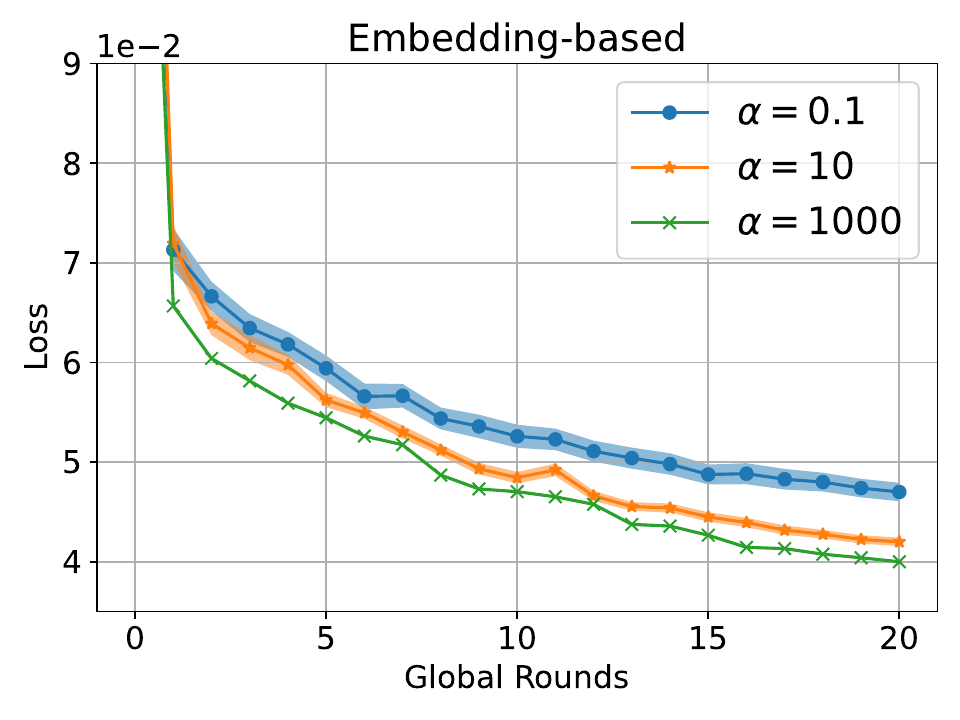}
    \end{tcolorbox}

     \vspace{-3.6mm}
    \begin{tcolorbox}[colback=BoxBg, colframe=BoxFrame, width=0.49\textwidth, left=0pt, right=0pt, top=-2pt, bottom=-2pt, after=\hspace{2mm}, title=Surface Normals --- \textbf{SCAFFOLD}, halign title=flush center,toptitle=-3pt,
    bottomtitle=-2pt]
    \includegraphics[width=0.49\textwidth]{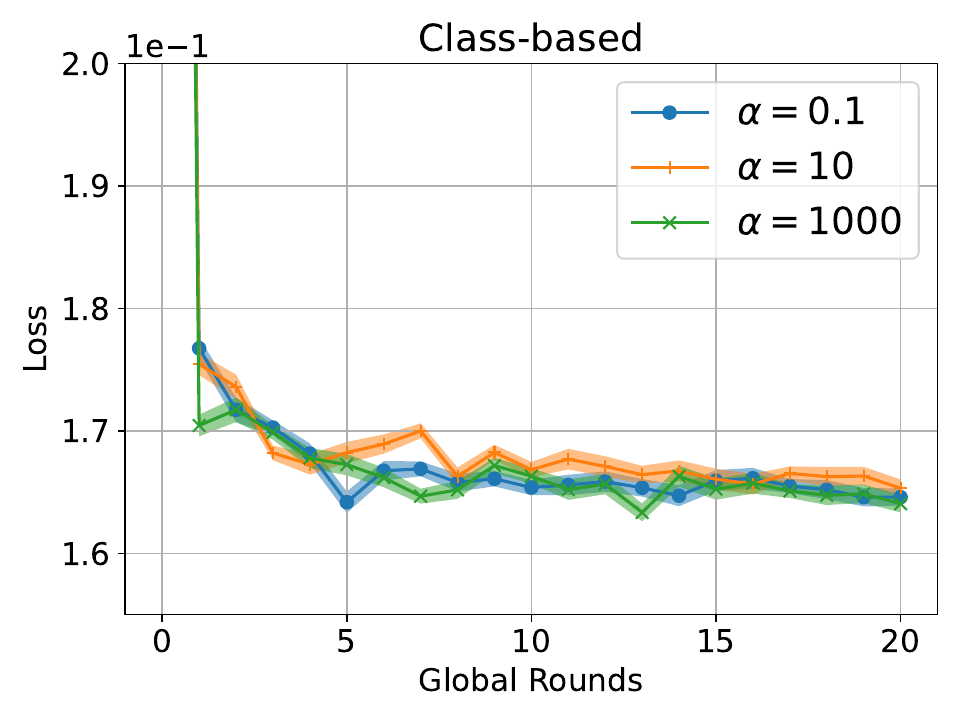}
    \includegraphics[width=0.49\textwidth]{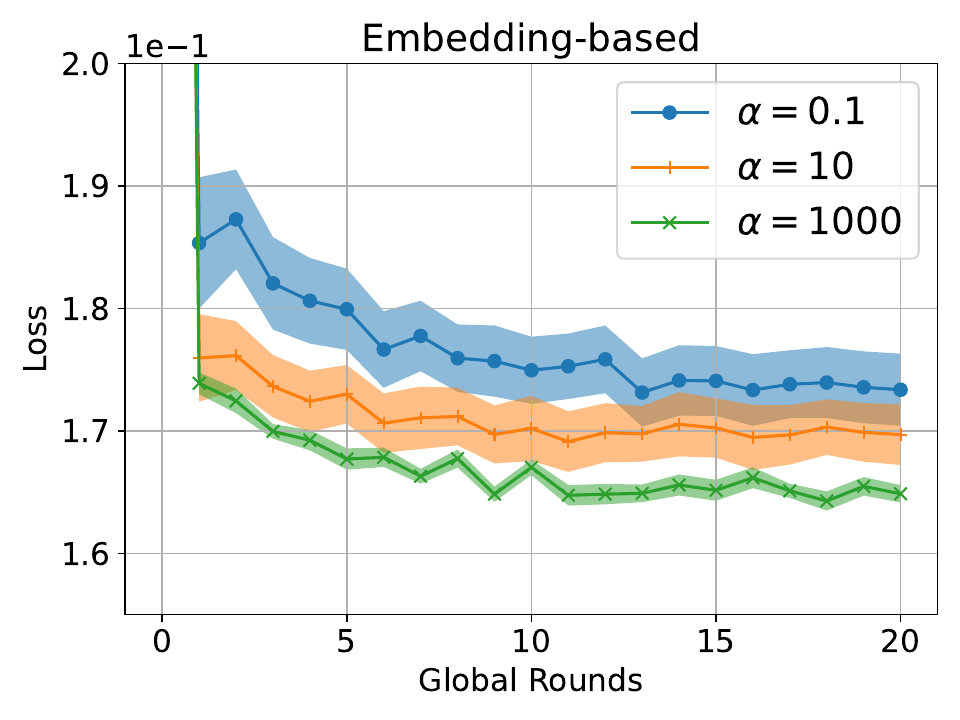}
    \end{tcolorbox}
    \hfill
    \begin{tcolorbox}[colback=BoxBg, colframe=BoxFrame, width=0.49\textwidth, left=0pt, right=0pt, top=-2pt, bottom=-2pt, before=, title=Semantic Segmentation --- \textbf{SCAFFOLD}, halign title=flush center,toptitle=-3pt,
    bottomtitle=-3pt]
    \includegraphics[width=0.49\textwidth]{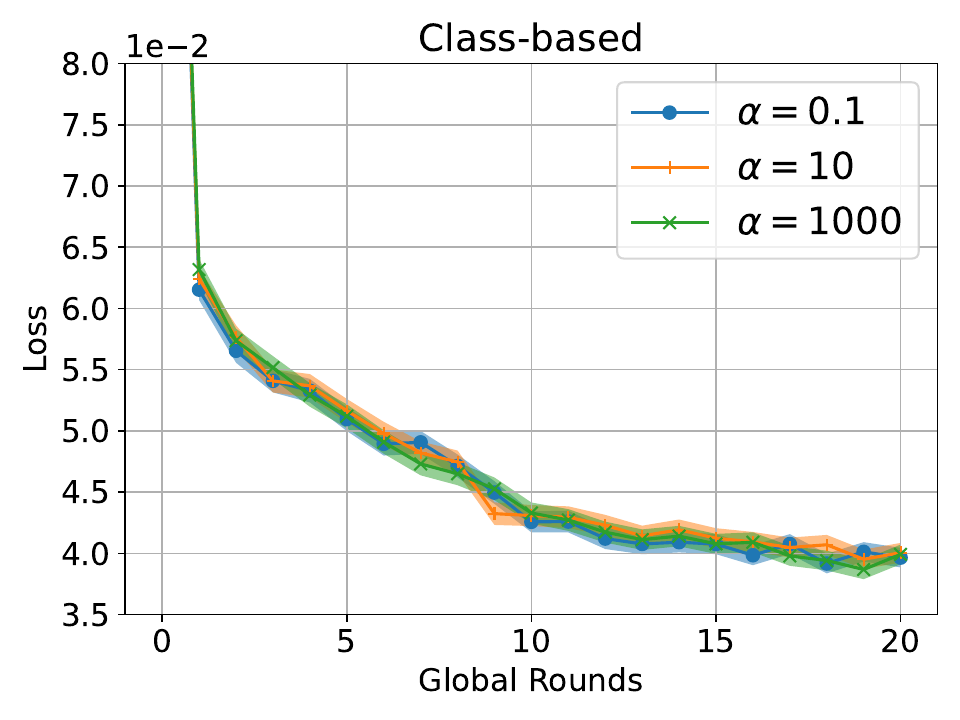}
    \includegraphics[width=0.49\textwidth]{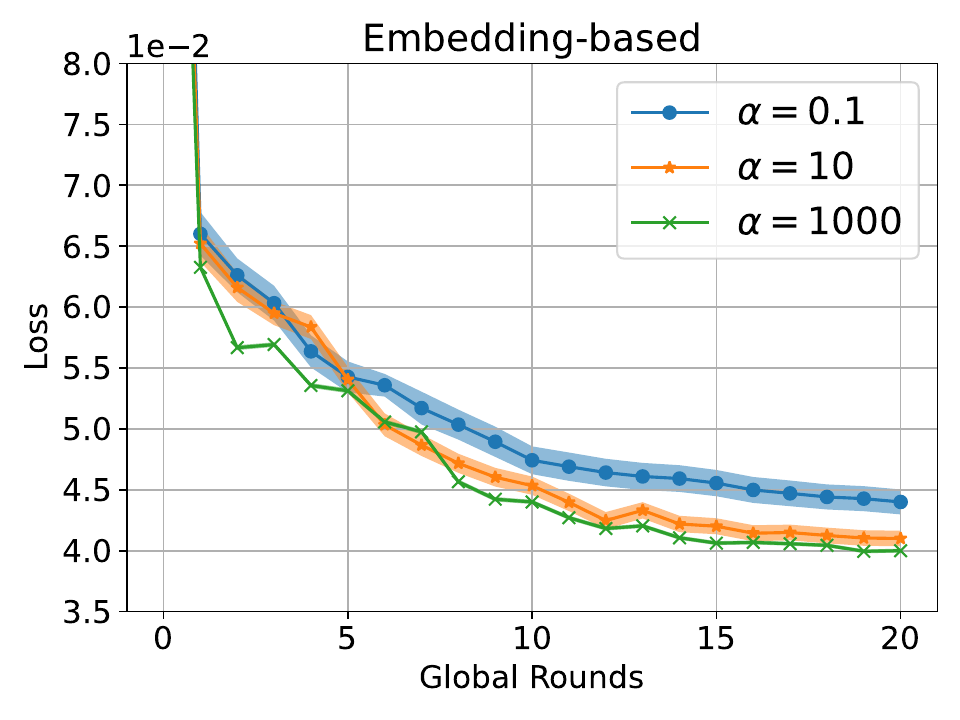}
    \end{tcolorbox}

\vspace{-3.6mm}
    \begin{tcolorbox}[colback=BoxBg, colframe=BoxFrame, width=0.49\textwidth, left=0pt, right=0pt, top=-2pt, bottom=-2pt, after=\hspace{2mm}, title=Euclidean Depth Estimation --- \textbf{SCAFFOLD}, halign title=flush center,toptitle=-3pt,
    bottomtitle=-3pt]
    \includegraphics[width=0.49\textwidth]{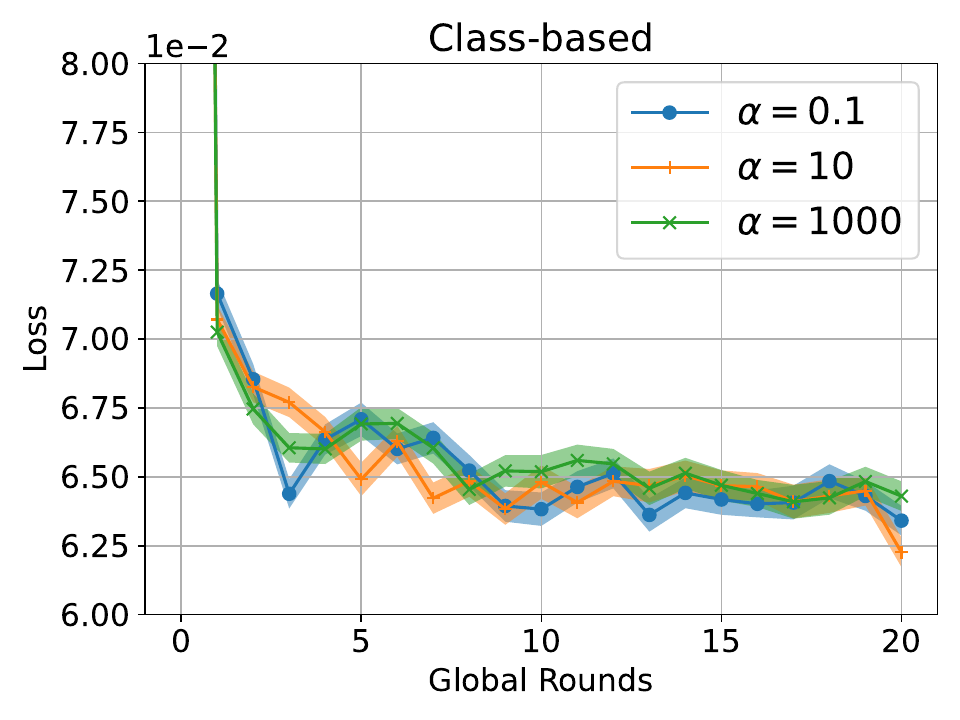}
    \includegraphics[width=0.49\textwidth]{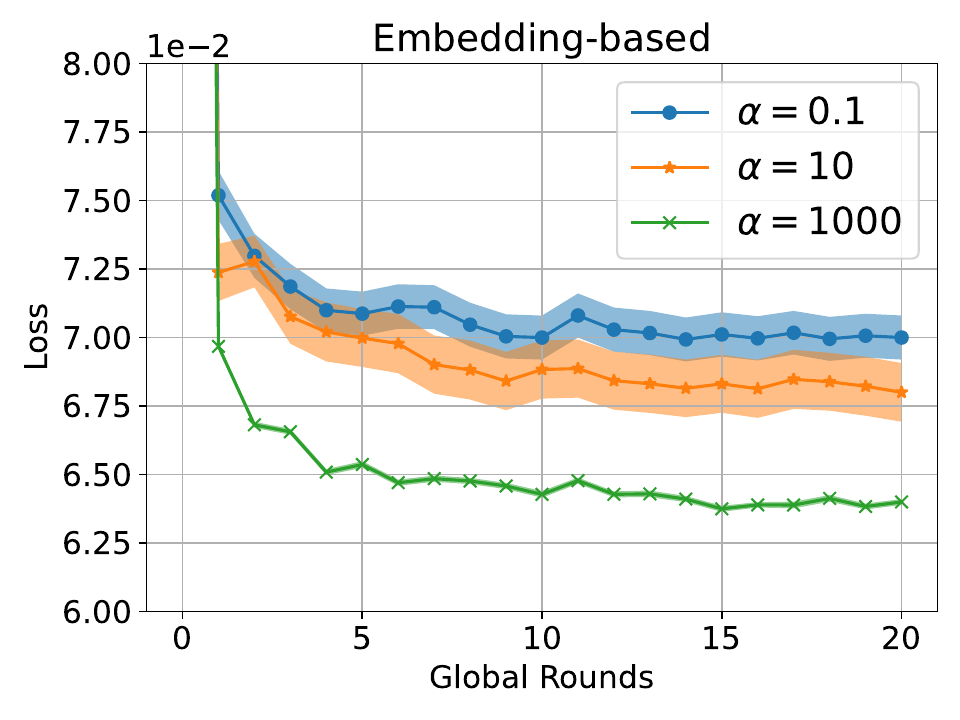}
    \end{tcolorbox}
    \hfill
    \begin{tcolorbox}[colback=BoxBg, colframe=BoxFrame, width=0.49\textwidth, left=0pt, right=0pt, top=-2pt, bottom=-2pt, title=3D Keypoints --- \textbf{SCAFFOLD}, halign title=flush center, before= ,toptitle=-3pt,
    bottomtitle=-3pt]
    \includegraphics[width=0.49\textwidth]{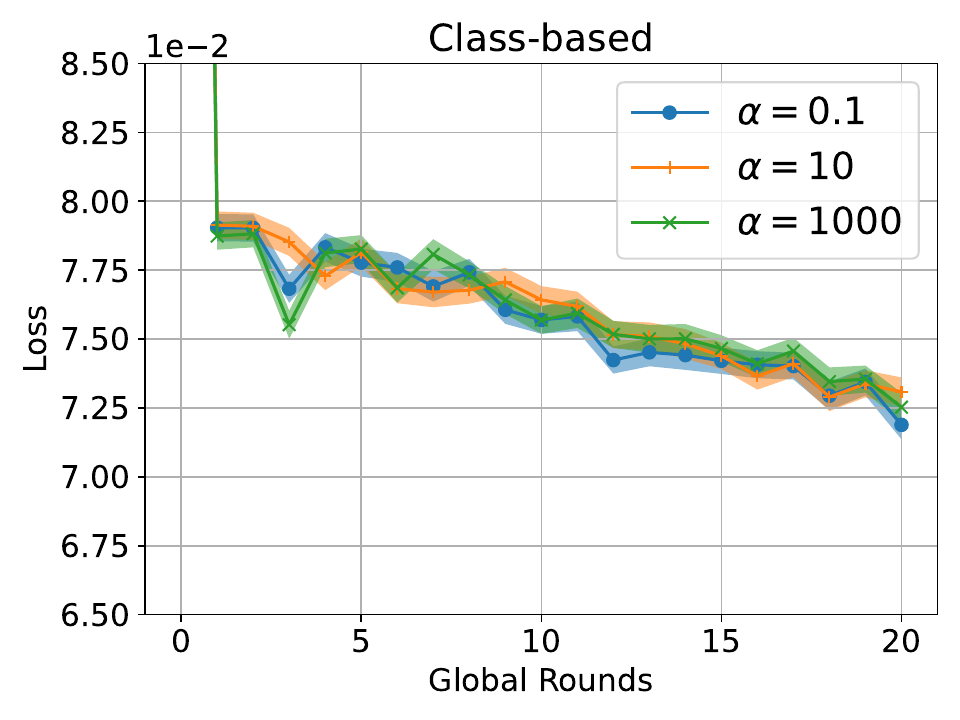}
    \includegraphics[width=0.49\textwidth]{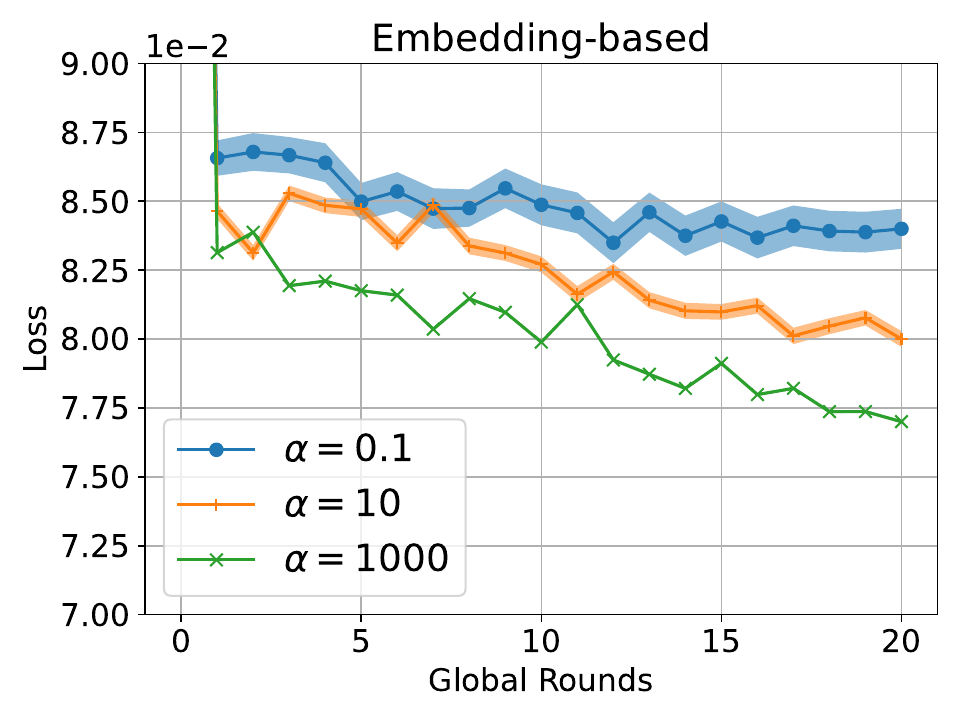}
    \end{tcolorbox}

    \vspace{-3.6mm}
    \begin{tcolorbox}[colback=BoxBg, colframe=BoxFrame, width=0.49\textwidth, left=0pt, right=0pt, top=-2pt, bottom=-2pt, after=\hspace{2mm}, title=Scene Classification --- \textbf{SCAFFOLD}, halign title=flush center,toptitle=-3pt,
    bottomtitle=-3pt]
    \includegraphics[width=0.49\textwidth]{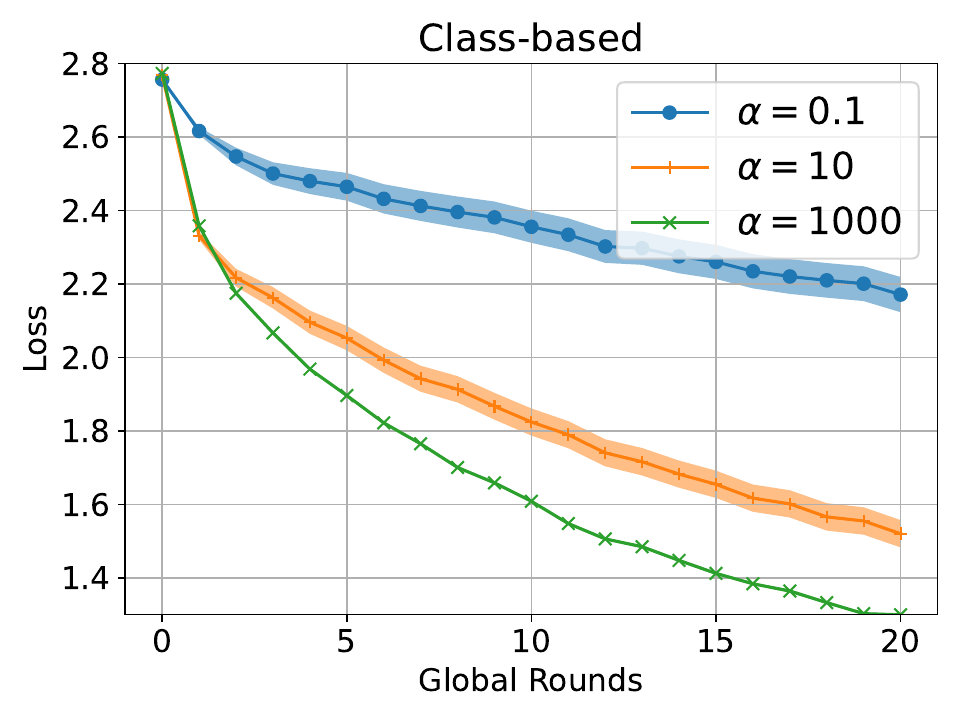}
    \includegraphics[width=0.49\textwidth]{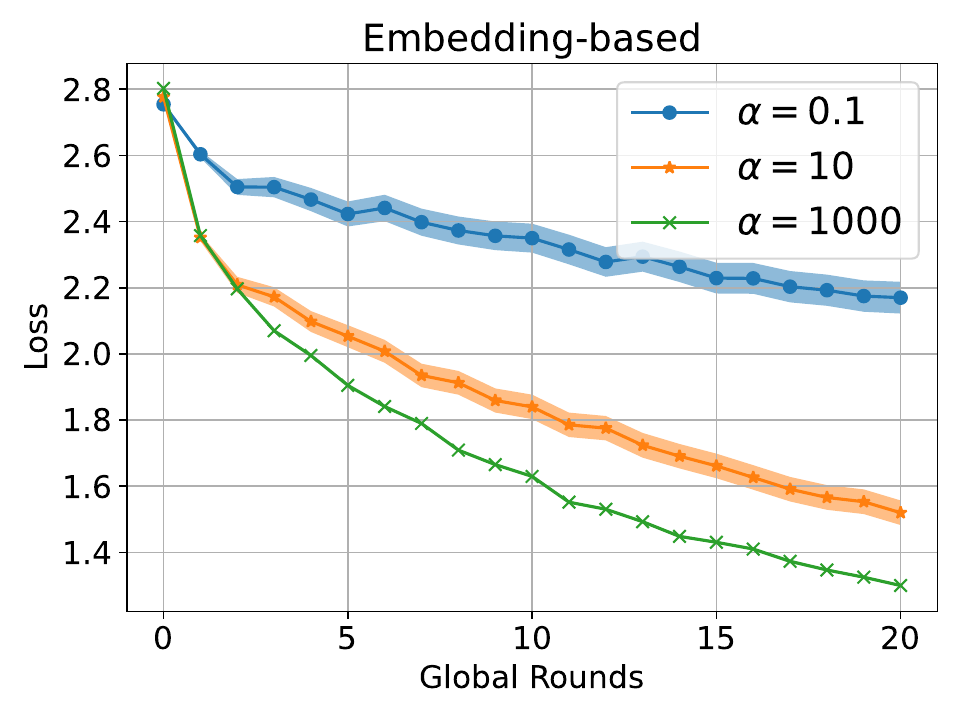}
    \end{tcolorbox}

    \caption{\textbf{Class-based vs. Embedding-based distribution:} Comparison of how performing Dirichlet distribution over the datapoints' labels (equivalently, the scene class feature) and the extracted embeddings affect the performance in FL when the SCAFFOLD technique is used.  Solid lines show the mean loss across clients; the shaded region denotes client-level dispersion (standard deviation of the loss across clients) at each global round.  The same phenomena observed in Fig.~\ref{fig:embedding-vs-class-p1} can be observed in the SCAFFOLD results.}
    \label{fig:scaffold_results}
    \vspace{-5mm}
\end{figure*}

\vspace{-3mm}
\subsection{FL Methods} \label{subsec:benchmark-methods}
\vspace{-.5mm}
To  observe the effect of embedding-based heterogeneity across various FL methods, we have considered the following six variations of FL:

\begin{enumerate}[leftmargin=4.75mm]
    \item \textbf{FedAvg}\cite{fed-avg}: One of the widely used methods in FL with the dynamics explained in Sec. \ref{sec:fed-sys-model}. The  results presented in Fig. \ref{fig:embedding-vs-class-p1} were also gathered by employing this method.
    \item \textbf{FedProx}\cite{fedprox}: FedProx extends FedAvg by incorporating a proximal term into the local objective function to address data heterogeneity across clients. This term penalizes each client's local model divergence from the global model, encouraging consistency and stability during aggregation.
    \item \textbf{SCAFFOLD}\cite{scaffold}: To address heterogeneous data distributions, SCAFFOLD maintains control variates, both on the server and clients, to adjust the local training updates and align them more closely with the global optimization direction. Clients send their updated control variates along with model parameters to the server. The server averages these control variates and redistributes them to clients in subsequent rounds.
    % \item \textbf{Ditto}\cite{ditto}: Each client in Ditto maintains its own personalized model while contributing to a shared global model. The interaction between clients and the server involves updating the global model using averaged gradients, while clients locally fine-tune their models to suit their specific data distributions. This  approach is meant to mitigate the performance drop caused by data heterogeneity.
    \item \textbf{FedRep}\cite{fedrep}: FedRep decouples the optimization of shared and personalized components. Clients optimize a shared representation (encoder) collaboratively while maintaining a personalized classifier layer for task-specific outputs. During interaction with the server, only the shared representation is aggregated, leaving the personalized components untouched. This decoupling aims to minimize the negative impact of data heterogeneity while allowing personalization.
    \item \textbf{FedAmp}\cite{fedamp}: FedAmp employs a model aggregation strategy that incorporates weighted averaging based on client-specific similarity measures. Clients send their local model updates and similarity scores to the server. The server aggregates these updates, prioritizing contributions from clients with higher similarity to the global objective. The method introduces an adaptive aggregation mechanism that accounts for the variability in clients' data distributions.
\end{enumerate}

\begin{figure*}[!h]
    \centering
    \begin{tcolorbox}[colback=BoxBg, colframe=BoxFrame, width=0.49\textwidth, left=0pt, right=0pt, top=-2pt, bottom=-2pt, after=\hspace{2mm}, title=2D Edges --- \textbf{FedRep}, halign title=flush center,toptitle=-3pt,
    bottomtitle=-3pt]
    \includegraphics[width=0.48\textwidth]{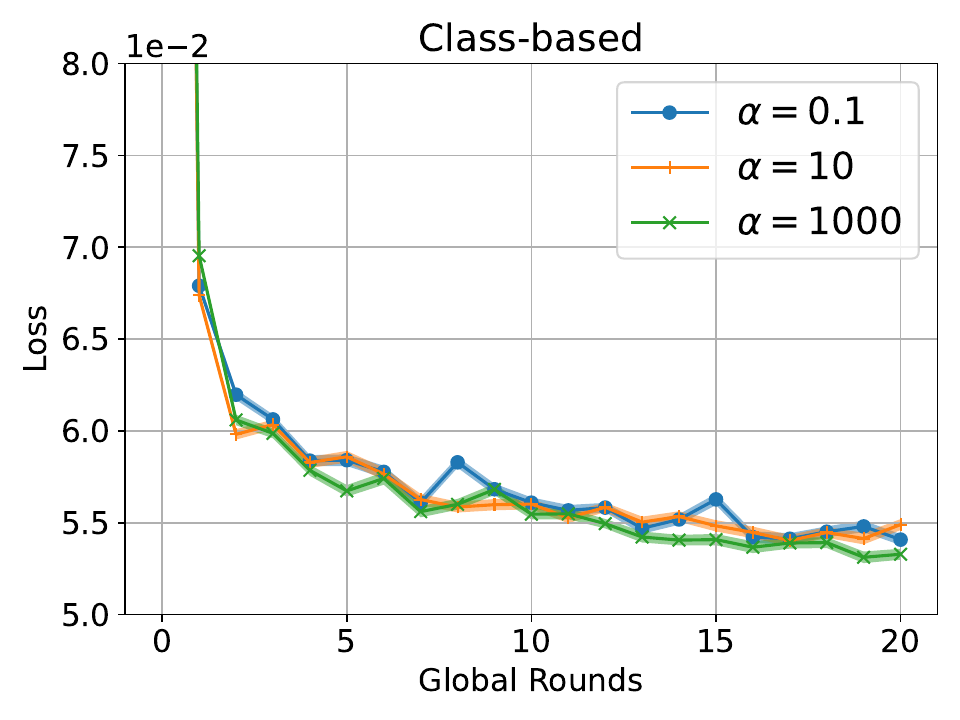}
    \includegraphics[width=0.48\textwidth]{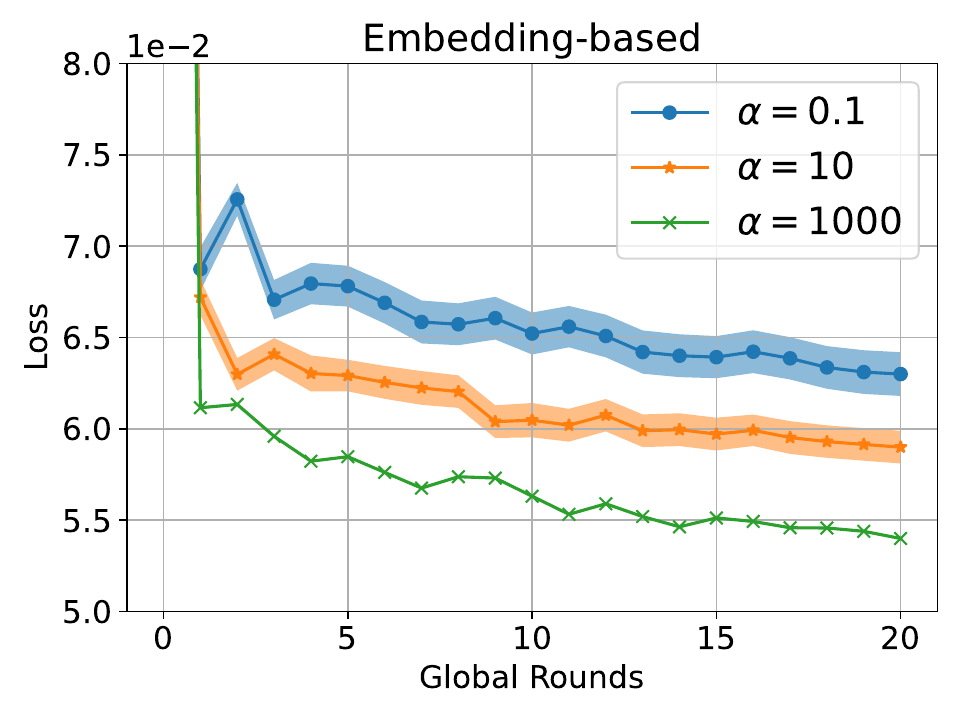}
    \end{tcolorbox}
    \hfill
    \begin{tcolorbox}[colback=BoxBg, colframe=BoxFrame, width=0.49\textwidth, left=0pt, right=0pt, top=-2pt, bottom=-2pt, before=, title=Reshading --- \textbf{FedRep}, halign title=flush center,toptitle=-3pt,
    bottomtitle=-3pt]
    \includegraphics[width=0.48\textwidth]{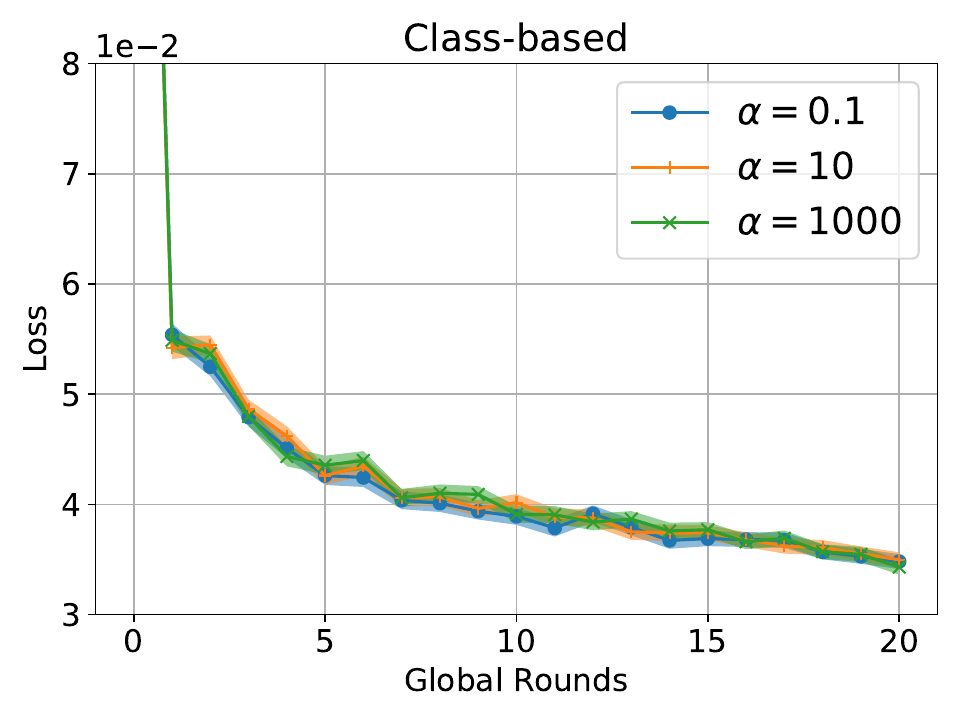}
    \includegraphics[width=0.48\textwidth]{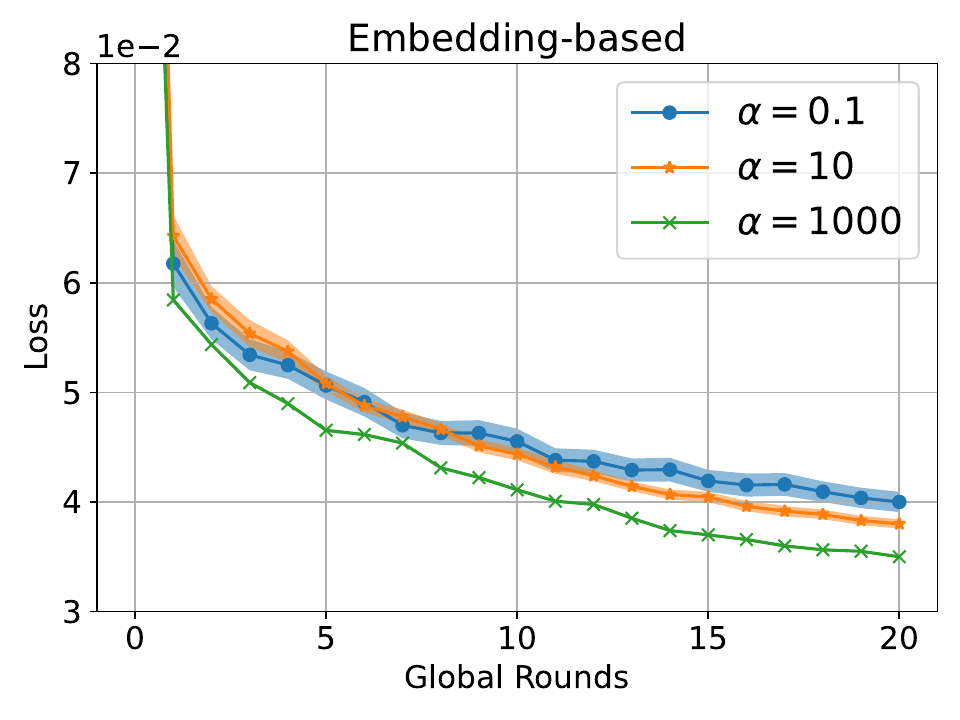}
    \end{tcolorbox}

    \vspace{-3.6mm}
    \begin{tcolorbox}[colback=BoxBg, colframe=BoxFrame, width=0.49\textwidth, left=0pt, right=0pt, top=-2pt, bottom=-2pt, after=\hspace{2mm}, title=Surface Normals --- \textbf{FedRep}, halign title=flush center,toptitle=-3pt,
    bottomtitle=-3pt]
    \includegraphics[width=0.49\textwidth]{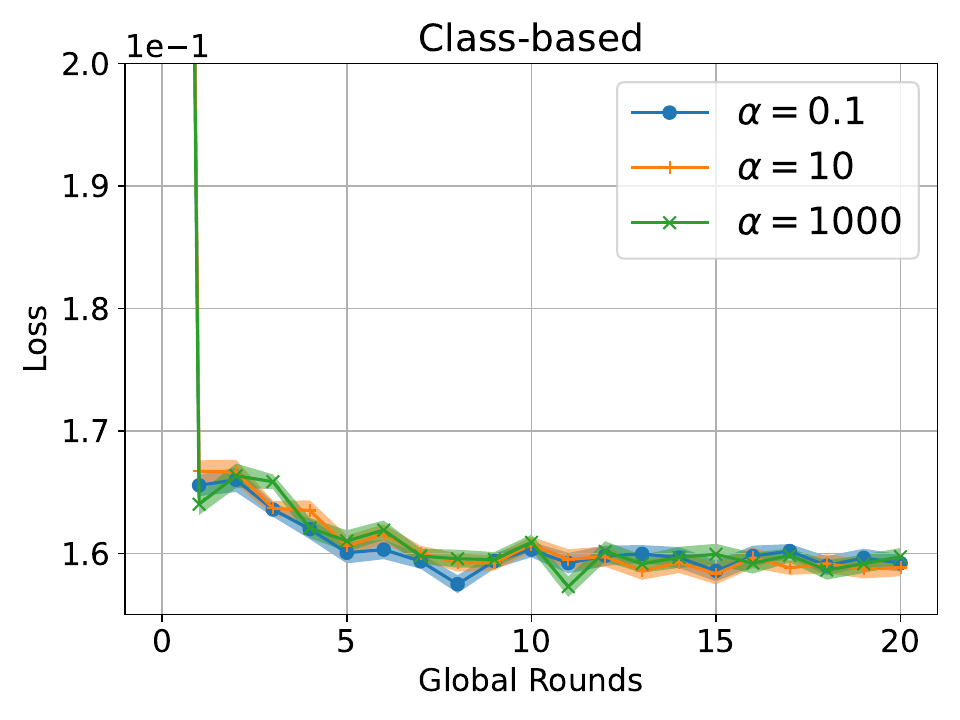}
    \includegraphics[width=0.49\textwidth]{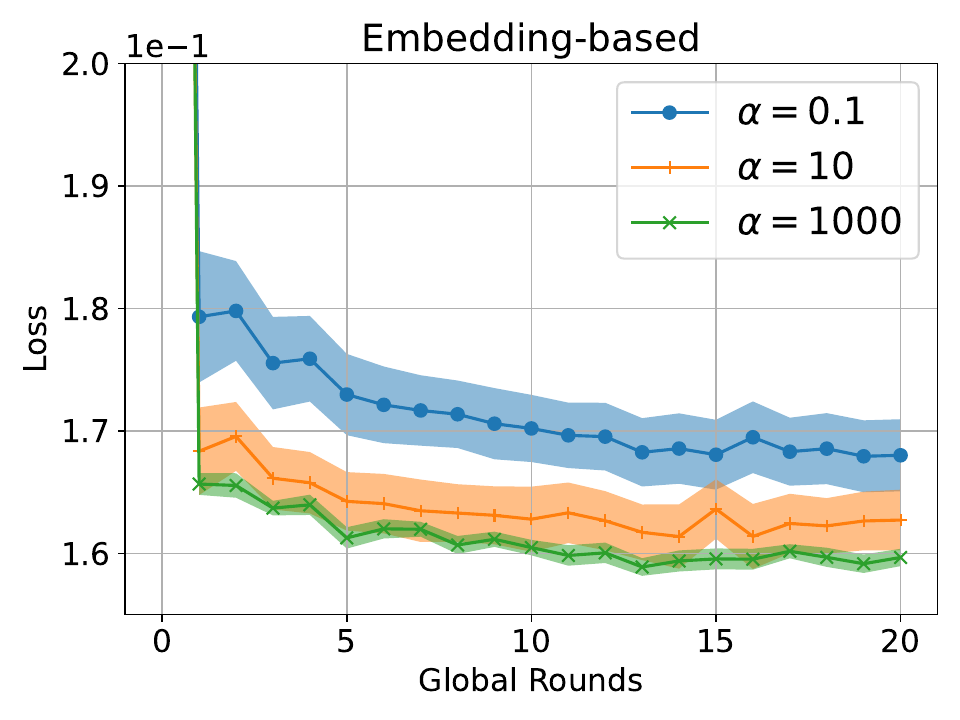}
    \end{tcolorbox}
    \hfill
    \begin{tcolorbox}[colback=BoxBg, colframe=BoxFrame, width=0.49\textwidth, left=0pt, right=0pt, top=-2pt, bottom=-2pt, before=, title=Semantic Segmentation --- \textbf{FedRep}, halign title=flush center,toptitle=-3pt,
    bottomtitle=-3pt]
    \includegraphics[width=0.49\textwidth]{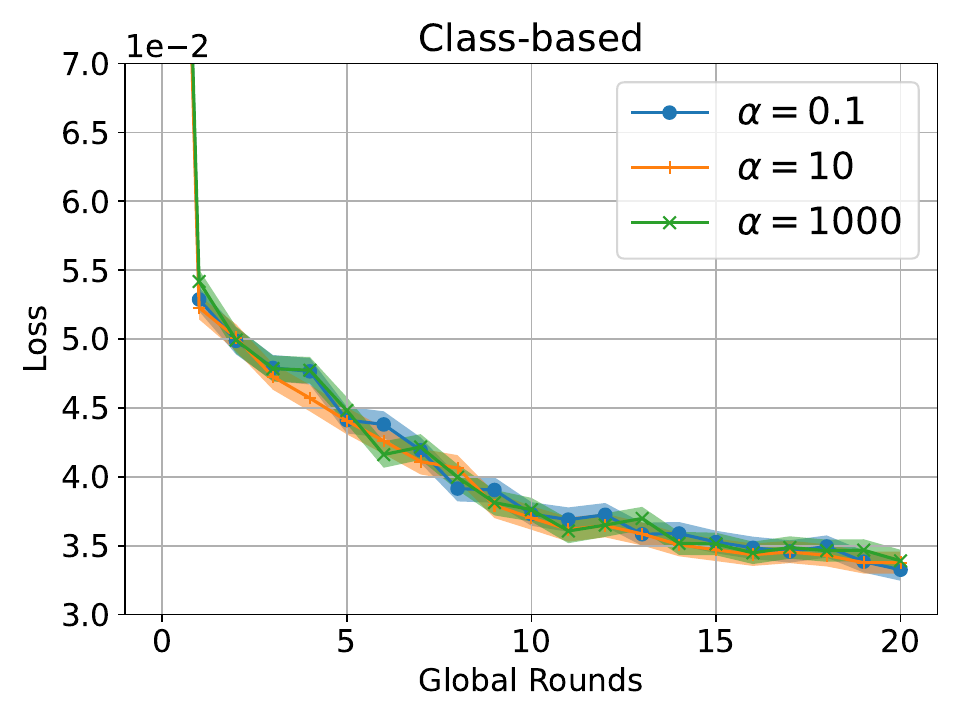}
    \includegraphics[width=0.49\textwidth]{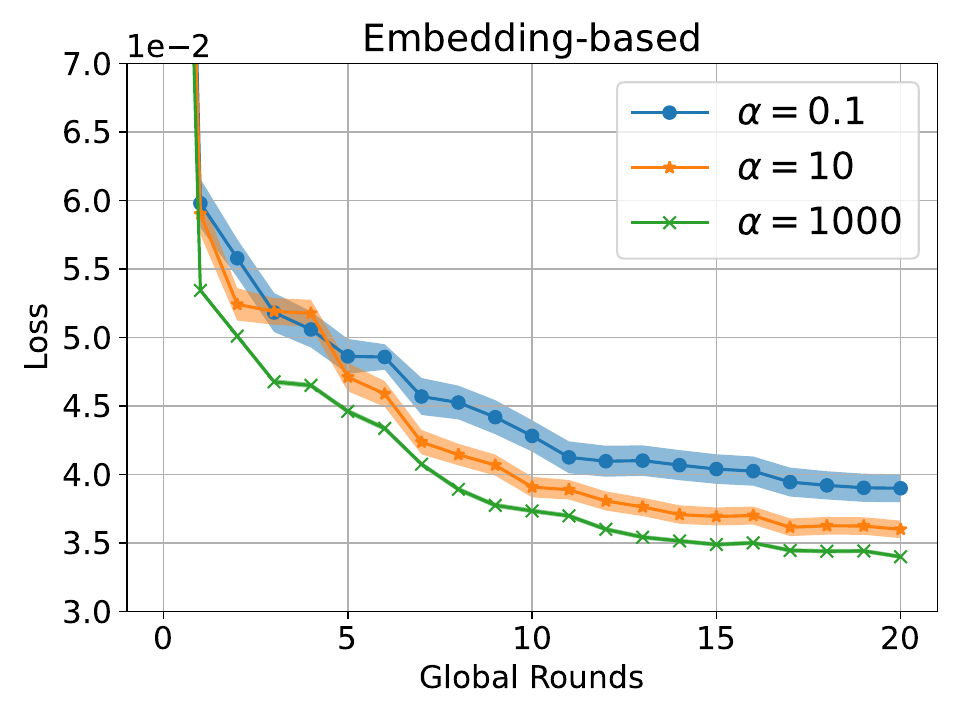}
    \end{tcolorbox}

\vspace{-3.6mm}
    \begin{tcolorbox}[colback=BoxBg, colframe=BoxFrame, width=0.49\textwidth, left=0pt, right=0pt, top=-2pt, bottom=-2pt, after=\hspace{2mm}, title=Euclidean Depth Estimation --- \textbf{FedRep}, halign title=flush center,toptitle=-3pt,
    bottomtitle=-3pt]
    \includegraphics[width=0.49\textwidth]{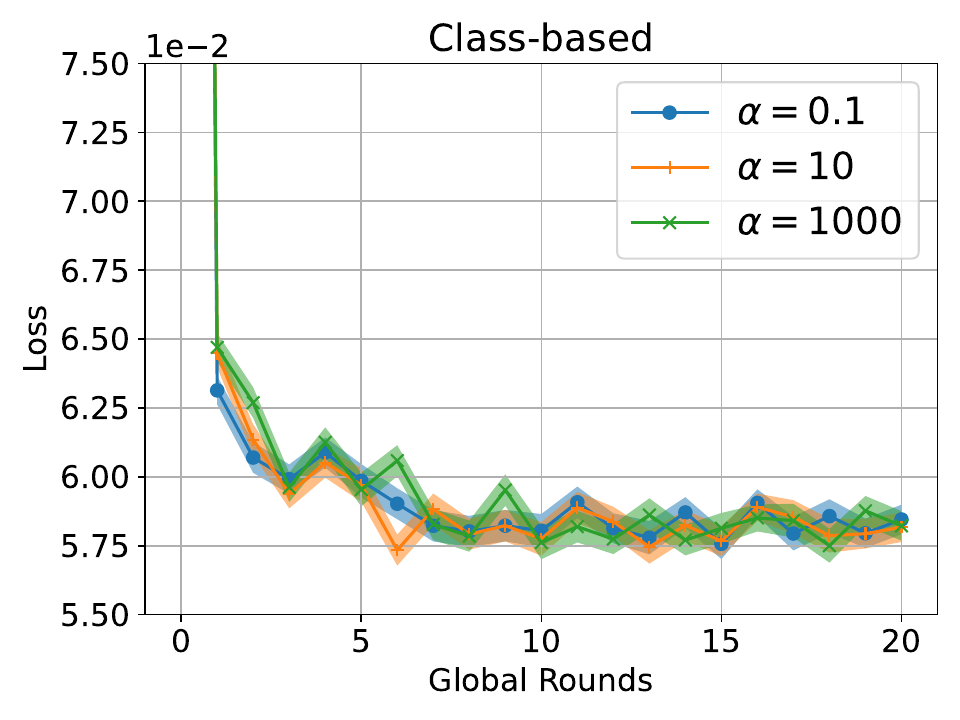}
    \includegraphics[width=0.49\textwidth]{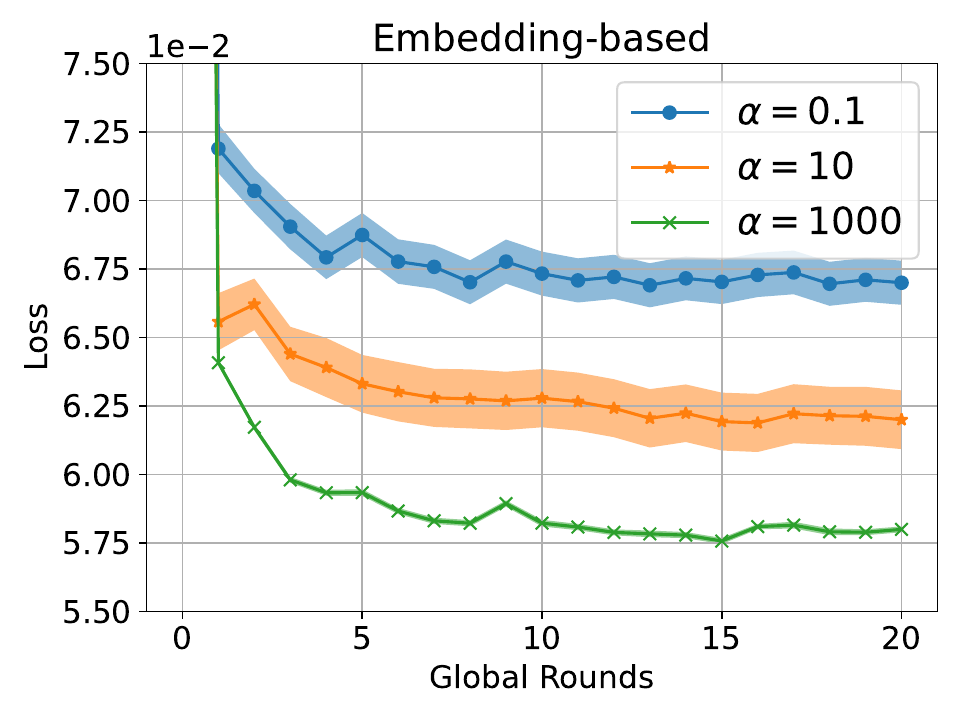}
    \end{tcolorbox}
    \hfill
    \begin{tcolorbox}[colback=BoxBg, colframe=BoxFrame, width=0.49\textwidth, left=0pt, right=0pt, top=-2pt, bottom=-2pt, title=3D Keypoints --- \textbf{FedRep}, halign title=flush center, before=, toptitle=-3pt,
    bottomtitle=-3pt]
    \includegraphics[width=0.49\textwidth]{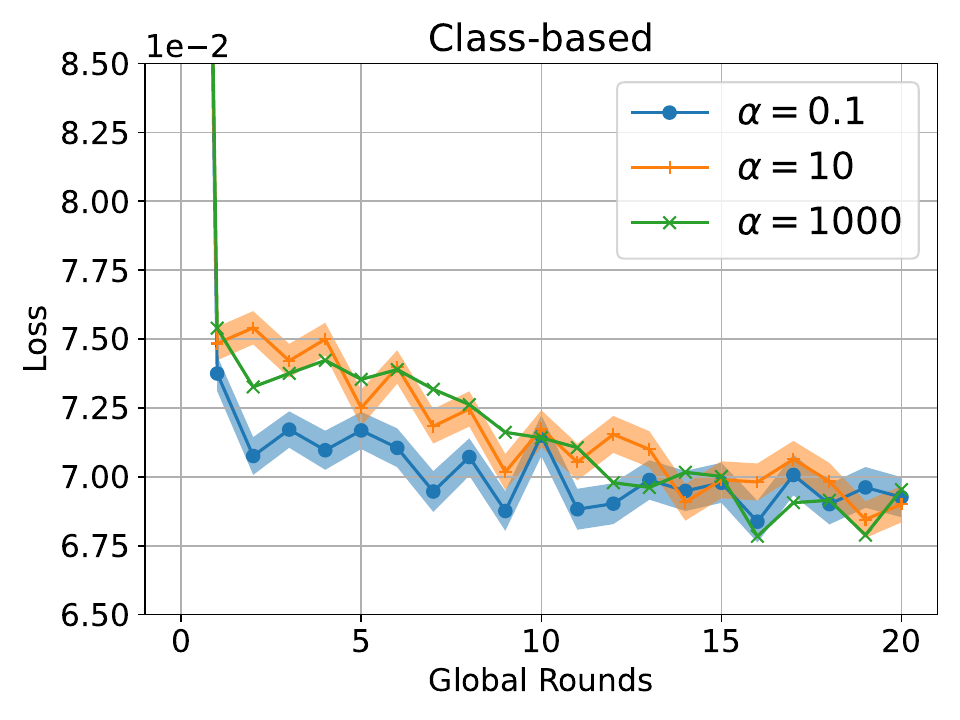}
    \includegraphics[width=0.49\textwidth]{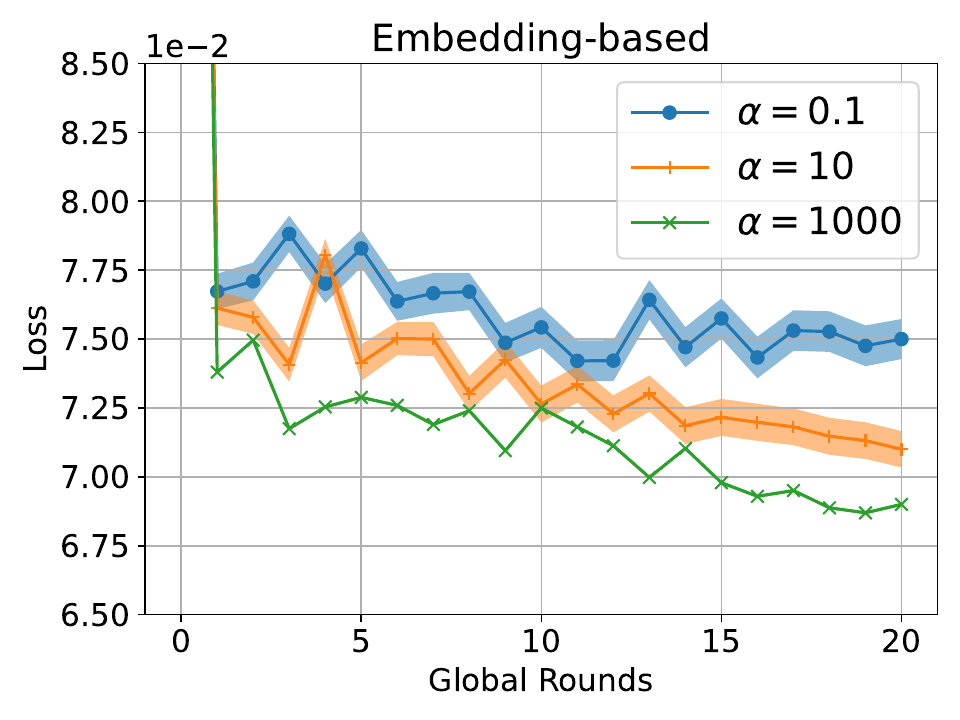}
    \end{tcolorbox}

     \vspace{-3.6mm}
    \begin{tcolorbox}[colback=BoxBg, colframe=BoxFrame, width=0.49\textwidth, left=0pt, right=0pt, top=-2pt, bottom=-2pt, after=\hspace{2mm}, title=Scene Classification --- \textbf{FedRep}, halign title=flush center,toptitle=-3pt,
    bottomtitle=-3pt]
    \includegraphics[width=0.49\textwidth]{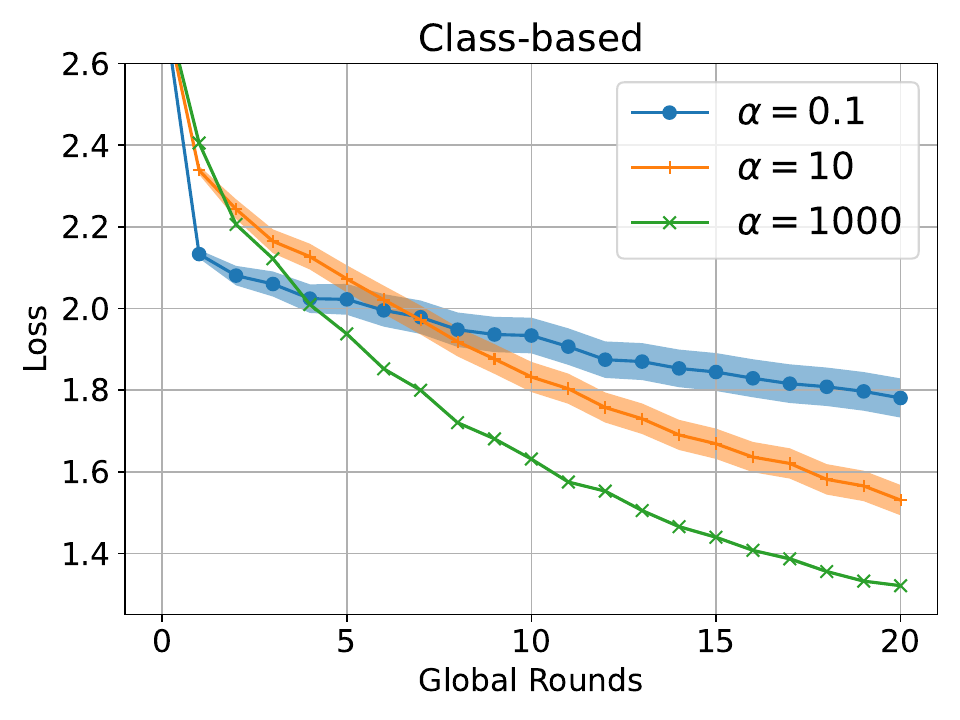}
    \includegraphics[width=0.49\textwidth]{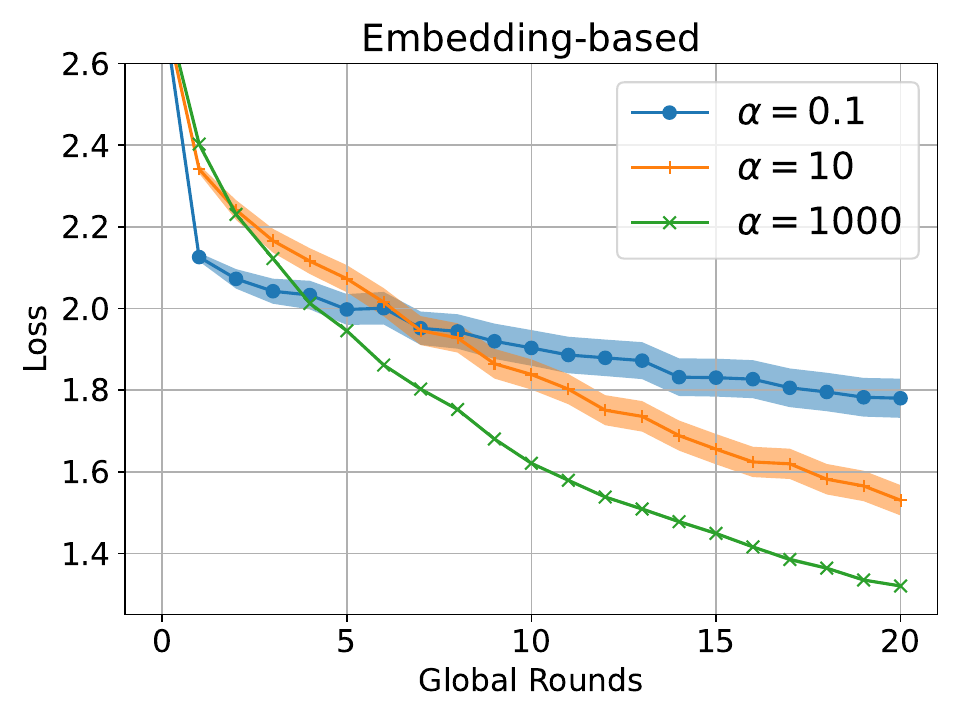}
    \end{tcolorbox}

    \caption{\textbf{Class-based vs. Embedding-based distribution:} Comparison of how performing Dirichlet distribution over the datapoints' labels (equivalently, the scene class feature) and the extracted embeddings affect the performance in FL when the FedRep technique is used.  Solid lines show the mean loss across clients; the shaded region denotes client-level dispersion (standard deviation of the loss across clients) at each global round.  The same phenomena observed in Fig.~\ref{fig:embedding-vs-class-p1} can be observed in the FedRep results.}
    \label{fig:fedrep_results}
    \vspace{-5mm}
\end{figure*}

\subsection{Adaptation of the Methods to Computer Vision Tasks}

\begin{figure*}[!h]
    \centering
    \begin{tcolorbox}[colback=BoxBg, colframe=BoxFrame, width=0.49\textwidth, left=0pt, right=0pt, top=-2pt, bottom=-2pt, after=\hspace{2mm}, title=2D Edges --- \textbf{FedAmp}, halign title=flush center,toptitle=-3pt,
    bottomtitle=-3pt]
    \includegraphics[width=0.48\textwidth]{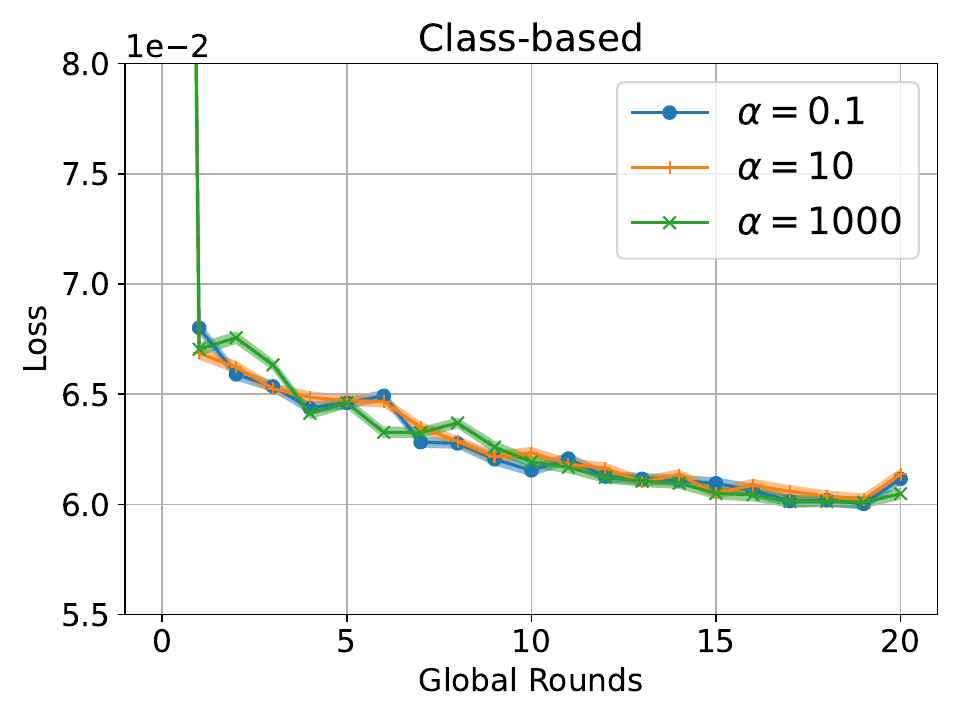}
    \includegraphics[width=0.48\textwidth]{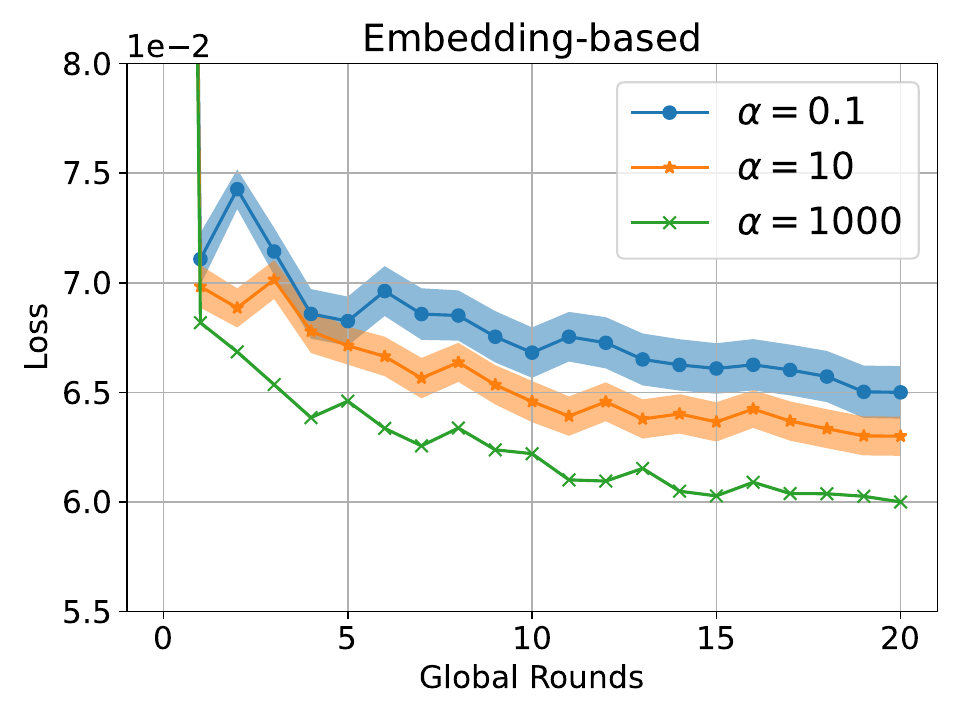}
    \end{tcolorbox}
    \hfill
    \begin{tcolorbox}[colback=BoxBg, colframe=BoxFrame, width=0.49\textwidth, left=0pt, right=0pt, top=-2pt, bottom=-2pt, before=, title=Reshading --- \textbf{FedAmp}, halign title=flush center,toptitle=-3pt,
    bottomtitle=-3pt]
    \includegraphics[width=0.48\textwidth]{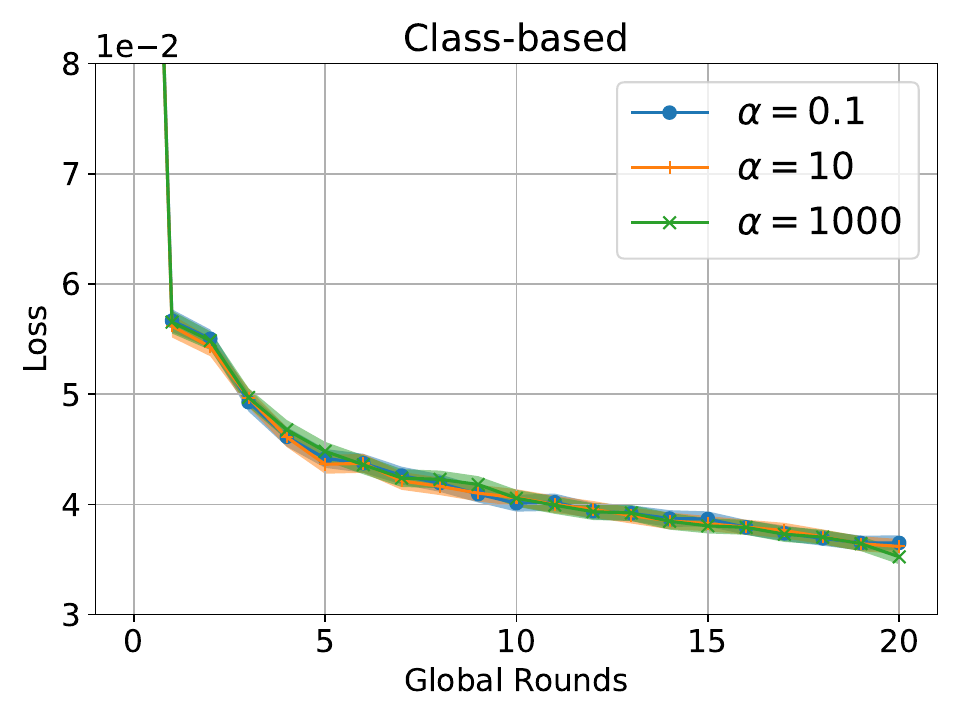}
    \includegraphics[width=0.48\textwidth]{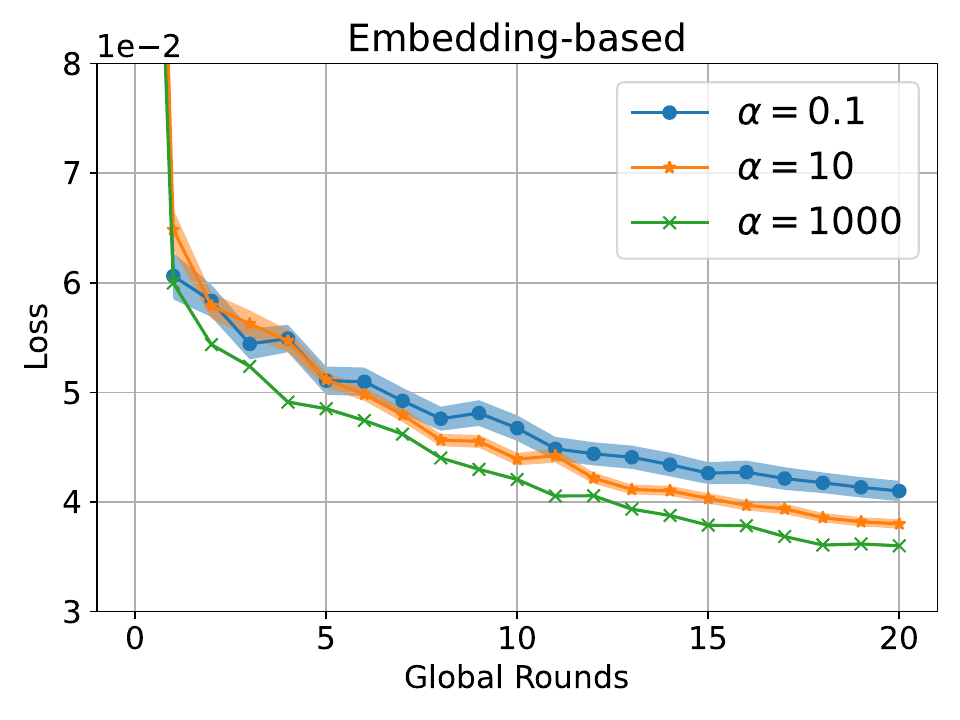}
    \end{tcolorbox}
    \vspace{-3.6mm}
     
    \begin{tcolorbox}[colback=BoxBg, colframe=BoxFrame, width=0.49\textwidth, left=0pt, right=0pt, top=-2pt, bottom=-2pt, after=\hspace{2mm}, title=Surface Normals --- \textbf{FedAmp}, halign title=flush center,toptitle=-3pt,
    bottomtitle=-3pt]
    \includegraphics[width=0.49\textwidth]{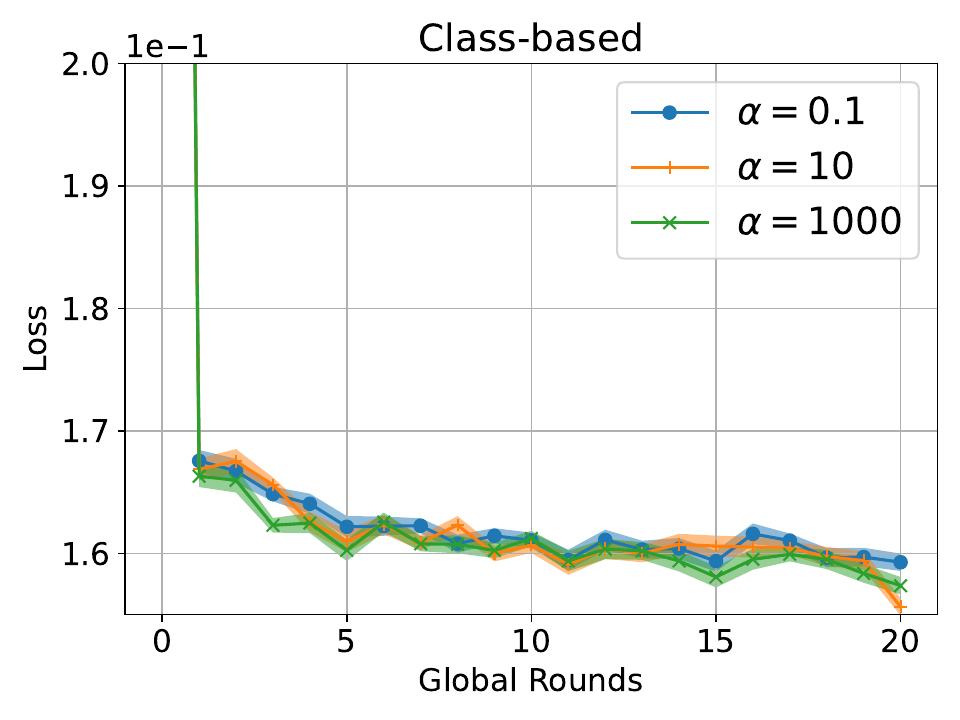}
    \includegraphics[width=0.49\textwidth]{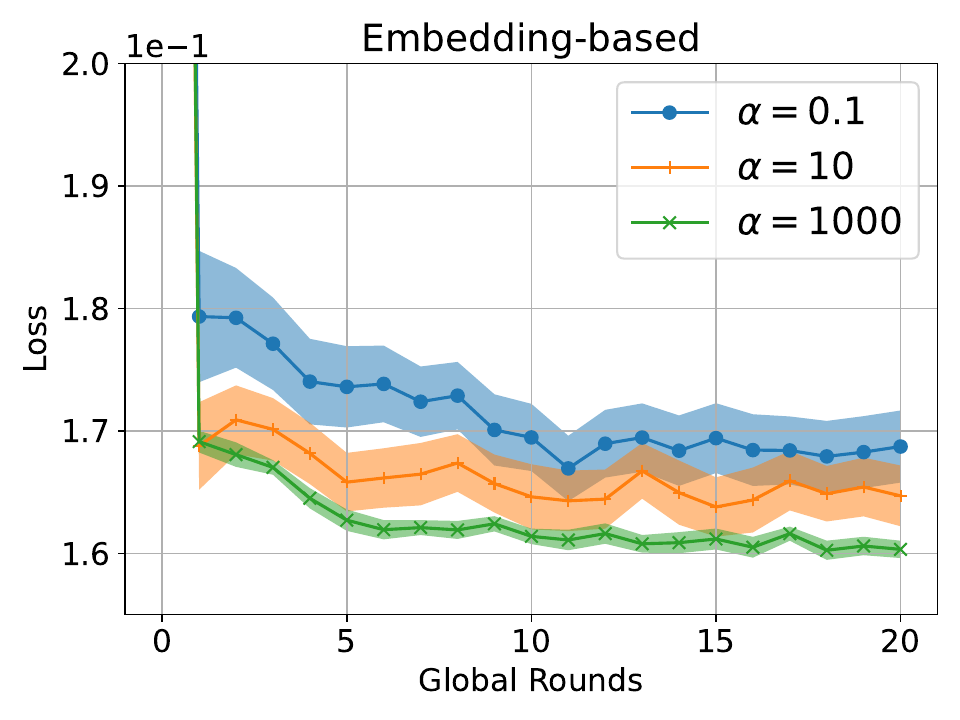}
    \end{tcolorbox}
    \hfill
    \begin{tcolorbox}[colback=BoxBg, colframe=BoxFrame, width=0.49\textwidth, left=0pt, right=0pt, top=-2pt, bottom=-2pt, before=, title=Semantic Segmentation --- \textbf{FedAmp}, halign title=flush center,toptitle=-3pt,
    bottomtitle=-3pt]
    \includegraphics[width=0.49\textwidth]{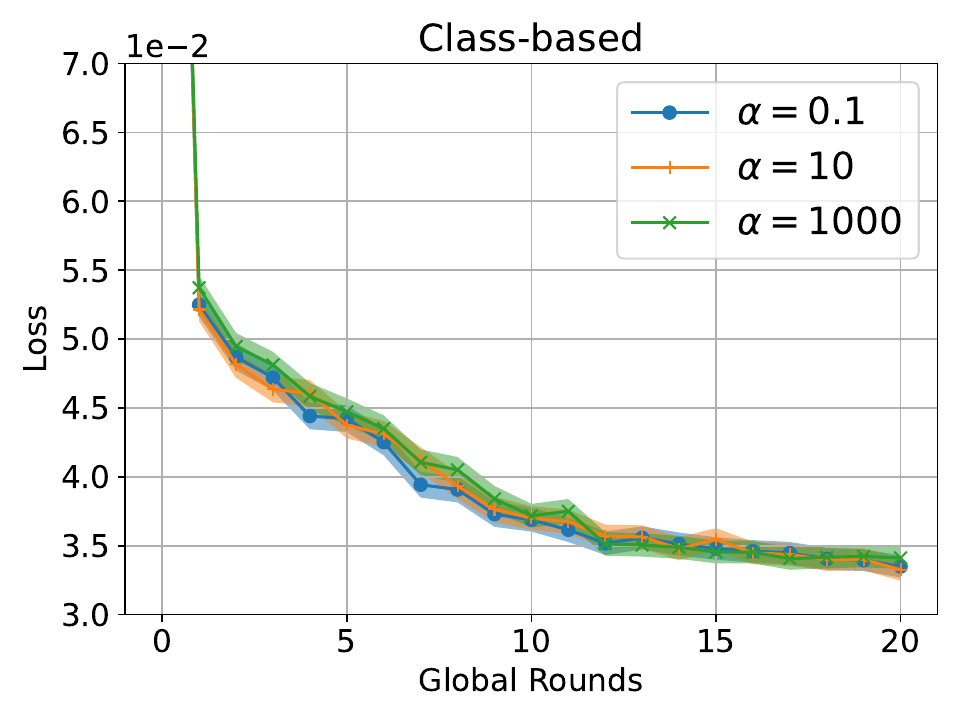}
    \includegraphics[width=0.49\textwidth]{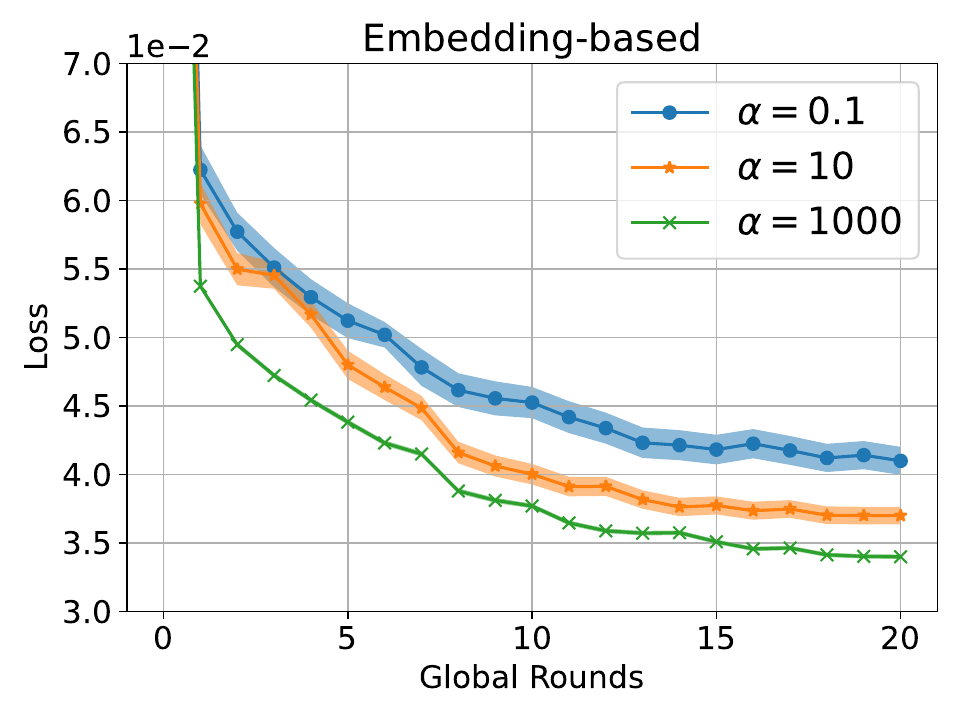}
    \end{tcolorbox}
 \vspace{-3.6mm}
 
    \begin{tcolorbox}[colback=BoxBg, colframe=BoxFrame, width=0.49\textwidth, left=0pt, right=0pt, top=-2pt, bottom=-2pt, after=\hspace{2mm}, title=Euclidean Depth Estimation --- \textbf{FedAmp}, halign title=flush center,toptitle=-3pt,
    bottomtitle=-3pt]
    \includegraphics[width=0.49\textwidth]{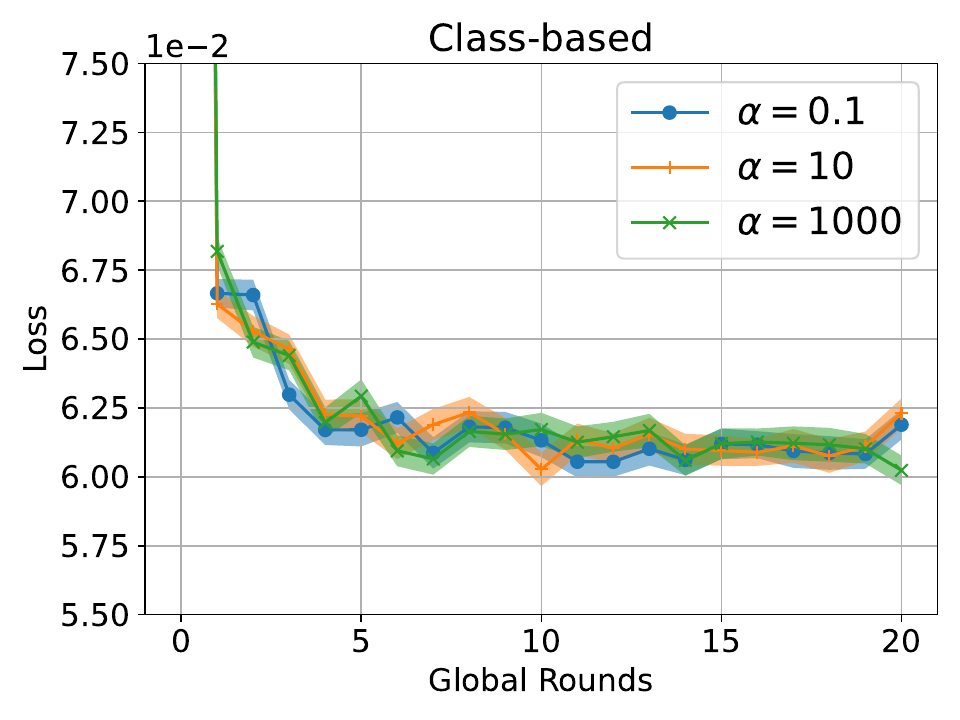}
    \includegraphics[width=0.49\textwidth]{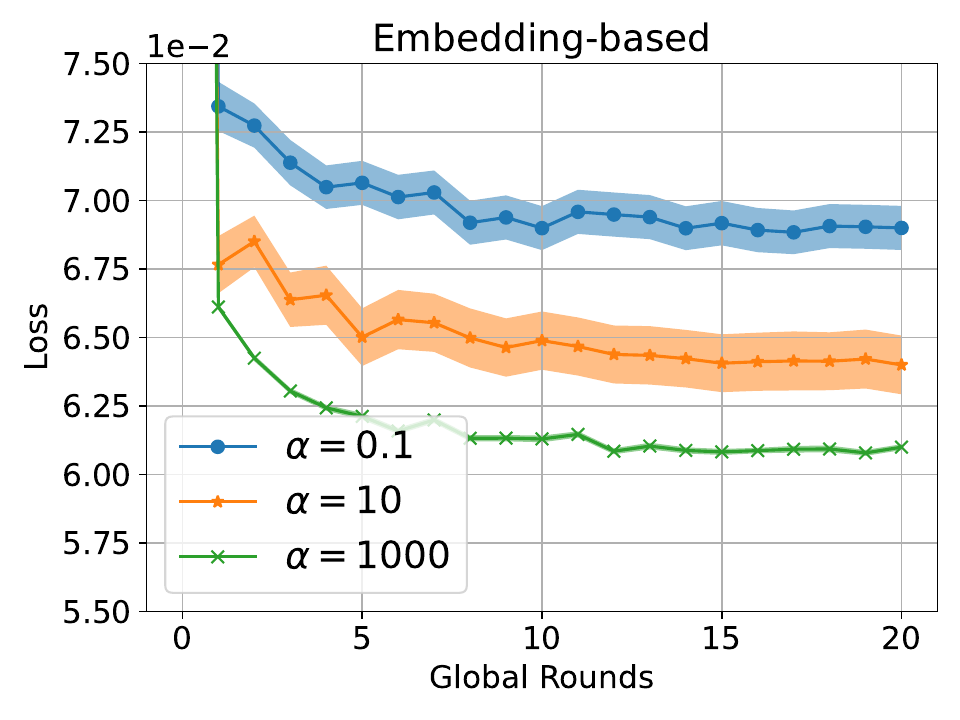}
    \end{tcolorbox}
    \hfill
    \begin{tcolorbox}[colback=BoxBg, colframe=BoxFrame, width=0.49\textwidth, left=0pt, right=0pt, top=-2pt, bottom=-2pt, title=3D Keypoints --- \textbf{FedAmp}, halign title=flush center, before=, toptitle=-3pt,
    bottomtitle=-3pt]
    \includegraphics[width=0.49\textwidth]{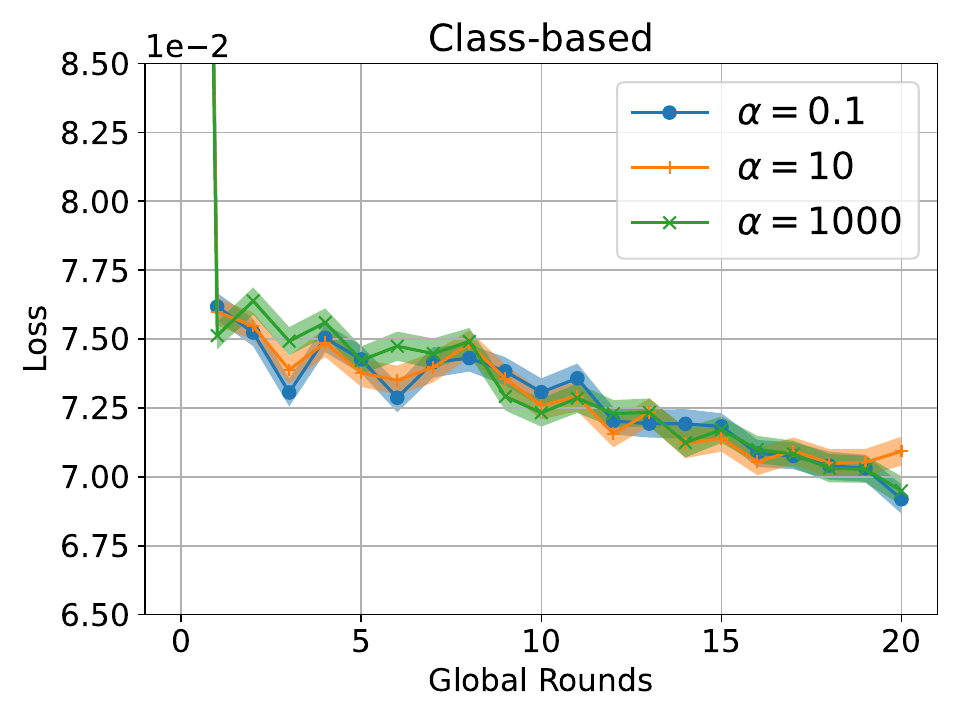}
    \includegraphics[width=0.49\textwidth]{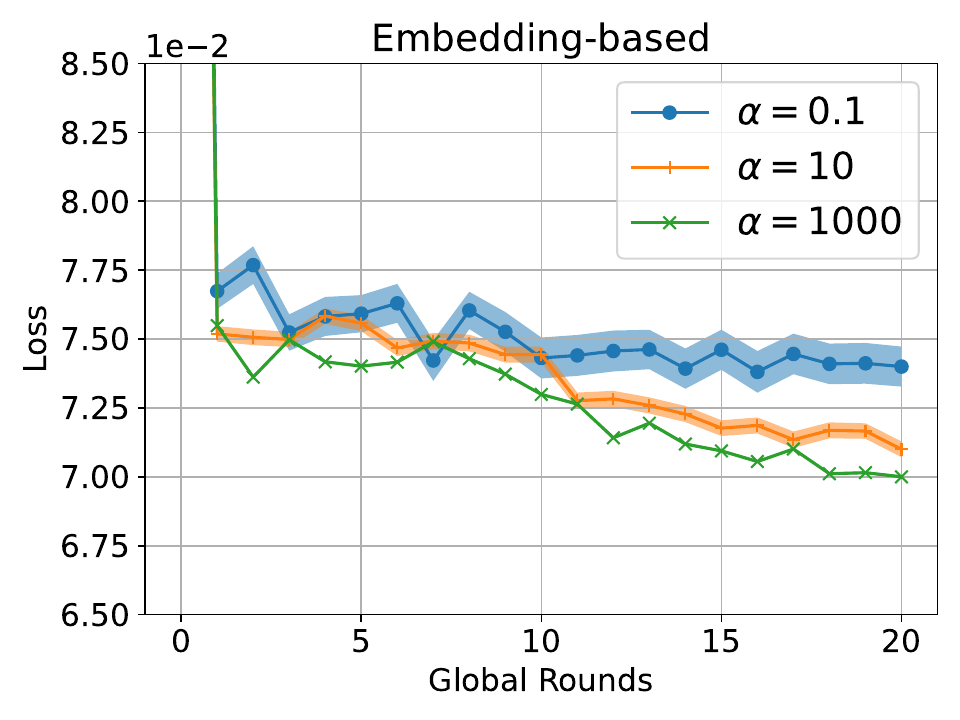}
    \end{tcolorbox}
     \vspace{-3.6mm}
     
    \begin{tcolorbox}[colback=BoxBg, colframe=BoxFrame, width=0.49\textwidth, left=0pt, right=0pt, top=-2pt, bottom=-2pt, after=\hspace{2mm}, title=Scene Classification --- \textbf{FedAmp}, halign title=flush center,toptitle=-3pt,
    bottomtitle=-3pt]
    \includegraphics[width=0.49\textwidth]{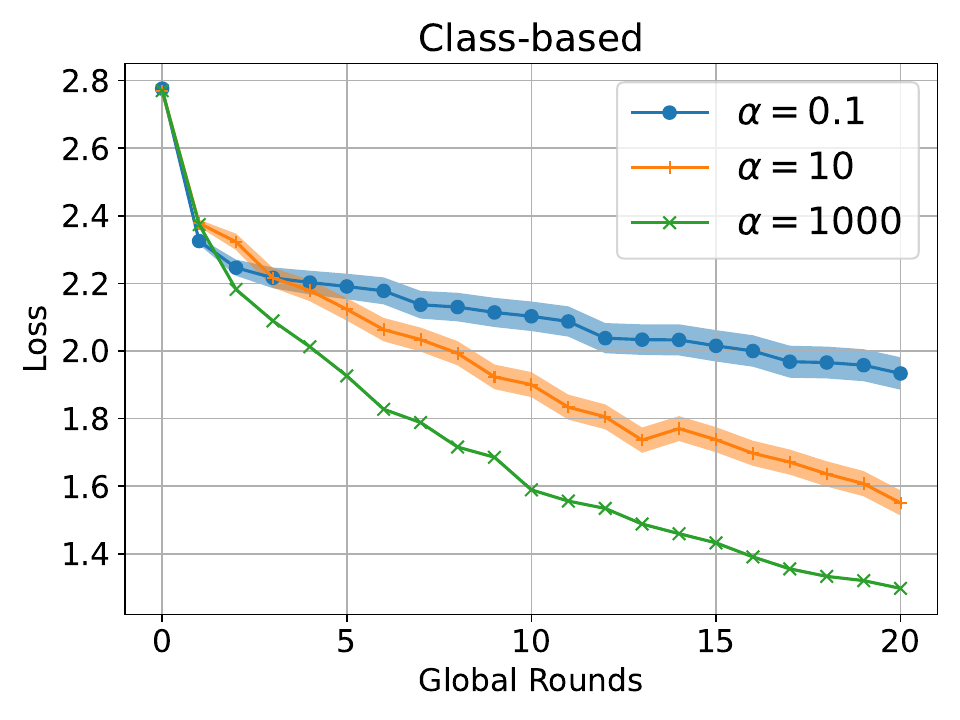}
    \includegraphics[width=0.49\textwidth]{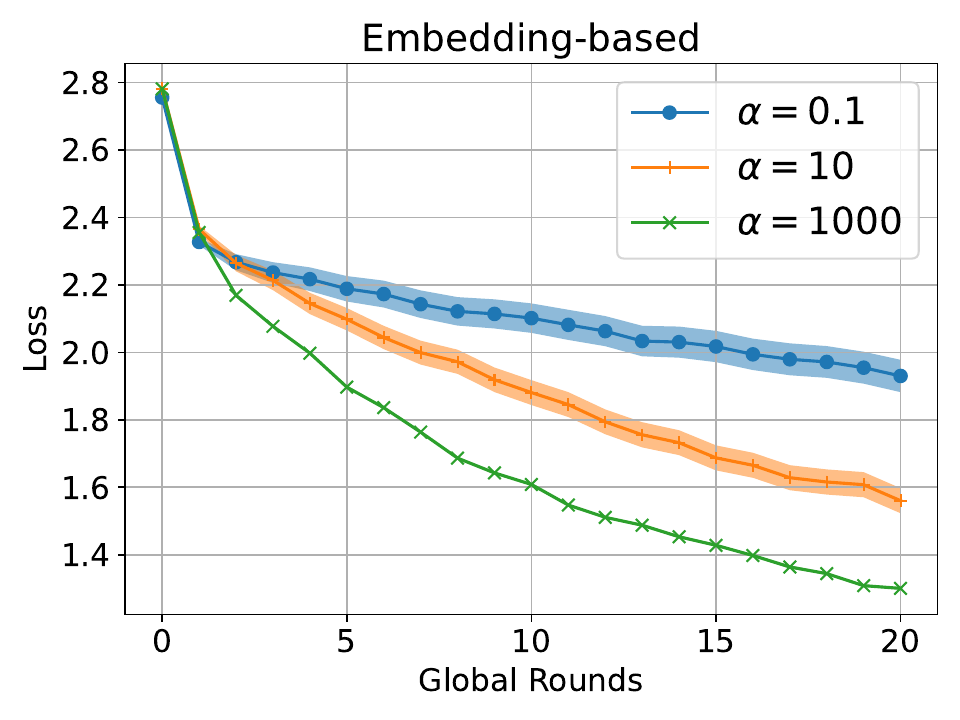}
    \end{tcolorbox}
    
    \caption{\textbf{Class-based vs. Embedding-based distribution:} Comparison of how performing Dirichlet distribution over the datapoints' labels (equivalently, the scene class feature) and the extracted embeddings affect the performance in FL when the FedAmp technique is used.  Solid lines show the mean loss across clients; the shaded region denotes client-level dispersion (standard deviation of the loss across clients) at each global round.  The same phenomena observed in Fig.~\ref{fig:embedding-vs-class-p1} can be observed in the FedAmp results.}
    \label{fig:fedamp_results}
    \vspace{-5mm}
\end{figure*}
Unlike classification benchmarks, the Taskonomy tasks used in this study (e.g., reshading, surface normals, etc.) require adapting FL baselines to encoder–decoder architectures with pixel-wise prediction heads. To ensure a fair comparison across methods, we adopt a consistent encoder–decoder backbone for all FL algorithms and specify, for each method, which components are treated as shared (aggregated) and which components are personalized below. 

\textbf{Shared vs. Personalized Architecture Components:}
For all methods, we decompose the network into an encoder that learns general feature representations and a decoder head that maps representations to pixel-wise or continuous outputs corresponding to the task being learned.
The encoder corresponds to the aggregated component in methods such as FedAvg, FedProx, SCAFFOLD, and FedAmp. The decoder head is handled according to each method’s personalization design, as detailed below.

\textbf{FedAvg, FedProx, and SCAFFOLD:}
For these three baselines, the entire encoder–decoder network is treated as fully shared across all clients, and all parameters are aggregated after each communication round. FedProx additionally incorporates the proximal term during local updates, and SCAFFOLD employs control variates to correct local gradients. No structural changes to the architecture are required.

\textbf{FedRep:}
FedRep’s original formulation uses a personalized classifier layer. For pixel-wise tasks, we extend this idea by treating the decoder output layer as the personalized component while keeping the encoder and the rest of the decoder shared (i.e., aggregated across clients). This adaptation maintains the spirit of FedRep (i.e., personalizing the task-specific output layer), while allowing consistent models to be learned collaboratively. In this regard, we follow the representation–head decoupling paradigm of~\cite{fedrep}, which emphasizes that the decoder or task head often captures client-specific variations in the downstream tasks. Our adaptations to FedRep adhere to this principle by personalizing decoder output layers where appropriate, while aggregating the rest of the model parameters.

\textbf{FedAmp:}
We make two modifications to adapt FedAmp to the pixel-wise prediction tasks in Taskonomy. First, we include the full model parameters (encoder + decoder) in the attentive message passed between clients during the proximal update. Second, to compute client–to-client similarity, we measure the distance between their decoder parameters (reflecting task-specific behavior) and apply softmax normalization to obtain similarity weights for aggregation.

\textbf{Loss Functions:}
The vision task's loss functions used across all methods are identical (represented in the y-axis of the figures). All extra loss terms correspond to the original FL method (e.g., for FedProx and FedAmp).

\textbf{Parameter Budget Alignment:}
All FL methods use the same encoder–decoder backbone, ensuring identical parameter counts (methods that personalize the decoder (e.g., FedRep) add no extra shared parameters). Thus, performance differences reflect method behavior (not architectural advantage).

\subsection{Further Results and Discussions} \label{sec:results}
% To provide insight into our proposed modeling of data heterogeneity, we have conducted two sets of experiments, focusing 
% on different implications of inducing heterogeneity through task perspective, which are
% \begin{enumerate}[label=(Set \arabic*), leftmargin=12mm]
%     \item Studying the performance of various benchmarks under different levels of data heterogeneity
%     \item Studying the task similarity through comparing the embedding clusters 
% \end{enumerate}
% These sets of experiments are discussed below in order.
\vspace{-.1mm}
\subsubsection{Studying the Performance of Various Benchmarks under Different Levels of Heterogeneity} \label{subsubsec:benchmark-results}
To provide a benchmark of FL performance, we compare the loss value of the benchmarks\footnote{Note that ``loss" is used as a unified metric that enables universal comparisons across the tasks. For example, other metrics such as \textit{accuracy}, which is mostly tailored to classification, do not benefit from such a universality.}
 explained above
under class-based and embedding-based data distribution (the same system of $25$ clients as in Sec.~\ref{sec:class-based-experiments} is considered). 
This benchmarking provides a baseline performance for each of these SoTA methods under various levels of data heterogeneity. 
Note that we have shown the results for FedAvg in Fig. \ref{fig:embedding-vs-class-p1}, which were used to explain our methodology. Subsequently, in the following, we discuss the results for FedProx (see Fig. \ref{fig:fedprox-results}), SCAFFOLD (see Fig. \ref{fig:scaffold_results}), FedRep (see Fig. \ref{fig:fedrep_results}), and FedAmp (see Fig. \ref{fig:fedamp_results}).

Observing the results in Figs. \ref{fig:fedprox-results}-\ref{fig:fedamp_results}, we can observe the same phenomenon as in Fig. \ref{fig:embedding-vs-class-p1} (see Sec.~\ref{sec:embedding-based-experiments}). In particular, the results indicate that, for the non-classification tasks (i.e., all the tasks except ``Scene Classification" presented in the bottom box of each figure), the change in the value of $\alpha$ in the dirichlet distribution in all methods results in a performance gap when our embedding-based distribution of data is utilized (see the right plot in each box). However, at the same time, it can be observed that such a performance gap does not hold for class-based distribution of data (see the left plot in each box).  
Further, our results verify that embedding-based data heterogeneity and class-based data heterogeneity result in a similar FL performance for the classification task (see the bottom box in Figs. \ref{fig:fedprox-results}-\ref{fig:fedamp_results}). The main two takeaways of the results are as follows. (i)
These observations further validate that our proposed method of inducing heterogeneity from the tasks' perspective is capable of creating a heterogeneous data setting more effectively for non-classification vision tasks. (ii)~In addition to revealing the performance of all the benchmarks under embedding-based data heterogeneity, the results reveal that all these SoTA methods are subject to overestimation of FL performance under class-based data heterogeneity for vision tasks: when class-based data heterogeneity is induced, the loss is often lower than that of the embedding-based data heterogeneity scenario. In other words, embedding-based data heterogeneity can induce a more pronounced impact on the FL performance (i.e., a higher value of loss).

\begin{table*}[t]
\vspace{-4mm}
\caption{Heatmap representation of task similarities based on the mean Adjusted Rand Index (ARI) with permutation-based $p$-values.}
{\notsotiny
\centering
\begin{tabularx}{\textwidth}{
| >{\centering\arraybackslash}m{28mm}
? >{\centering\arraybackslash}X
  >{\centering\arraybackslash}X
  >{\centering\arraybackslash}X
  >{\centering\arraybackslash}X
  >{\centering\arraybackslash}X
  >{\centering\arraybackslash}X
  >{\centering\arraybackslash}X|}
\hline
\rowcolor{DarkerBlue}
\diagbox[height=5mm,width=32mm,linecolor=white]{\tiny \color{white}Task 1}{\tiny \color{white}Task 2} &
{\color{white}Euclidean Depth Estimation} &
{\color{white}2D Edges} &
{\color{white}3D Keypoints} &
{\color{white}Surface Normals} &
{\color{white}Reshading} &
{\color{white}Semantic Segmentation} &
{\color{white}Scene Classification} \\
\specialrule{1.5pt}{0pt}{0pt}

% ---------------- ROW 1: DEPTH --------------------
\cellcolor{DarkerBlue}{\color{white}Euclidean Depth Estimation} &
\cellcolor{purple!0}   1.000 &
\cellcolor{purple!72}  0.079 (p=0.0099) &
\cellcolor{purple!40}  0.044 (p=0.0099) &
\cellcolor{purple!100} 0.115 (p=0.0099) &
\cellcolor{purple!100} 0.127 (p=0.0099) &
\cellcolor{purple!35}  0.042 (p=0.0099) &
\cellcolor{purple!25}  0.031 (p=0.0099) \\

% ---------------- ROW 2: EDGES --------------------
\cellcolor{DarkerBlue}{\color{white}2D Edges} &
\cellcolor{purple!72}  0.079 (p=0.0099) &
\cellcolor{purple!0}   1.000 &
\cellcolor{purple!55}  0.060 (p=0.0099) &
\cellcolor{purple!90}  0.096 (p=0.0099) &
\cellcolor{purple!60}  0.066 (p=0.0099) &
\cellcolor{purple!80}  0.091 (p=0.0099) &
\cellcolor{purple!35}  0.043 (p=0.0099) \\

% ---------------- ROW 3: KEYPOINTS3D --------------------
\cellcolor{DarkerBlue}{\color{white}3D Keypoints} &
\cellcolor{purple!40}  0.044 (p=0.0099) &
\cellcolor{purple!55}  0.060 (p=0.0099) &
\cellcolor{purple!0}   1.000 &
\cellcolor{purple!52}  0.055 (p=0.0099) &
\cellcolor{purple!38}  0.041 (p=0.0099) &
\cellcolor{purple!25}  0.031 (p=0.0099) &
\cellcolor{purple!37}  0.045 (p=0.0099) \\

% ---------------- ROW 4: NORMALS --------------------
\cellcolor{DarkerBlue}{\color{white}Surface Normals} &
\cellcolor{purple!100} 0.115 (p=0.0099) &
\cellcolor{purple!90}  0.096 (p=0.0099) &
\cellcolor{purple!52}  0.055 (p=0.0099) &
\cellcolor{purple!0}   1.000 &
\cellcolor{purple!95}  0.102 (p=0.0099) &
\cellcolor{purple!50}  0.063 (p=0.0099) &
\cellcolor{purple!30}  0.040 (p=0.0099) \\

% ---------------- ROW 5: RESHADING --------------------
\cellcolor{DarkerBlue}{\color{white}Reshading} &
\cellcolor{purple!100} 0.127 (p=0.0099) &
\cellcolor{purple!60}  0.066 (p=0.0099) &
\cellcolor{purple!38}  0.041 (p=0.0099) &
\cellcolor{purple!95}  0.102 (p=0.0099) &
\cellcolor{purple!0}   1.000 &
\cellcolor{purple!45}  0.054 (p=0.0099) &
\cellcolor{purple!36}  0.044 (p=0.0099) \\

% ---------------- ROW 6: SEMANTIC SEGMENTATION --------------------
\cellcolor{DarkerBlue}{\color{white}Semantic Segmentation} &
\cellcolor{purple!35}  0.042 (p=0.0099) &
\cellcolor{purple!80}  0.091 (p=0.0099) &
\cellcolor{purple!25}  0.031 (p=0.0099) &
\cellcolor{purple!50}  0.063 (p=0.0099) &
\cellcolor{purple!45}  0.054 (p=0.0099) &
\cellcolor{purple!0}   1.000 &
\cellcolor{purple!53}   0.061 (p=0.0099) \\

% ---------------- ROW 7: SCENE CLASSIFICATION --------------------
\cellcolor{DarkerBlue}{\color{white}Scene Classification} &
\cellcolor{purple!25}  0.031 (p=0.0099) &
\cellcolor{purple!35}  0.043 (p=0.0099) &
\cellcolor{purple!37}  0.045 (p=0.0099) &
\cellcolor{purple!30}  0.040 (p=0.0099) &
\cellcolor{purple!36}  0.044 (p=0.0099) &
\cellcolor{purple!53}  0.061 (p=0.0099) &
\cellcolor{purple!0}   1.000 \\
\hline
\end{tabularx}}
\label{tab:ari_heatmap}
\vspace{-5mm}
\end{table*}

Further, a major takeaway of implementing these techniques were as follows: the tasks exhibit model bias at different numbers of local SGD iterations ($K$), which depends on the task complexity and the underlying data distributions. In other words, some tasks may need to be trained over longer durations of local training (i.e., higher values of $K$) to exhibit the model bias and subsequently a gap in performance under various levels of data heterogeneity. Nevertheless, to have fair comparisons between class-based and embedding-based methods, we maintained the same number of local SGDs per round for both class-based and embedding-based experiments across all benchmark methods and all tasks. The details of these hyperparameters are available at our aforementioned GitHub repository.  Also, for an at-a-glance comparison of all FL methods, refer to Appendix~A.

\subsubsection{Studying the Task Similarity Through Comparing the Embedding Clusters} \label{subsubsec:heatmap}
Revisiting Remarks~\ref{remark:task-similarity} and \ref{remark:multi-task}, it can be noted that the extracted embeddings of the datapoints and the way that they get clustered in our proposed approach may have further use cases. 
% As a preliminary investigation in this direction, we measure the similarities between the embedding clusters, which provide nuanced insights.
  As a preliminary investigation in this direction, in Table~\ref{tab:ari_heatmap}, we calculate the similarity of embedding clusters across pairs of tasks through \textit{Adjusted Rand Index (ARI)}, a standard clustering measure with well-understood statistical properties~\cite{hubert1985comparing}, and visualize the resulting cross-task similarities using a heatmap. 
In essence, ARI measures how consistently two different tasks cluster the same set of datapoints: it counts how often pairs of data samples are placed in the same cluster or in different clusters by both tasks, and then corrects this agreement for what would be expected purely by chance. As a result, ARI takes the value 1 when the two tasks clusterings match perfectly, 0 when the alignment is no better than random, and can even become negative when the agreement is worse than random.

Inspecting the results in Table \ref{tab:ari_heatmap}, the \textit{Euclidean Depth Estimation} and the \textit{Reshading} tasks have the highest average ARI across their embedding clusters ($0.127$) among the pairs included in the table.
The similarity between the output domains of these two tasks and the general aim each one is accomplishing can help explain the higher score: both of these tasks are driven by the same geometric structure of a scene (e.g., how surfaces are shaped, oriented, and spatially arranged) and since their outputs rely on these shared cues, they partition images along similar geometric patterns, resulting in more closely aligned embedding clusters and a higher ARI. On the other hand, the two lowest similarity ratios (i.e., both $0.031$) belong to the \textit{3D Keypoints} task when compared with \textit{Semantic Segmentation} task and \textit{Euclidean Depth Estimation} when compared with \textit{Scene Classification} task.
This can also be explained through the fact that both 3D Keypoints and Euclidean Depth Estimation are tasks that consider the very details of the image which is contrary to the higher-level inspection conducted in the Semantic Segmentation and Scene Classification tasks.
The rest of the pairwise comparisons fall in between these two extremes. In particular, many of the numerical values are hard to interpret intuitively since human-level visual explanations may not suffice to describe the innate characteristics of these tasks. This makes the reach of the methodology used to obtain the heatmap in Table \ref{tab:ari_heatmap} more pronouncedly felt in scenarios that human-level intuitions are non-existent. The extensions of the implications of these results to task grouping and multi-task training is left as future work (see Remarks~\ref{remark:task-similarity} and \ref{remark:multi-task}).
Further, it can be noted that the ARI values reported in Table \ref{tab:ari_heatmap} are numerically small (reflecting that the tasks are largely distinct from one another), where the accompanying permutation-based 
p-values confirm that these similarities are statistically meaningful/reliable. In particular, the consistently low p-values ($p = 0.0099\simeq0.01$, measured across $100$ permutations) show that even the modest ARI scores observed are significantly above what would arise by chance, indicating task-induced structure rather than noise.

\subsubsection{Studying the Effect of the Number of Clusters on the Silhouette Score} \label{subsubsec:fedavg-seed}
To assess the structural quality of the learned embeddings, we employ the \textit{Silhouette coefficient}, a standard cluster validity index~\cite{shahapure2020cluster}. In words, Silhouette coefficient measures how well each datapoint fits within its assigned cluster compared to its proximity to neighboring clusters, summarizing both compactness and separation. Its values range in $[-1,1]$, where higher values indicate clearer and more meaningful cluster structure.
We compute this coefficient for $k \in \{2,4,10,16,32\}$ across all tasks. 
As shown in Appendix~B, the value of Silhouette score increases marginally from $K=16$ onward. This diminishing return, combined with the  increase in computational cost associated with larger $k$, motivates our choice of $n_{\mathrm{clusters}} = 16$ for all analyses in this paper.

% mong these, $k = 16$ consistently yields the highest average Silhouette scores with low variance, indicating a stable and well-separated clustering structure. 

\subsubsection{Studying the Effect of Randomness in Clustering on the Stability of the Observed Gaps in Performance} \label{subsubsec:fedavg-seed}
We analyze the effect of different random seeds used in K-means clustering on the performance gaps observed under our embedding-based data heterogeneity formulation. To this end, we revisit our main results in Fig.~\ref{fig:embedding-vs-class-p1} and repeat the experiments using three distinct K-means seeds.  We then report the mean of the resulting loss values alongside their uncertainty intervals (standard deviation) for each $\alpha$ in Table~\ref{tab:fedavg_cross_seed}.  The results show that the performance degradation at lower $\alpha$  values (i.e., higher loss values) consistently persists even after averaging across seeds, demonstrating that our embedding-based heterogeneity remains effective regardless of the seed initialization. Moreover, the standard deviations are significantly smaller than the corresponding mean losses, indicating that the observed performance gaps are stable and not driven by seed-dependent fluctuations: intuitively, this robustness arises because the embedding space already encodes strong geometric and structural cues, causing K-means to produce similar cluster partitions across different initializations.

\subsubsection{Studying Task Performance on Another Dataset}
To evaluate the generality of our approach beyond Taskonomy, we further compare embedding-based and class-based data heterogeneity on a different vision dataset: PASCAL~\cite{mottaghi2014role}, a widely used dataset spanning diverse object categories and scene structures. For this study, we apply FedAvg and report the results in Appendix~C. Consistent with our main findings, the embedding-based data splits on PASCAL induce clear performance gaps across various data heterogeneity levels, whereas the class-based data splits fail to meaningfully perturb the task-relevant feature space and have only a negligible effect on the trained FL model’s performance.

\subsubsection{Studying Task-Specific Metrics} As mentioned earlier, our choice of loss functions (namely, $\ell_1$ loss for Euclidean Depth Estimation, 2D Edges, Surface Normals, and 3D Keypoints; mean squared error for Reshading; and cross-entropy for Scene Classification and Semantic Segmentation) was motivated by their generic applicability across a broad range of vision tasks. For concreteness, we additionally provide a study using more task-specific metrics, where we employ F-measure for 2D Edges, 
PSNR for Reshading, mIoU for Semantic Segmentation, RMSE for Euclidean Depth Estimation, and Accuracy for Scene Classification. The results of these experiments are presented in Appendix~D. Consistent with our main findings, the embedding-based data heterogeneity again produces clear performance gaps across different heterogeneity levels, whereas class-based heterogeneity yields negligible performance variation.

\begin{table}[!t]
% \vspace{-1mm}
\centering
\caption{Average and standard deviation of the loss for each task in Taskonomy using FedAvg, across random seeds of K-means.}
\label{tab:fedavg_cross_seed}
\vspace{-1.5mm}
{\notsotiny
\setlength{\tabcolsep}{2pt}
\begin{tblr}{
  colspec = {
    Q[c,m,wd=5mm]    % alpha
    Q[c,m,wd=10mm]    % 2D Edges
    Q[c,m,wd=10mm]    % Reshading
    Q[c,m,wd=10mm]   % Surface Normals
    Q[c,m,wd=12mm]   % Semantic Seg.
    Q[c,m,wd=10mm]    % 3D Keypoints
    Q[c,m,wd=10mm]   % Depth
    Q[c,m,wd=10mm]    % Scene Cls.
  },
  hlines = {black,0.08mm},
  vlines = {black,0.08mm},
  cell{1}{1} = {r=2, c=1}{m, DarkBlue},
  cell{1}{2} = {r=1, c=7}{m, DarkBlue},
  row{2} = {bg=LightBlue},
  colsep=1pt,
  rowsep=1pt
}
\textbf{$\alpha$} & \textbf{Tasks} & & & & & & \\

&
\textbf{\tiny 2D Edges} &
\textbf{\tiny Reshading} &
\textbf{\tiny Surface Normals} &
\textbf{\tiny Semantic Segmentation} &
\textbf{\tiny 3D Keypoints} &
\textbf{\tiny Euclidean Depth Estimation} &
\textbf{\tiny Scene Classification} \\
\hline

0.1 &
$0.077$ {\tiny $\pm\!0.002$} &
$0.057$ {\tiny $\pm\!0.001$} &
$0.196$ {\tiny $\pm\!0.005$} &
$0.052$ {\tiny $\pm\!0.001$} &
$0.090$ {\tiny $\pm\!0.002$} &
$0.075$ {\tiny $\pm\!0.002$} &
$2.324$ {\tiny $\pm\!0.065$} \\

10 &
$0.074$ {\tiny $\pm\!0.002$} &
$0.054$ {\tiny $\pm\!0.001$} &
$0.190$ {\tiny $\pm\!0.004$} &
$0.048$ {\tiny $\pm\!0.001$} &
$0.085$ {\tiny $\pm\!0.002$} &
$0.074$ {\tiny $\pm\!0.002$} &
$1.779$ {\tiny $\pm\!0.053$} \\

1000 &
$0.072$ {\tiny $\pm\!0.002$} &
$0.048$ {\tiny $\pm\!0.001$} &
$0.186$ {\tiny $\pm\!0.005$} &
$0.048$ {\tiny $\pm\!0.001$} &
$0.084$ {\tiny $\pm\!0.002$} &
$0.071$ {\tiny $\pm\!0.002$} &
$1.357$ {\tiny $\pm\!0.033$} \\
\end{tblr}
}
\vspace{-.1mm}
\end{table}
 
\subsubsection{Studying other Embedding Extraction Methods} \label{sec:pretraining_ablation}

We conduct an ablation study comparing generic and task-aligned embedding extraction strategies, including frozen CLIP, self-supervised SimCLR, and task-supervised CLIP, to examine their ability to induce task-perspective data heterogeneity. The study shows that only task-supervised embeddings consistently reveal clear performance separation between low- and high-heterogeneity regimes, while generic embeddings fail to do so. Detailed  results and analysis are provided in Appendix~F.

\subsubsection{Studying the Effect of Embedding Extraction Layer}\label{sec:layer_ablation}
We conduct an ablation study comparing embeddings extracted from multiple decoder layers to assess sensitivity to the extraction point. The qualitative separation between low- and high-heterogeneity regimes persists across all layers, while the penultimate layer yields slightly larger performance gaps under severe heterogeneity. Full results are reported in Appendix~G.

\subsubsection{Effect of Training Backbone on Embedding-Based Heterogeneity} \label{sec:backbone_ablation} 
We further assess robustness to architectural coupling by training the federated global model with a ResNet-50 backbone while retaining client partitions generated from task-supervised CLIP embeddings. The embedding-based heterogeneity effect persists despite this backbone change, indicating that the observed performance degradation is not an artifact of encoder alignment. Full results are reported in Appendix~H.

\subsubsection{Effect of the Number of Clusters $k$}
\label{sec:vary_k}
We further analyze the sensitivity of the proposed method to the number of clusters $k$ used during embedding-based partitioning. We vary $k$ over a meaningful range and observe that, for moderate to large values, the monotonic relationship between data heterogeneity and performance and the induced separation between low- and high-heterogeneity regimes are preserved. Detailed results and analysis are provided in Appendix~I.

\subsubsection{Qualitative Behavior of Client Data Partitions Under Varying $K$}
Although we use $k=16$ as the default number of embedding clusters, we additionally evaluate the effect of varying 
$k \in \{2,4,10,16,32\}$. As detailed in Appendix~K, increasing $k$ refines the granularity of embedding partitions without introducing drastic cluster imbalance or effectively alter client sample compositions. Across all tested values of $k$, the qualitative heterogeneity structure induced by the Dirichlet distribution is preserved, which in turn explains the observed stability of the monotonic $\alpha$-performance trends.

\section{Conclusion and Future Work}
\vspace{-.5mm}
\noindent  In this paper, we demonstrated that inducing data heterogeneity in FL via a label/class-based approach alone fails to fully encapsulate the data heterogeneity encountered in computer vision tasks beyond classification.  
To address this limitation, we extracted task-specific data embeddings and proposed a new perspective on data heterogeneity in FL:  \textit{embedding-based data heterogeneity}. By clustering data points based on their embeddings and distributing them among clients using the Dirichlet distribution, we introduce a more nuanced formulation of data heterogeneity in FL, particularly for non-classification vision tasks.
Our experiments validated the impact of this revised framework, providing new benchmark performance measures and offering various nuanced insights. 
Beyond these contributions, we highlighted several open research directions, paving the way for future advancements in FL under this revised notion of data heterogeneity.

Further, it is worth mentioning that 
 % while this work introduces a task-aware embedding-based emulation of data heterogeneity, an interesting avenue for future research is the exploration of alternative embedding generators, such as self-supervised models (e.g., DINO~\cite{caron2021emerging}, SimCLR~\cite{chen2020simple}). Along the same line, investigating how various embedding generators influence data heterogeneity modeling and FL robustness remains an important direction for future work. 
 % Further, 
 although our embedding extraction is performed centrally for benchmarking, consistent with standard Dirichlet-based emulation, future work may incorporate secure aggregation protocols~\cite{bonawitz2017practical} to enable clients to compute embeddings locally and share only aggregated or privatized statistics to the server to obtain global knowledge about data heterogeneity across the clients. Finally, while our current study treats the vision tasks as a flat collection, future extensions of our benchmark could incorporate hierarchical structure and uncertainty-aware evaluation, inspired by~\cite{barkan2023forecasting}. For example, related tasks (e.g., depth, surface normals, and reshading) could be grouped into higher-level families, allowing data heterogeneity trends to be analyzed not only per task but also at the level of task families.

% \section*{Acknowledgments}
% This should be a simple paragraph before the References to thank those individuals and institutions who have supported your work on this article.

% {\appendices
% \section*{Proof of the First Zonklar Equation}
% \section*{Proof of the Second Zonklar Equation}
% Appendix two text goes here.}

\bibliographystyle{IEEEtran}
\bibliography{main.bib}

\newpage
\begin{IEEEbiography}
[\vspace{-5mm}{\includegraphics[width=1in,height=1.25in,clip,keepaspectratio]{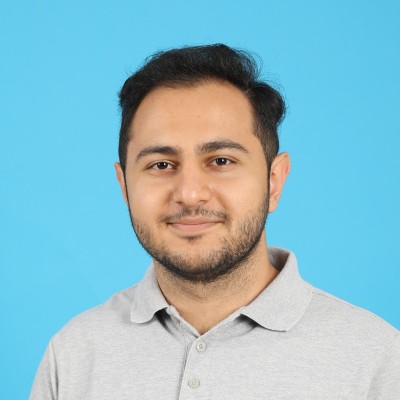}}]{Kasra Borazjani}[SM] received his B.Sc. degree in Electrical Engineering from University of Tehran, Tehran, Iran, in 2022. He has since been a Ph.D. student in Electrical Engineering at the University at Buffalo--SUNY. He was the recipient of the silver medal in the \textit{Iranian National Olympiad on Astronomy and Astrophysics} in 2016. His research interests include federated learning (FL), computer vision, and medical image processing, currently focusing on multi-modal and multi-task FL in computer vision applications.
\end{IEEEbiography}

\begin{IEEEbiography}
[{\includegraphics[width=1in,height=1.25in,clip,keepaspectratio]{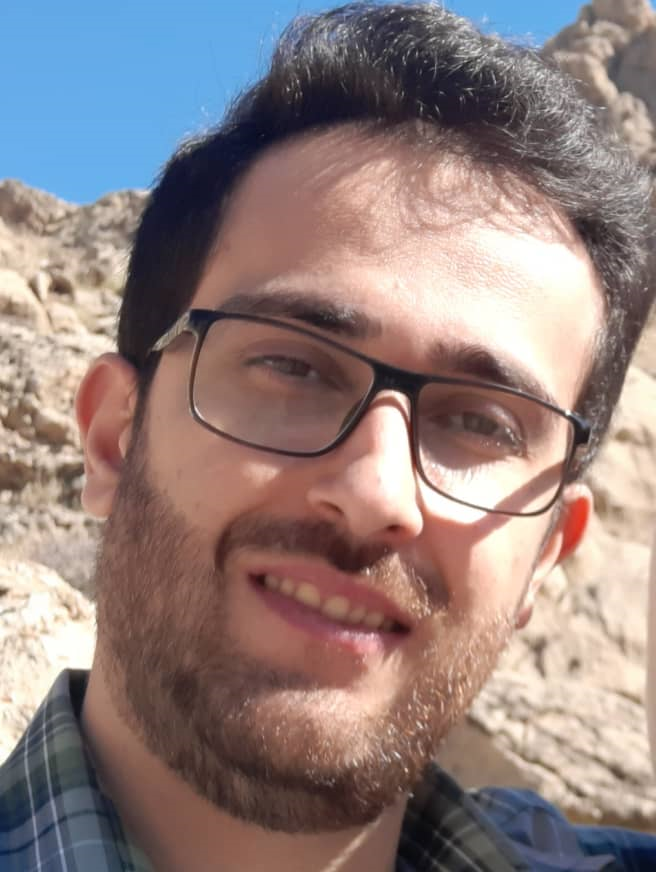}}]{Payam Abdisarabshali}[SM] received the M.Sc. degree in Computer Engineering from Razi University, Iran, with top-rank recognition in 2018. He was a teaching assistant professor at Razi University from 2018 to 2022. He is currently a Ph.D. student at the Department of Electrical Engineering, University at Buffalo--SUNY, NY, USA. His research interests include distributed and federated machine learning, computer vision, the design of neural network architectures, and mathematical modeling.  
\end{IEEEbiography}

\begin{IEEEbiography}
[{\vspace{-2.2mm}\includegraphics[width=1.05in,height=1.1in, trim= 1 1 1 1, clip]{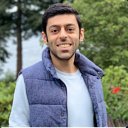}}]{Naji Khosravan}[IEEE/CVF Member] received his B.Sc. in Electrical Engineering from Amirkabir University of Technology (Tehran Polytechnique),  Tehran, Iran, in 2015. He received his M.Sc. and Ph.D. in Computer Science from the Center for Research In Computer Vision (CRCV) at the University of Central Florida, FL, USA in 2019. His research focus spans Human Centered AI, Computer Vision, Deep Learning and Medical Image Analysis. He has been an active contributor to IEEE CVF conferences in the form of workshop organizer and reviewer. He has been publishing his work in different venues such as CVPR, WACV, MICCAI, IEEE GLOBECOM, IEEE EMBC and ACM. He is currently a \textit{Senior Applied Science Manager} at Zillow group where his teams are  developing cutting edge AI solutions with an emphasis on the computer vision area.
\end{IEEEbiography}

\begin{IEEEbiography}
[{\includegraphics[width=1in,height=1.25in,clip,keepaspectratio]{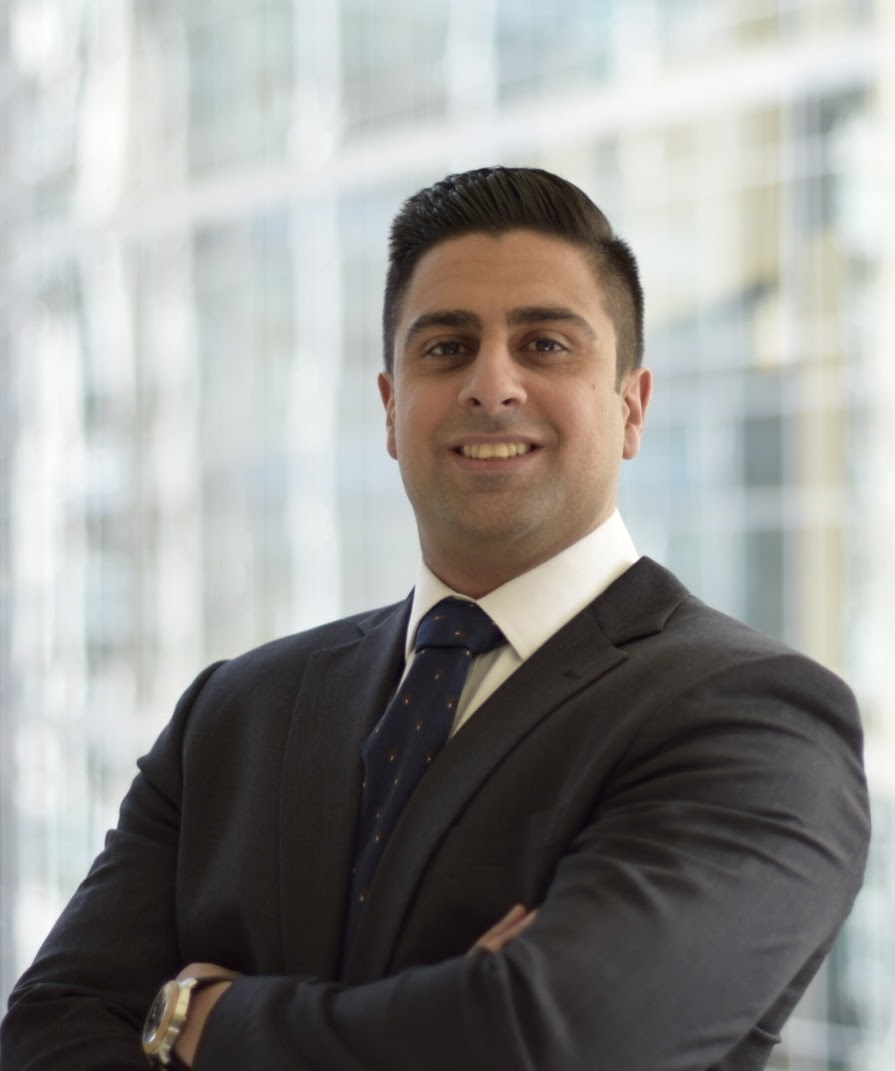}}]{Seyyedali Hosseinalipour}[M] received the B.S. degree in electrical engineering from Amirkabir University of Technology, Tehran, Iran, in 2015 with high honor and top-rank recognition. He then received the M.S. and Ph.D. degrees in electrical engineering from North Carolina State University, NC, USA, in 2017 and 2020, respectively; and was a postdoctoral researcher at Purdue University, IN, USA from 2020 to 2022. 
He is currently an assistant professor at the Department of Electrical Engineering, University at Buffalo-SUNY, NY, USA. 
He was the recipient of the \textit{ECE Doctoral Scholar of the Year Award} (2020) and \textit{ECE Distinguished Dissertation Award} (2021) at NC State University; and \textit{Students’ Choice Teaching Excellence Award} (2023) at University at Buffalo--SUNY. Furthermore, he was the first author of a paper published in IEEE/ACM Transactions on Networking that received the \textit{2024 IEEE Communications Society William Bennett Prize}.
He has served as the TPC Co-Chair of workshops/symposiums related to machine learning and edge computing for IEEE INFOCOM, GLOBECOM, ICC, CVPR, ICDCS, SPAWC, WiOpt, and VTC. He has also served as the guest editor of IEEE Internet of Things Magazine for the special issue on \textit{Federated Learning for Industrial Internet of Things} (2023). Since Feb. 2025, he has been serving as an \textit{Associate Editor} for the \textit{IEEE Transactions on Signal and Information Processing over Networks}.
His research interests include the analysis of
modern wireless networks, synergies between machine learning methods and fog/edge
computing systems, distributed/federated machine learning, and network optimization.
\end{IEEEbiography}

\appendices

\begin{table*}[t]
\vspace{-4mm}
\centering
\caption{{Final performance values (in terms of model loss) of all FL methods across seven Taskonomy vision tasks under three levels of embedding-based data heterogeneity (controlled by the Dirichlet parameter $\alpha$). Similar to the performance evaluations conducted in Figs. \ref{fig:embedding-vs-class-p1} and \ref{fig:fedprox-results}-\ref{fig:fedamp_results} of the paper, smaller loss values correspond to better model performance.}}
\label{tab:benchmark_results}
{\footnotesize
\begin{tblr}{
colspec ={Q[c,m,wd=42mm]  Q[c,m,wd=13mm] Q[c,m,wd=21mm] Q[c,m,wd=21mm] Q[c,m,wd=21mm] Q[c,m,wd=21mm] Q[c,m,wd=21mm]},
hlines = {black,0.1mm},
vlines = {black,0.1mm},
row{2} = {LightBlue},
cell{1}{1} = {r=2, c=1}{m, DarkBlue},
cell{1}{2} = {r=2, c=1}{m, DarkBlue},
cell{1}{3} = {r=1, c=5}{m, DarkBlue},
cell{3}{1} = {r=3, c=1}{m, white},
cell{6}{1} = {r=3, c=1}{m, white},
cell{9}{1} = {r=3, c=1}{m, white},
cell{12}{1} = {r=3, c=1}{m, white},
cell{15}{1} = {r=3, c=1}{m, white},
cell{18}{1} = {r=3, c=1}{m, white},
cell{21}{1} = {r=3, c=1}{m, white},
colsep=3pt, rowsep=3pt
                }
Vision Task & $\alpha$ & Method & & & &   \\
& & FedAvg & FedProx & SCAFFOLD & FedAmp & FedRep \\
\hline
2D Edges & 0.1 & 0.074 & 0.072 & 0.070 & 0.065 & 0.063\\
 & 10 & 0.073 & 0.071 & 0.069 & 0.063 & 0.059\\
 & 1000 & 0.071 & 0.070 & 0.064 & 0.060 & 0.054\\
 \hline
Reshading & 0.1 & 0.055 & 0.052 & 0.047 & 0.041& 0.040 \\
 & 10 & 0.052 & 0.049 & 0.042 & 0.038 & 0.038\\
 & 1000 & 0.049 & 0.045 & 0.040 & 0.036 & 0.035\\
 \hline
 Surface Normals & 0.1 & 0.192 & 0.183 & 0.174 & 0.169 & 0.168\\
 & 10 & 0.191 & 0.180 & 0.170 & 0.165 & 0.163\\
 & 1000 & 0.186 & 0.178 & 0.165 & 0.160 & 0.159\\
 \hline
Semantic Segmentation & 0.1 & 0.052 & 0.048 & 0.044 & 0.041 & 0.039\\
 & 10 & 0.048 & 0.046 & 0.041 & 0.037 & 0.036\\
 & 1000 & 0.047 & 0.044 & 0.040 & 0.034 & 0.034\\
 \hline
3D Keypoints & 0.1 & 0.089 & 0.084 & 0.079 & 0.074 & 0.075 \\
 & 10 & 0.085 & 0.080 & 0.076 & 0.071 & 0.071 \\
 & 1000 & 0.083 & 0.077 & 0.073 & 0.070 & 0.069 \\
 \hline
Euclidean Depth Estimation & 0.1 & 0.074 & 0.073 & 0.070 & 0.069 & 0.067\\
 & 10 & 0.073 & 0.071 & 0.068 & 0.064 & 0.062\\
 & 1000 & 0.071 & 0.069 & 0.064 & 0.061 & 0.058\\
 \hline
Scene Classification & 0.1 & 2.31 & 2.29 & 2.17 & 1.93 & 1.78 \\
 & 10 & 1.81 & 1.70 & 1.52 & 1.56 & 1.53 \\
 & 1000 & 1.43 & 1.36 & 1.30 & 1.30 & 1.32 \\
\end{tblr}
}
\vspace{1mm}
\end{table*}

\newpage
\section{Performance Summary of FL Algorithms}\label{App:Glance}

Table~\ref{tab:benchmark_results} summarizes the converged performance of FedAvg, FedProx, FedRep, FedAMP, and SCAFFOLD under the embedding-based data heterogeneity levels reported in Figs. \ref{fig:embedding-vs-class-p1} and \ref{fig:fedprox-results}-\ref{fig:fedamp_results} of the paper. {Overall, the performance ranks from best to worst as: FedRep, FedAMP, SCAFFOLD, FedProx, and FedAvg.} Further, the key takeaway is consistent across all algorithms: each method exhibits a clear decline in performance as the embedding-based heterogeneity level increases, confirming that they are all susceptible to our revamped heterogeneity formulation. Notably, this monotonic performance degradation is not captured under conventional label-based heterogeneity, as evidenced by the flat or inconsistent trends observed in Figs. \ref{fig:embedding-vs-class-p1} and \ref{fig:fedprox-results}-\ref{fig:fedamp_results}. This reinforces the central motivation of our work -- namely, that class-based heterogeneity fails to impactfully hinder the training of FL methods in computer vision tasks, while our embedding-based heterogeneity can do so.

\begin{figure*}[t]
    \centering
    \begin{tcolorbox}[colback=BoxBg, colframe=BoxFrame, width=0.49\textwidth, left=0pt, right=0pt, top=-2pt, bottom=-2pt, after=\hspace{2mm}, title=Edge Detection --- \textbf{FedAvg, PASCAL}, halign title=flush center,toptitle=-3pt,
    bottomtitle=-3pt]
    \includegraphics[width=0.48\textwidth]{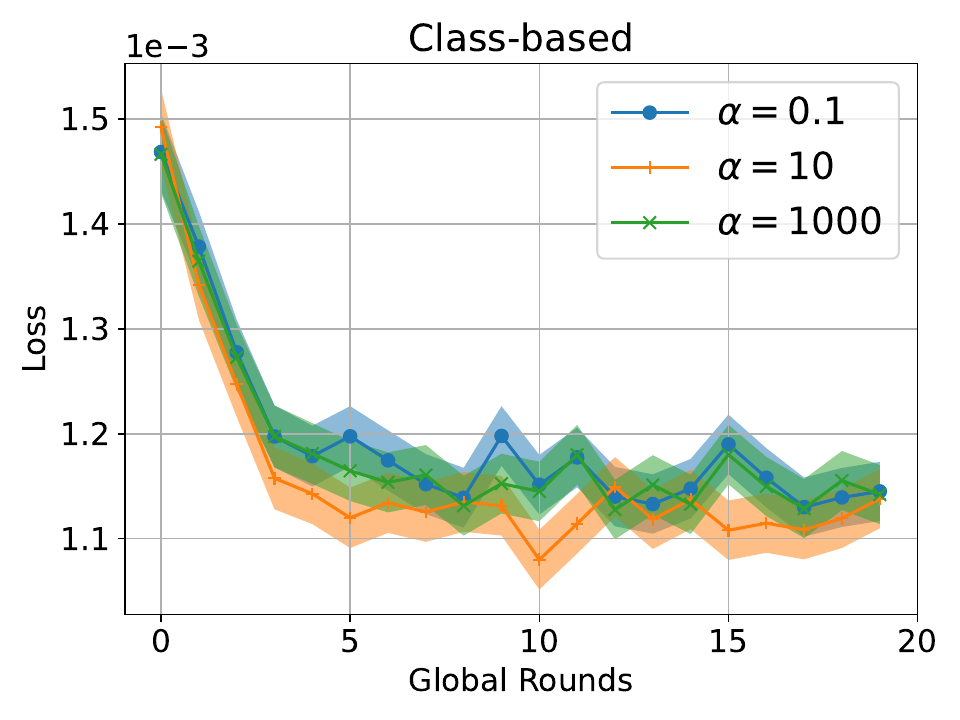}
    \includegraphics[width=0.48\textwidth]{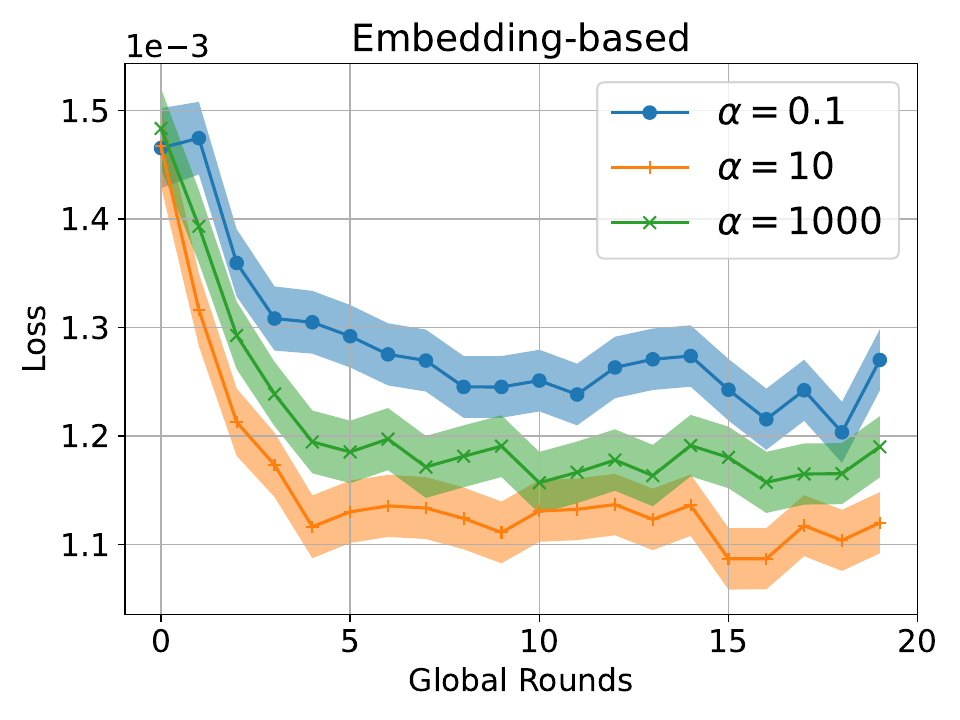}
    \end{tcolorbox}
    \hfill
    \begin{tcolorbox}[colback=BoxBg, colframe=BoxFrame, width=0.49\textwidth, left=0pt, right=0pt, top=-2pt, bottom=-2pt, before=, title=Surface Normals --- \textbf{FedAvg, PASCAL}, halign title=flush center,toptitle=-3pt,
    bottomtitle=-3pt]
    \includegraphics[width=0.48\textwidth]{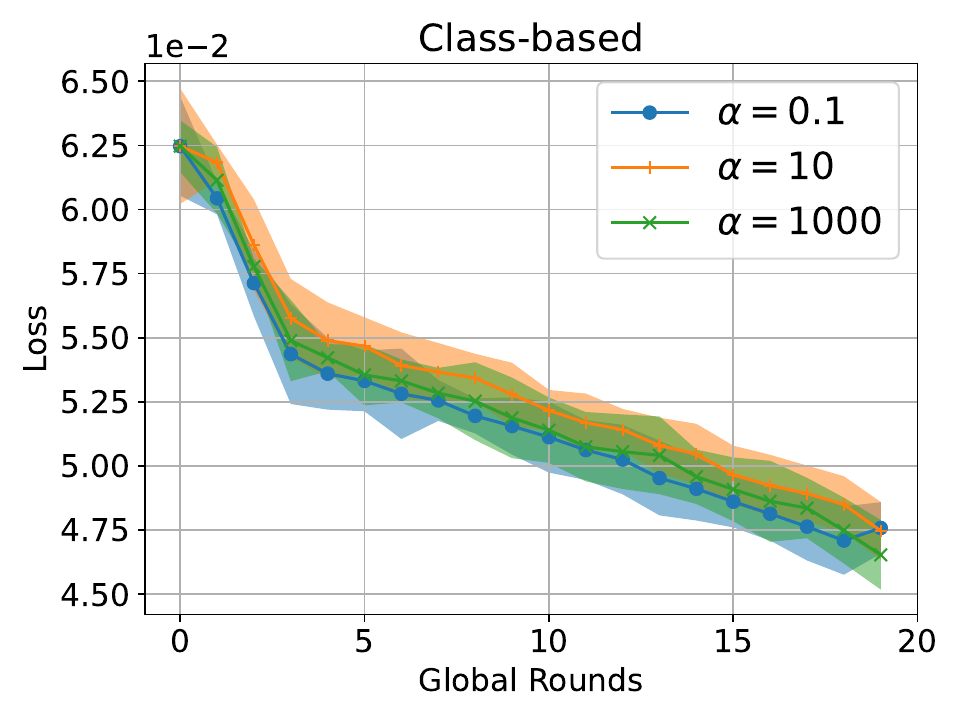}
    \includegraphics[width=0.48\textwidth]{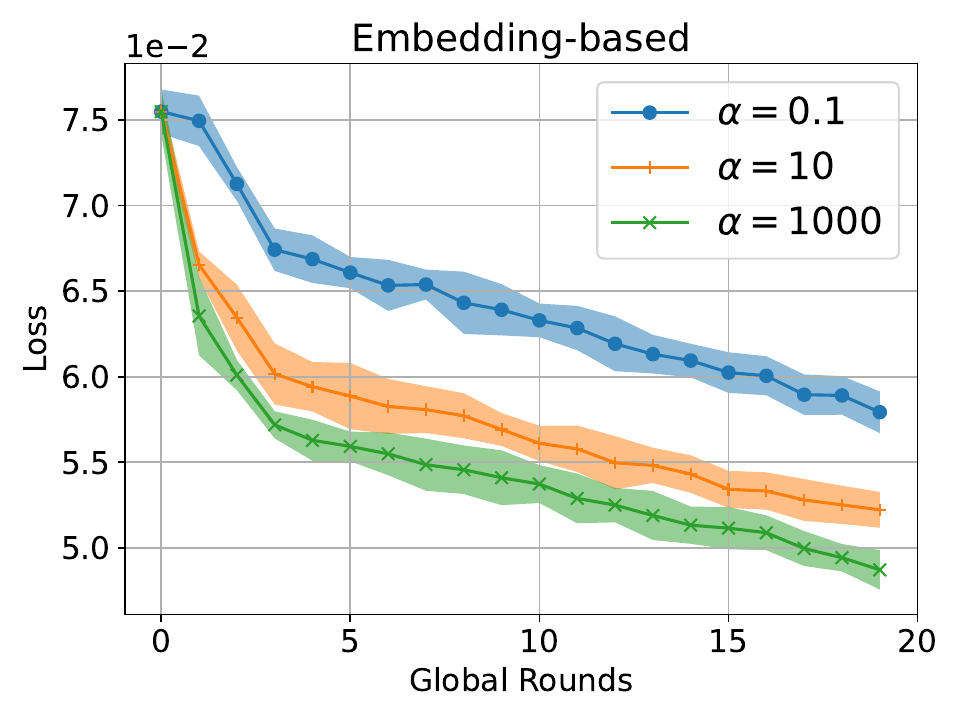}
    \end{tcolorbox}
    \vspace{-3.6mm}
     
    \begin{tcolorbox}[colback=BoxBg, colframe=BoxFrame, width=0.49\textwidth, left=0pt, right=0pt, top=-2pt, bottom=-2pt, after=\hspace{2mm}, title=Saliency Detection --- \textbf{FedAvg, PASCAL}, halign title=flush center,toptitle=-3pt,
    bottomtitle=-3pt]
    \includegraphics[width=0.49\textwidth]{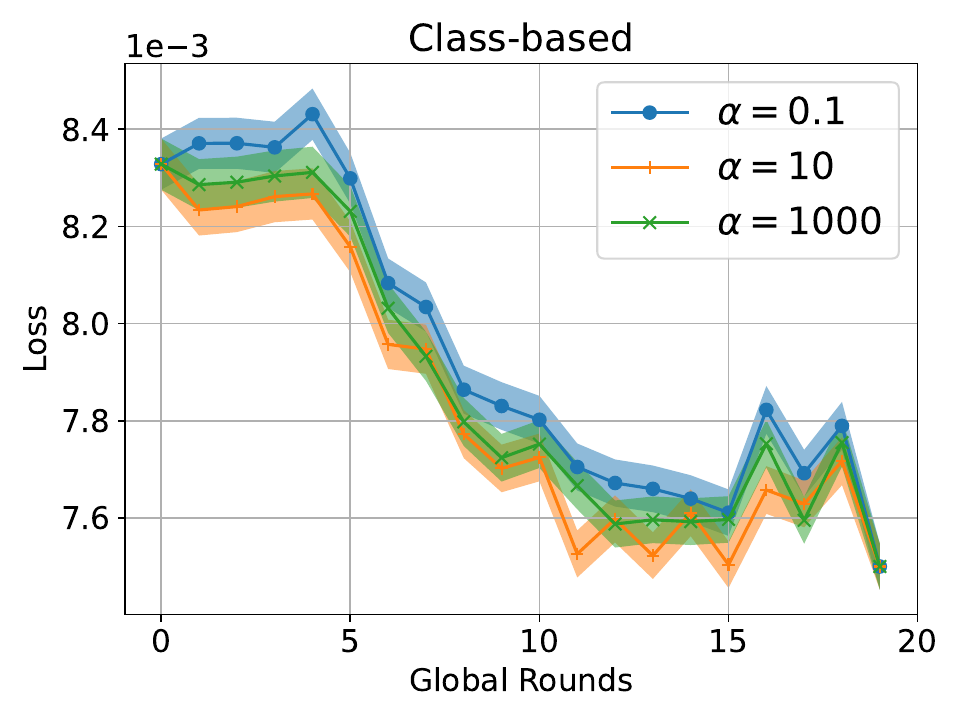}
    \includegraphics[width=0.49\textwidth]{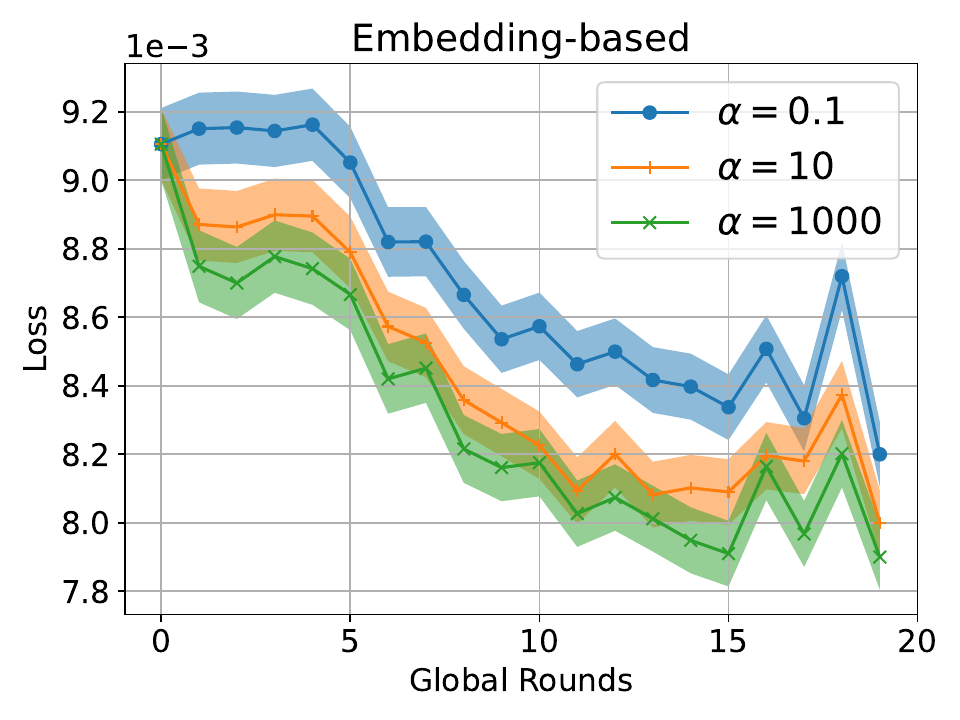}
    \end{tcolorbox}
    \hfill
    \begin{tcolorbox}[colback=BoxBg, colframe=BoxFrame, width=0.49\textwidth, left=0pt, right=0pt, top=-2pt, bottom=-2pt, before=, title=Semantic Segmentation --- \textbf{FedAvg, PASCAL}, halign title=flush center,toptitle=-3pt,
    bottomtitle=-3pt]
    \includegraphics[width=0.49\textwidth]{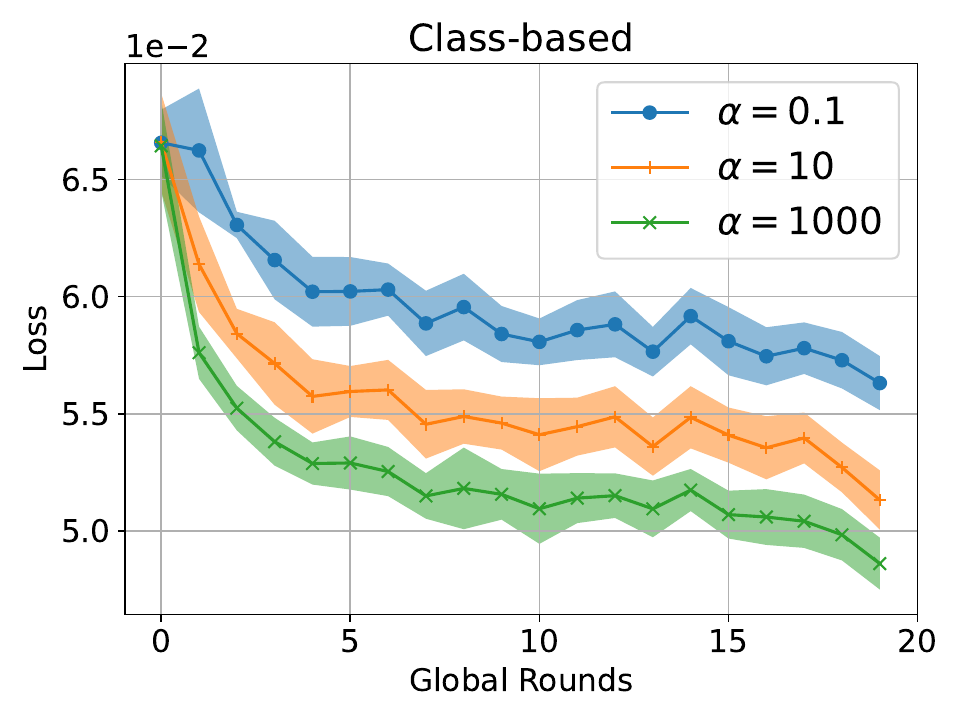}
    \includegraphics[width=0.49\textwidth]{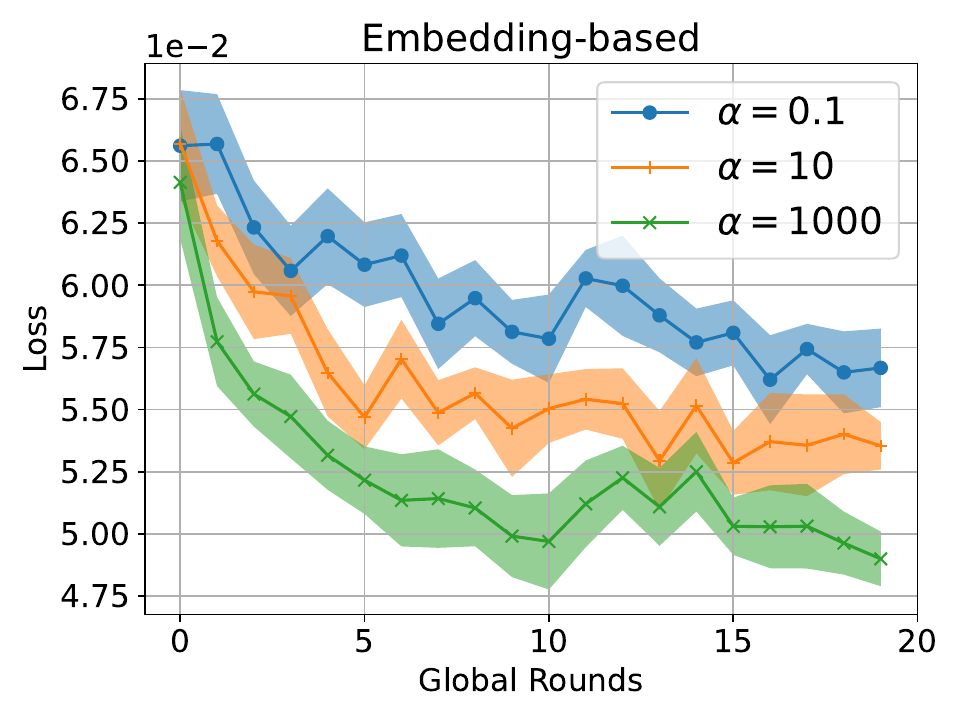}
    \end{tcolorbox}
 \vspace{-3.6mm}
    \caption{\textbf{Class-based vs. Embedding-based distribution:} Comparison of how performing Dirichlet distribution over the datapoints' labels (equivalently, the primary object in the PASCAL dataset) and the extracted embeddings affect the performance in FL. Consistent with our findings on the Taskonomy dataset, the embedding-based data partitioning yields  performance gaps across data heterogeneity levels, whereas class-based data partitioning produces minimal or unstable performance variations.} 
    \label{fig:pascal_results}
    \vspace{-3.5mm}
\end{figure*}

\section{Cluster Validity Index Results}\label{App:CVI}
\begin{figure}[H]
    \centering
    \includegraphics[width=\linewidth]{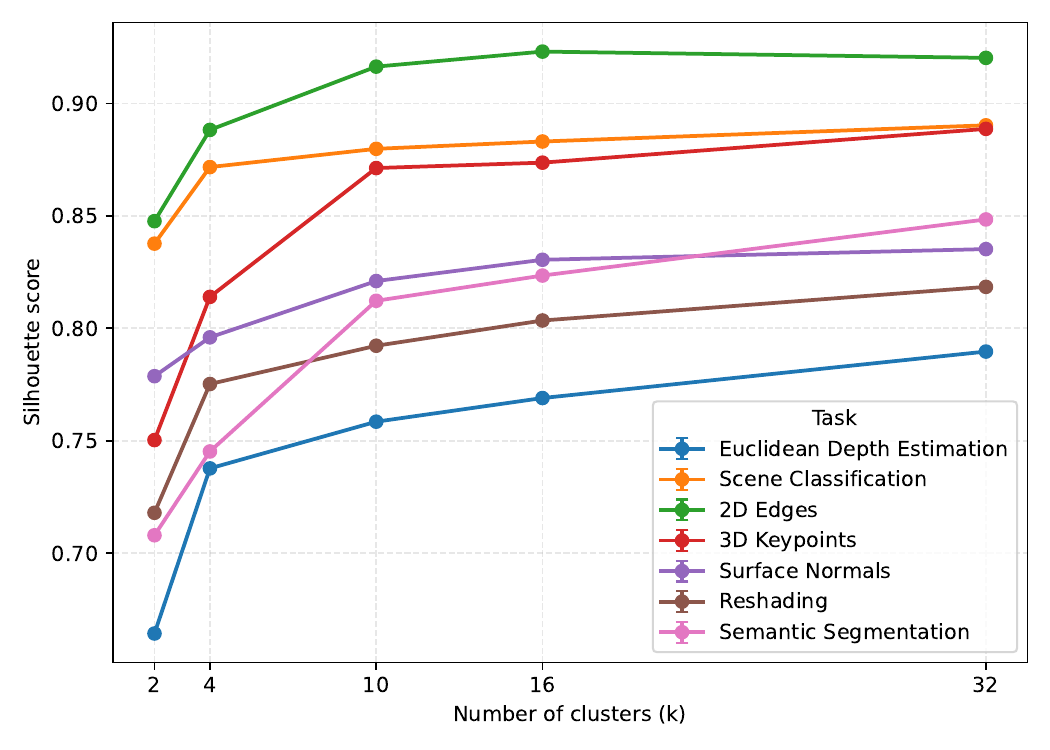}
    \caption{Silhouette coefficient values for the evaluated choices of $k\in\{2,4,10,16,32\}$ across the seven Taskonomy tasks.}
    \label{Fig:Silhouette}
\end{figure}

 Fig.~\ref{Fig:Silhouette} reports the  Silhouette coefficient values for all evaluated choices of $k\in\{2,4,10,16,32\}$ across the seven Taskonomy tasks. The Silhouette scores exhibit an improvement as $k$ increases, but the gains become marginal beyond $k=16$.
 
\section{Results of the PASCAL Dataset}\label{App:PASCAL}

\subsection{Data Preprocessing}

The PASCAL dataset includes four tasks: Semantic Segmentation, Edge Detection, Saliency Detection, and Surface Normals. To construct a class-based setting analogous to Taskonomy, we assign each image a \textit{primary object} defined as the semantic class occupying the largest pixel area. Images are then partitioned across clients based on this primary label in class-based data heterogeneity. Because these labels originate from the semantic segmentation task, this task serves as the analogue of scene classification in Taskonomy when comparing class-based versus embedding-based heterogeneity.

Similar to the Taskonomy dataset, when performing the federated training, we distribute the data across the clients based on the Dirichlet distribution of the embeddings extracted, where $10\%$ of local data points are used for validation. For global model validation, we report the average loss of the global model across the clients on their validation data. For the centralized training, we use the early stopping method to halt training if no improvement in loss was observed after $5$ epochs. For the federated training, we train all methods for $20$ global aggregations, each comprising $100$ SGD rounds with the mini-batch size of $16$. In addition, we sample all clients during each global aggregation round as our method is not focused on client sampling.

\subsection{Results}
Fig.~\ref{fig:pascal_results} illustrates the FL model performance under varying data heterogeneity levels for both class-based and embedding-based data splits. The results mirror the trends observed in Taskonomy: class-based data splits produce nearly overlapping loss curves across all 
$\alpha$ values, indicating that label skew does not meaningfully disrupt the features relevant to the vision tasks considered. In contrast, embedding-based splits consistently generate pronounced performance degradation as  $\alpha$ decreases, revealing the model’s sensitivity to task-driven variations in the data distribution.

\section{Results of the Task-Specific Metrics}\label{App:Task-Metric}
In this appendix, we complement the main-text analysis by reporting results under a set of task-specific evaluation metrics that are commonly used in the computer vision literature. Specifically, we evaluate: F-measure for 2D Edges, PSNR for Reshading, RMSE for Euclidean Depth Estimation, mIoU for Semantic Segmentation, and Accuracy for Scene Classification. The complete results are summarized in Table~\ref{tab:task_metrics_reformatted}. As shown in the table, the qualitative trend remains consistent with our main findings: embedding-based data heterogeneity induces more clear performance gaps across data heterogeneity levels, while class-based data splits continue to exhibit negligible performance variations for the vision tasks (except for Scene Classification task, as expected).

\begin{table}[!ht]
\centering
\caption{Task-specific metrics for embedding-based vs.\ class-based data distributions across different $\alpha$ values. 
Arrows indicate the desirable direction for each metric: $\downarrow$ denotes that lower values correspond to better performance, while $\uparrow$ denotes that higher values correspond to better performance.}
\label{tab:task_metrics_reformatted}
{\footnotesize
\begin{tblr}{
  colspec = {
    Q[c,m,wd=27mm]   % Task
    Q[c,m,wd=7mm]   % alpha
    Q[c,m,wd=25mm]   % embedding-based
    Q[c,m,wd=20mm]   % class-based
  },
  hlines = {black,0.1mm},
  vlines = {black,0.1mm},
  row{1} = {DarkBlue},
  cell{2}{1} = {r=3, c=1}{m, white},
  cell{5}{1} = {r=3, c=1}{m, white},
  cell{8}{1} = {r=3, c=1}{m, white},
  cell{11}{1} = {r=3, c=1}{m, white},
  cell{14}{1} = {r=3, c=1}{m, white},
  colsep=3pt, rowsep=3pt
}
\textbf{Task} & \textbf{$\alpha$} & \textbf{Embedding-based} & \textbf{Class-based} \\
\hline

% ----------------- 2D Edges ------------------
\SetCell[r=3]{c, bg=LightBlue} \textbf{2D Edges\\(F-measure~$\uparrow$)} 
  & 0.1   & 0.142 & 0.155 \\
& 10    & 0.146 & 0.156 \\
& 1000  & 0.152 & 0.156 \\
\hline

% ----------------- Reshading ------------------
\SetCell[r=3]{c, bg=LightBlue} \textbf{Reshading\\(PSNR (dB) $\uparrow$)} 
  & 0.1 & 13.41 & 14.10 \\
& 10 & 13.82 & 14.10 \\
& 1000 & 14.03 & 14.13 \\
\hline

% ----------------- Semantic Segmentation ------------------
\SetCell[r=3]{c, bg=LightBlue} \textbf{Semantic Segmentation\\(mIoU (\%)~$\uparrow$)} 
  & 0.1   & 36.48 & 46.14 \\
& 10    & 44.39 & 47.44 \\
& 1000  & 45.37 & 47.92 \\
\hline

% ----------------- Euclidean Depth ------------------
\SetCell[r=3]{c, bg=LightBlue} \textbf{Euclidean Depth \\(RMSE~$\downarrow$)} 
  & 0.1   & 0.274 & 0.265 \\
& 10    & 0.272 & 0.265 \\
& 1000  & 0.266 & 0.265 \\
\hline

% ----------------- Scene Classification ------------------
\SetCell[r=3]{c, bg= v} \textbf{Scene Classification\\(Accuracy (\%) $\uparrow$)} 
  & 0.1   & 52.45 & 54.68 \\
& 10    & 60.32 & 61.13 \\
& 1000  & 67.24 & 69.34 \\
\end{tblr}}
\end{table}

\section{Simulation Hyperparameters}\label{App:HyperS}
We summarize all hyperparameters and training protocol details used across our experiments in Table~\ref{tab:training_protocol}.

\begin{table}[h]
\centering
\caption{Hyperparameters and training protocol used in all experiments.}
\label{tab:training_protocol}
{\footnotesize
\begin{tblr}{
colspec = {Q[l,m,wd=35mm] Q[l,m,wd=45mm]},
hlines = {black,0.1mm},
vlines = {black,0.1mm},
row{1} = {DarkBlue},
colsep=3pt, rowsep=3pt
}
\hspace{11mm}\textbf{Parameter} & \hspace{11mm}\textbf{Value / Setting} \\
\hline
Client sampling rate & All clients participate in each round \\
Number of clients & 25 \\
Number of global rounds & 20 \\
Local SGD iterations per round & 100 \\
Mini-batch size & 16 \\
Optimizer & SGD \\
Learning-rate (LR) schedule & Initial LR: $1\times10^{-4}$; exponential decay $\gamma = 0.99$ \\
Early stopping (centralized pre-training) & Patience of 5 epochs \\
Local validation split & 10\% of each client’s local data \\
Global validation set & 100 uniformly sampled datapoints \\
Data augmentations & None used \\
\end{tblr}
}
\end{table}

\section{Studying other Embedding Extraction Methods} \label{app:pretraining_ablation}
As discussed in Sec.~\ref{sec:method}, our objective is to induce data heterogeneity from the unique perspective of each vision task. Since different tasks attend to different attributes of the same image, inducing task-perspective data heterogeneity inherently requires representations that encode task-relevant structure. Consequently, representations that are generic or task-agnostic are not expected to provide a meaningful basis for constructing task-perspective heterogeneous data distributions.
To empirically examine this premise, we study how the extent of the data heterogeneity-induced performance gap varies under different embedding extraction strategies. In particular, we compare three settings:
\begin{enumerate}
    \item \textbf{No Pre-training}: embeddings are extracted from a frozen CLIP vision encoder backbone,
    \item \textbf{Self-supervised Pre-training}: embeddings are extracted from a SimCLR model~\cite{chen2020simple} pre-trained on the Taskonomy dataset,
    \item \textbf{Task-supervised Pre-training (our method)}: embeddings are extracted from a CLIP model fine-tuned on the Taskonomy dataset for each corresponding task.
\end{enumerate}

The results for all tasks and data heterogeneity levels are reported in Table~\ref{tab:PreTrainingAblationResults}. Focusing on the Surface Normals task, we observe that under the frozen CLIP embedding extraction method, the trained global models across different data heterogeneity levels do not exhibit a monotonic change in performance as heterogeneity increases (i.e., when $\alpha$ decreases). A similar behavior is observed for the 3D Keypoints task. In other cases where performance worsens (i.e., the loss increases) with increasing data heterogeneity (i.e., when $\alpha$ decreases), such as the Scene Classification task, the resulting gap between the worst and best performance ($1.54-1.44=0.10$) remains substantially smaller than the gap induced under our proposed task-supervised pre-training (right column, $2.31-1.43=0.88$).
Inspecting the results obtained using embeddings extracted from SimCLR reveals a similar pattern to that of frozen CLIP, indicating that self-supervised pre-training alone does not ensure representations that are sufficiently aligned with the target tasks to induce meaningful performance gap.
In contrast, embeddings obtained from our proposed task-supervised pre-training consistently yield higher loss values under severe data heterogeneity ($\alpha = 0.1$) across all tasks and a clearer separation between low- and high-heterogeneity regimes. These observations indicate that task supervision is a necessary ingredient for inducing task-perspective data heterogeneity during the simulation phase of FL methods, as it yields representations that capture the variations most relevant to the task.

Taken together, these results support our central claim that meaningful data heterogeneity for non-classification vision tasks must be defined relative to task-aligned representations. Generic embeddings that lack task supervision are therefore expected to be less effective in revealing performance sensitivity of FL to data heterogeneity.

\begin{table}[!ht]
% \vspace{-4mm}
\centering
\caption{Comparison of the performance gaps reached through embeddings extracted using various pre-training scenarios based on the $20$-th global aggregation round loss value trained using FedAvg. The higher loss value achieved through our method (task-supervised pre-training) and the inconsistency in loss trend in the two other embedding extraction methods (frozen CLIP and SimCLR) show that the task-relative pre-training of the embedding extraction can lead to a better performance gap across different levels of data heterogeneity.}
\label{tab:PreTrainingAblationResults}
{\footnotesize
\begin{tblr}{
colspec ={Q[c,m,wd=25mm]  Q[c,m,wd=7mm] Q[c,m,wd=14mm] Q[c,m,wd=14mm] Q[c,m,wd=14mm]},
hlines = {black,0.1mm},
vlines = {black,0.1mm},
row{2} = {LightBlue},
cell{1}{1} = {r=2, c=1}{m, DarkBlue},
cell{1}{2} = {r=2, c=1}{m, DarkBlue},
cell{1}{3} = {r=1, c=3}{m, DarkBlue},
cell{3}{1} = {r=3, c=1}{m, white},
cell{6}{1} = {r=3, c=1}{m, white},
cell{9}{1} = {r=3, c=1}{m, white},
cell{12}{1} = {r=3, c=1}{m, white},
cell{15}{1} = {r=3, c=1}{m, white},
cell{18}{1} = {r=3, c=1}{m, white},
cell{21}{1} = {r=3, c=1}{m, white},
colsep=3pt, rowsep=3pt
                }
Vision Task & $\alpha$ & Method &  \\
& & Frozen CLIP \hspace{4mm}(No Pre-Training) & SimCLR (Self-supervised Pre-Training) & Our Method (Task-supervised Pre-Training) \\
\hline
2D Edges & 0.1 & 0.072 & 0.073 & 0.074 \\
 & 10 & 0.071 & 0.071 & 0.073 \\
 & 1000 & 0.071 & 0.073 & 0.071 \\
 \hline
Reshading & 0.1 & 0.048 & 0.049 & 0.055 \\
 & 10 & 0.047 & 0.049 & 0.052 \\
 & 1000 & 0.046 & 0.047 & 0.049 \\
 \hline
 Surface Normals & 0.1 & 0.189 & 0.185 & 0.192 \\
 & 10 & 0.187 & 0.186 &  0.191 \\
 & 1000 & 0.188 & 0.185 & 0.186 \\
 \hline
Semantic Segmentation & 0.1 & 0.047 & 0.049 & 0.052 \\
 & 10 & 0.046 & 0.048 & 0.048 \\
 & 1000 & 0.046 & 0.046 & 0.047 \\
 \hline
3D Keypoints & 0.1 & 0.084 & 0.085 & 0.089 \\
 & 10 & 0.086 & 0.085 &  0.085 \\
 & 1000 & 0.082 & 0.084 & 0.083 \\
 \hline
Euclidean Depth Estimation & 0.1 & 0.072 & 0.072 & 0.074 \\
 & 10 & 0.072 & 0.071 &  0.073 \\
 & 1000 & 0.070 & 0.071 & 0.071 \\
 \hline
Scene Classification & 0.1 & 1.54 & 1.48 & 2.31 \\
 & 10 & 1.52 & 1.50 & 1.81 \\
 & 1000 & 1.44 & 1.44 & 1.43 \\
\end{tblr}
}
\end{table}

\section{Studying the Effect of Embedding Extraction Layer} \label{app:layer_ablation}

To evaluate whether the observed data heterogeneity effects depend on the specific layer from which embeddings are extracted, we compare embeddings obtained from three different locations within the task-trained decoder: \textit{(i) the decoder input} (i.e., encoder output, corresponding to the extracted features), \textit{(ii) an intermediate decoder layer}, and \textit{(iii) the decoder penultimate layer} used in our default setting.
The results across all tasks and data heterogeneity levels are reported in Table~\ref{tab:EmbeddingLayerResults}. Across tasks, embeddings extracted from all three layers produce a consistent monotonic relationship between data heterogeneity level and performance, where increasing heterogeneity (smaller $\alpha$) leads to higher loss. For example, in the Surface Normals task, the loss decreases as $\alpha$ increases from $0.1$ to $10$ to $1000$ for embeddings extracted from the input layer ($0.190 \rightarrow 0.188 \rightarrow 0.185$), middle layer ($0.190 \rightarrow 0.188 \rightarrow 0.186$), and penultimate layer ($0.192 \rightarrow 0.191 \rightarrow 0.186$).
Nevertheless, while this qualitative behavior is preserved across all layers, embeddings extracted from the penultimate layer consistently yield the worst performance (i.e., the highest loss) under severe data heterogeneity. For instance, in the Surface Normals task, at $\alpha=0.1$, the loss under the penultimate layer is $0.192$ compared to $0.190$ for the input and middle layers, and a similar trend holds at $\alpha=10$ ($0.191$ versus $0.188$ and $0.187$). Also, under  low  data heterogeneity ($\alpha=1000$), the converged loss values are nearly identical across layers (e.g., $0.185$ and $0.186$), indicating that the embedding extraction layer does not materially affect performance in near-IID regimes, which is an expected outcome. 
Similar patterns are observed across other tasks in Table~\ref{tab:EmbeddingLayerResults}. 

These results indicate that the qualitative conclusions of our benchmark (i.e., embedding-based data heterogeneity induces a clear separation between low and high data heterogeneity regimes) are robust to the choice of embedding extraction layer, while the penultimate layer provides slightly more pronounced performance gaps and is therefore adopted in the main simulations of our paper.

\begin{table}[!h]
% \vspace{-4mm}
\centering
\caption{Comparison of the performance gaps reached through embeddings extracted from various parts of the model based on the $20$-th global aggregation round loss value trained using FedAvg. Although an increasing trend in loss with increasing data heterogeneity level (i.e., smaller $\alpha$ values) is observed throughout different layers for embedding extraction, the higher loss values are reached using the embeddings of the penultimate layer that better captures task-specific features.}
\label{tab:EmbeddingLayerResults}
{\footnotesize
\begin{tblr}{
colspec ={Q[c,m,wd=24mm]  Q[c,m,wd=7mm] Q[c,m,wd=14mm] Q[c,m,wd=14mm] Q[c,m,wd=14mm]},
hlines = {black,0.1mm},
vlines = {black,0.1mm},
row{2} = {LightBlue},
cell{1}{1} = {r=2, c=1}{m, DarkBlue},
cell{1}{2} = {r=2, c=1}{m, DarkBlue},
cell{1}{3} = {r=1, c=3}{m, DarkBlue},
cell{3}{1} = {r=3, c=1}{m, white},
cell{6}{1} = {r=3, c=1}{m, white},
cell{9}{1} = {r=3, c=1}{m, white},
cell{12}{1} = {r=3, c=1}{m, white},
cell{15}{1} = {r=3, c=1}{m, white},
cell{18}{1} = {r=3, c=1}{m, white},
cell{21}{1} = {r=3, c=1}{m, white},
colsep=3pt, rowsep=3pt
                }
Vision Task & $\alpha$ & Embedding Extraction Layer & &  \\
& & Decoder Input (Backbone Extracted Features) & Middle Decoder Block & Penultimate Layer (Proposed) \\
\hline
2D Edges & 0.1 & 0.072 & 0.073 & 0.074 \\
 & 10 & 0.071 & 0.072 & 0.073 \\
 & 1000 & 0.070 & 0.070 & 0.071 \\
 \hline
Reshading & 0.1 & 0.053 & 0.054 & 0.055 \\
 & 10 & 0.051 & 0.051 & 0.052 \\
 & 1000 & 0.048 & 0.047 & 0.049 \\
 \hline
 Surface Normals & 0.1 & 0.190 & 0.190 & 0.192 \\
 & 10 & 0.187 & 0.188 & 0.191 \\
 & 1000 & 0.185 & 0.186 & 0.186 \\
 \hline
Semantic Segmentation & 0.1 & 0.050 & 0.050 & 0.052 \\
 & 10 & 0.047 & 0.048 & 0.048 \\
 & 1000 & 0.046 & 0.046 & 0.047 \\
 \hline
3D Keypoints & 0.1 & 0.086 & 0.088 & 0.089 \\
 & 10 & 0.084 & 0.085 & 0.085 \\
 & 1000 & 0.083 & 0.082 & 0.083 \\
 \hline
Euclidean Depth Estimation & 0.1 & 0.073 & 0.074 & 0.074 \\
 & 10 & 0.073 & 0.073 & 0.073 \\
 & 1000 & 0.071 & 0.072 & 0.071 \\
 \hline
Scene Classification & 0.1 & 2.08 & 2.15 & 2.31 \\
 & 10 & 1.45 & 1.73 & 1.81 \\
 & 1000 & 1.34 & 1.39 & 1.43 \\
\end{tblr}
}
\end{table}

\section{Effect of Training the Backbone on Embedding-Based Heterogeneity} \label{app:backbone_ablation}
We further examine the effect of coupling between the encoder architectures of the embedding generator and the trained global model during FL. To this end, we replace the CLIP vision encoder with a ResNet-50 backbone~\cite{he2016deep} and conduct FL while keeping the client data partitions identical to those generated using task-supervised CLIP embeddings.
The results are presented in Table~\ref{tab:CNNResults} for all tasks across data heterogeneity levels under both class-based and embedding-based data heterogeneity. Focusing on the Reshading task, the embedding-based data heterogeneity (second column) continues to induce a monotonic increase in the converged loss as data heterogeneity increases (i.e., as $\alpha$ decreases), even though the federated model (i.e., ResNet-50) backbone differs from the embedding generator (i.e., CLIP). In contrast, class-based data heterogeneity under the ResNet backbone (first column) exhibits minimal performance separation across $\alpha$ values. This phenomenon closely resembles the difference in behavior between the scenarios where class-based and the embedding-based data heterogeneity were induced and a CLIP model was used as the backbone (third and fourth columns, respectively).
These results indicate that the performance degradation induced by embedding-based heterogeneity is not entirely dependent on the backbone alignment between the embedding generator and the federated trained model. Instead, the effect persists under architectural changes, suggesting that the induced data partitions capture task-relevant structure that generalizes beyond a specific model.

\begin{table}[!h]
% \vspace{-4mm}
\centering
\caption{Comparison of the performance gaps reached through training the federated models using a ResNet-50 encoder and a CLIP encoder under class-based (Cl.) and embedding-based (Em.) data heterogeneity based on the $20$-th global aggregation round loss value trained using FedAvg. Upon using embedding-based data heterogeneity (i.e., our method), similar loss trends can be observed when ResNet encoder is used compared to when CLIP encoder is used. Further, class-based data heterogeneity fails to affect the performance for both models.}
\label{tab:CNNResults}
{\footnotesize
\begin{tblr}{
colspec ={Q[c,m,wd=24mm]  Q[c,m,wd=7mm] Q[c,m,wd=12.5mm] Q[c,m,wd=12.5mm] Q[c,m,wd=10mm] Q[c,m,wd=10mm]},
hlines = {black,0.1mm},
vlines = {black,0.1mm},
row{2} = {LightBlue},
cell{1}{1} = {r=2, c=1}{m, DarkBlue},
cell{1}{2} = {r=2, c=1}{m, DarkBlue},
cell{1}{3} = {r=1, c=4}{m, DarkBlue},
cell{3}{1} = {r=3, c=1}{m, white},
cell{6}{1} = {r=3, c=1}{m, white},
cell{9}{1} = {r=3, c=1}{m, white},
cell{12}{1} = {r=3, c=1}{m, white},
cell{15}{1} = {r=3, c=1}{m, white},
cell{18}{1} = {r=3, c=1}{m, white},
cell{21}{1} = {r=3, c=1}{m, white},
colsep=3pt, rowsep=3pt
                }
Vision Task & $\alpha$ & Backbone & & & \\
& & ResNet-50 (Cl.) & ResNet-50 (Em.) & CLIP (Cl.) & CLIP (Em.) \\
\hline
2D Edges & 0.1 & 0.083 & 0.098 & 0.072 & 0.074 \\
 & 10 & 0.082 & 0.094 & 0.072 & 0.073 \\
 & 1000 & 0.082 & 0.087 & 0.072 & 0.071 \\
 \hline
Reshading & 0.1 & 0.063 & 0.071 & 0.048 & 0.055 \\
 & 10 & 0.062 & 0.068 & 0.048 & 0.052 \\
 & 1000 & 0.063 & 0.063 & 0.048 & 0.049 \\
 \hline
 Surface Normals & 0.1 & 0.208  & 0.231 & 0.190 & 0.192 \\
 & 10 & 0.207 & 0.215 & 0.190 & 0.191 \\
 & 1000 & 0.208 & 0.209 & 0.190 & 0.186 \\
 \hline
Semantic Segmentation & 0.1 & 0.062 & 0.072 & 0.047 & 0.052 \\
 & 10 & 0.064 & 0.068 & 0.047 & 0.048 \\
 & 1000 & 0.063 & 0.063 & 0.047 & 0.047 \\
 \hline
3D Keypoints & 0.1 & 0.089 & 0.102 & 0.082 & 0.089 \\
 & 10 & 0.089 & 0.098 & 0.080 & 0.085 \\
 & 1000 & 0.089 & 0.090 & 0.084 & 0.083 \\
 \hline
Euclidean Depth Estimation & 0.1 & 0.078& 0.091 & 0.070 & 0.074 \\
 & 10 & 0.078 & 0.082 & 0.070 & 0.073 \\
 & 1000 & 0.077 & 0.078 & 0.070 & 0.071 \\
 \hline
Scene Classification & 0.1 & 2.42 & 2.60 & 2.19 & 2.31 \\
 & 10 & 2.01 & 2.21 & 1.60 & 1.81 \\
 & 1000 & 1.73 & 1.96 & 1.35 & 1.43 \\
\end{tblr}
}
\end{table}

\section{Effect of the Number of Clusters $k$} \label{app:vary_k}
To evaluate whether the results of our proposed method for inducing data heterogeneity depend on the choice of $k$, we vary $k \in \{2,4,10,16,32\}$ and report the final converged loss after $20$ global aggregation rounds for each task and data heterogeneity level. The results are summarized in Table~\ref{tab:VaryKReults}.

Across tasks, we observe that for moderate to large values of $k$ (i.e., $k \in \{10,16,32\}$), the qualitative behavior of our method is consistent. In particular, the monotonic relationship between data heterogeneity and performance is preserved, with the highest heterogeneity level ($\alpha=0.1$) yielding the highest loss and larger $\alpha$ values yielding lower loss. For example, in the Surface Normals task, the loss under $\alpha=0.1$ increases from $0.190$ ($k=10$) to $0.192$ ($k=16$) and $0.193$ ($k=32$), while decreasing to $0.186$ and $0.187$ under $\alpha=1000$ across all three $k$ values (i.e., $k \in \{10,16,32\}$). Similar monotonic trends are observed for Reshading, 2D Edges, Euclidean Depth Estimation, and 3D Keypoints. Note that $\alpha=1000$ results in the lowest loss across all $k$ values and is often not sensitive to the choice of $k$ as it corresponds to a near-data homogeneity scenario and uniform data sampling across the clusters.

In contrast, very small values of $k$ ($k=2$ and $k=4$) produce loss values comparable to those observed under the lowest heterogeneity setting across most tasks (e.g., Surface Normals: $0.187$ at $\alpha=0.1$ for $k=2$ versus $0.186$ at $\alpha=1000$), indicating insufficient separation between clusters and therefore inability to induce meaningful data heterogeneity.

While the qualitative conclusions are invariant for $k \in \{10,16,32\}$, we observe that $k=16$ and $k=32$ often yield the largest performance gaps under severe heterogeneity. For instance, in Scene Classification at $\alpha=0.1$, the loss reaches $2.31$ for $k=16$ and $2.37$ for $k=32$, compared to $1.80$ for $k=10$. Given the need to adopt a single $k$ value across all tasks and our cluster validity analysis in Appendix~\ref{App:CVI}, where the Silhouette score for $k=32$ is only marginally higher than that of $k=16$, we adopt $k=16$ as the default setting.

Overall, these results indicate that the substantive conclusions of our method are stable across a meaningful range of cluster counts, and that the choice of $k$, as long as it is chosen to be sufficiently large, does not determine whether embedding-based data heterogeneity induces clear separation between low- and high-heterogeneity scenarios.

\begin{table}[!ht]
% \vspace{-4mm}
\centering
\caption{\textit{Effect of varying the number of clusters $k$ on embedding-based data heterogeneity.} Final converged loss after $20$ global aggregation rounds for different values of $k$ across tasks and data heterogeneity levels. For moderate to large values of $k$ (i.e., $k\in\{10,16,32\}$), the monotonic relationship between data heterogeneity level and performance is preserved and our method continues to induce clear separation between low- and high-heterogeneity scenarios, while very small values of $k$ fail to produce meaningful performance gaps.}
\label{tab:VaryKReults}
{\footnotesize
\begin{tblr}{
colspec ={Q[c,m,wd=24mm]  Q[c,m,wd=7mm] Q[c,m,wd=7mm] Q[c,m,wd=7mm] Q[c,m,wd=7mm] Q[c,m,wd=7mm] Q[c,m,wd=7mm]},
hlines = {black,0.1mm},
vlines = {black,0.1mm},
row{2} = {LightBlue},
cell{1}{1} = {r=2, c=1}{m, DarkBlue},
cell{1}{2} = {r=2, c=1}{m, DarkBlue},
cell{1}{3} = {r=1, c=5}{m, DarkBlue},
cell{3}{1} = {r=3, c=1}{m, white},
cell{6}{1} = {r=3, c=1}{m, white},
cell{9}{1} = {r=3, c=1}{m, white},
cell{12}{1} = {r=3, c=1}{m, white},
cell{15}{1} = {r=3, c=1}{m, white},
cell{18}{1} = {r=3, c=1}{m, white},
cell{21}{1} = {r=3, c=1}{m, white},
colsep=3pt, rowsep=3pt
                }
Vision Task & $\alpha$ & $k$ & & & \\
& & 2 & 4 & 10 & 16 & 32 \\
\hline
2D Edges & 0.1 & 0.072 & 0.072 & 0.073 & \textbf{0.074} & 0.073\\
 & 10 & 0.070 & 0.071 & 0.072 & \textbf{0.073} & 0.072 \\
 & 1000 & 0.070 & 0.070 & 0.070 & \textbf{0.071} & \textbf{0.071} \\
 \hline
Reshading & 0.1 & 0.050 & 0.052 & 0.053 & \textbf{0.055} & 0.054 \\
 & 10 & 0.049 & 0.051 & 0.052 & 0.052 & \textbf{0.053}\\
 & 1000 & 0.049 & \textbf{0.050} & \textbf{0.050} & 0.049 & 0.049\\
 \hline
 Surface Normals & 0.1 & 0.187 & 0.189 & 0.190 & 0.192 & \textbf{0.193} \\
 & 10 & 0.186 & 0.187 & 0.189 & 0.191 & \textbf{0.192} \\
 & 1000 & 0.186 & 0.186 & \textbf{0.187} & 0.186 & 0.186 \\
 \hline
Semantic Segmentation & 0.1 & 0.046 & 0.047 & 0.050 & \textbf{0.052} & \textbf{0.052} \\
 & 10 & 0.047 & 0.047 & 0.047 & \textbf{0.048} & 0.047 \\
 & 1000 & \textbf{0.047} & \textbf{0.047 }& \textbf{0.047} & \textbf{0.047} & 0.046 \\
 \hline
3D Keypoints & 0.1 & 0.082 & 0.084 & 0.087 & \textbf{0.089} & 0.088 \\
 & 10 & 0.083 & 0.083 & 0.085 & \textbf{0.085} & 0.086 \\
 & 1000 & 0.082 & 0.082 & \textbf{0.083} & \textbf{0.083} & \textbf{0.083} \\
 \hline
Euclidean Depth Estimation & 0.1 & 0.071 & 0.071 & 0.072 & 0.074 & \textbf{0.075} \\
 & 10 & 0.070 & 0.071 & 0.071 & 0.073 & \textbf{0.073} \\
 & 1000 & 0.070 & 0.070 & 0.070 & \textbf{0.071} & 0.070 \\
 \hline
Scene Classification & 0.1 & 1.48 & 1.50 & 1.80 & 2.31 & \textbf{2.37} \\
 & 10 & 1.43 & 1.45 & 1.67 & 1.81 & \textbf{1.92} \\
 & 1000 & 1.44 & 1.43 & 1.43 & 1.43 & \textbf{1.45} \\
\end{tblr}
}
\end{table}

\section{Effect of Clustering Family on Embedding-Based Data Heterogeneity}
\label{app:alt_clustering}

The embedding-based client data partitioning procedure in our method relies on clustering data embeddings to construct heterogeneous data distributions. While K-means is used as the default clustering method, it is important to note that the substantive conclusions of the benchmark are not tied to a particular clustering family or to Euclidean K-means geometry.
To show this, we repeat the embedding-based data partitioning procedure using \textit{agglomerative clustering} in place of K-means, while keeping all other components of the pipeline fixed, including the embedding extractor, number of clusters, Dirichlet concentration parameter $\alpha$, training protocol, and evaluation procedure. We compare three scenarios: (i) class-based data heterogeneity, (ii) embedding-based data heterogeneity constructed using K-means, and (iii) embedding-based data heterogeneity constructed using agglomerative clustering. The results for all tasks and data heterogeneity levels are reported in Table~\ref{tab:AClusteringResults}.

\begin{figure*}[t]
    \centering
    \includegraphics[width=\linewidth]{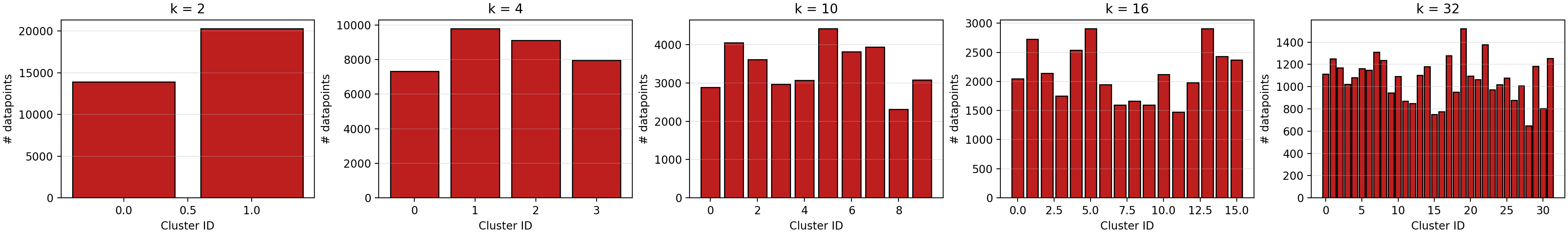}
    \caption{Number of datapoints per embedding cluster for the Euclidean Depth Estimation task with varying $K\in\{2,4,10,16,32\}$.}
    \label{fig:cluster-sizes-by-k}
\end{figure*}

\begin{table}[!h]
\centering
\caption{\textit{Effect of clustering family on embedding-based data heterogeneity.} Final converged loss after $20$ global aggregation rounds for class-based data heterogeneity and embedding-based data heterogeneity constructed using K-means and agglomerative clustering across tasks and data heterogeneity levels. Both clustering methods preserve the monotonic relationship between data heterogeneity level and performance and induce substantially larger performance gaps than class-based splits, indicating that the benchmark conclusions are not an artifact of Euclidean K-means geometry.}
\label{tab:AClusteringResults}
{\footnotesize
\begin{tblr}{
colspec ={Q[c,m,wd=24mm]  Q[c,m,wd=7mm] Q[c,m,wd=14mm] Q[c,m,wd=18mm] Q[c,m,wd=14mm]},
hlines = {black,0.1mm},
vlines = {black,0.1mm},
row{2} = {LightBlue},
cell{1}{1} = {r=2, c=1}{m, DarkBlue},
cell{1}{2} = {r=2, c=1}{m, DarkBlue},
cell{1}{3} = {r=1, c=3}{m, DarkBlue},
cell{3}{1} = {r=3, c=1}{m, white},
cell{6}{1} = {r=3, c=1}{m, white},
cell{9}{1} = {r=3, c=1}{m, white},
cell{12}{1} = {r=3, c=1}{m, white},
cell{15}{1} = {r=3, c=1}{m, white},
cell{18}{1} = {r=3, c=1}{m, white},
cell{21}{1} = {r=3, c=1}{m, white},
colsep=3pt, rowsep=3pt
}
Vision Task & $\alpha$ & Data Heterogeneity Method & & \\
& & Class-based & Agglomerative (Embedding-based) & K-Means (Embedding-based) \\
\hline
2D Edges & 0.1  & 0.072 & 0.083 & 0.074 \\
 & 10  & 0.072 & 0.080 & 0.073 \\
 & 1000  & 0.072 & 0.073 & 0.071 \\
 \hline
Reshading & 0.1  & 0.048 & 0.067 & 0.055 \\
 & 10  & 0.048 & 0.064 & 0.052 \\
 & 1000  & 0.048 & 0.049 & 0.049 \\
 \hline
 Surface Normals & 0.1  & 0.190 &0.204 & 0.192 \\
 & 10  & 0.190 & 0.193 & 0.191 \\
 & 1000 & 0.190 & 0.187 & 0.186 \\
 \hline
Semantic Segmentation & 0.1 & 0.047 & 0.073 & 0.052 \\
 & 10 & 0.047 & 0.070 & 0.048 \\
 & 1000 & 0.047 & 0.048 & 0.047 \\
 \hline
3D Keypoints & 0.1 & 0.082 & 0.095 & 0.089 \\
 & 10 & 0.080 & 0.094 & 0.085 \\
 & 1000 & 0.084 & 0.082 & 0.083 \\
 \hline
Euclidean Depth Estimation & 0.1  & 0.070 & 0.083 & 0.074 \\
 & 10 & 0.070 & 0.080 & 0.073 \\
 & 1000 & 0.070 & 0.072 & 0.071 \\
 \hline
Scene Classification & 0.1 & 2.19 & 2.43 & 2.31 \\
 & 10 & 1.60 & 1.92 & 1.81 \\
 & 1000 & 1.35 & 1.40 & 1.43 \\
\end{tblr}
}
\end{table}

Across tasks, both K-means and agglomerative clustering consistently induce substantially larger performance degradation under high data heterogeneity ($\alpha=0.1$) compared to class-based data heterogeneity. For example, in the Reshading task at $\alpha=0.1$, the loss increases from $0.048$ under class-based heterogeneity to $0.055$ using K-means and to $0.067$ using agglomerative clustering. Similar behavior is observed for Semantic Segmentation ($0.047$ vs.\ $0.052$ and $0.073$) and Scene Classification ($2.19$ vs.\ $2.31$ and $2.43$).

Moreover, for both clustering methods, the loss decreases as $\alpha$ increases under embedding-based data heterogeneity, indicating that the monotonic relationship between data heterogeneity level and performance is preserved regardless of the clustering family. Overall, these results demonstrate that the qualitative conclusions of our method are not an artifact of Euclidean K-means geometry. Instead, embedding-based data heterogeneity induces a clear separation between low- and high-heterogeneity regimes and exposes performance sensitivity not captured by class-based splits independent of the clustering method used.

\section{Qualitative Behavior of Client Data Partitions Under Varying $K$}
\label{app:cluster_sizes}

We study the sensitivity of the proposed embedding-based data heterogeneity formulation to the choice of the number of embedding clusters. While $k=16$ is used as the default value throughout the main experiments, we perform an additional analysis by sweeping $k \in \{2,4,10,16,32\}$ using the Euclidean Depth Estimation task as a representative example (similar qualitative behavior is observed across the remaining tasks, and we therefore report results for a representative task to avoid redundancy).

Fig.~\ref{fig:cluster-sizes-by-k} reports the number of datapoints assigned to each embedding cluster for different values of $k$. As expected, increasing $k$ reduces the average number of datapoints per cluster. However, no cluster collapses or dominates across the tested values of $k$, and cluster sizes remain within comparable orders of magnitude. This indicates that increasing $k$ refines the granularity of the embedding partition without introducing drastic imbalance or artificially extreme partitions.

\begin{figure*}[!ht]
    \centering
    \includegraphics[width=1\linewidth]{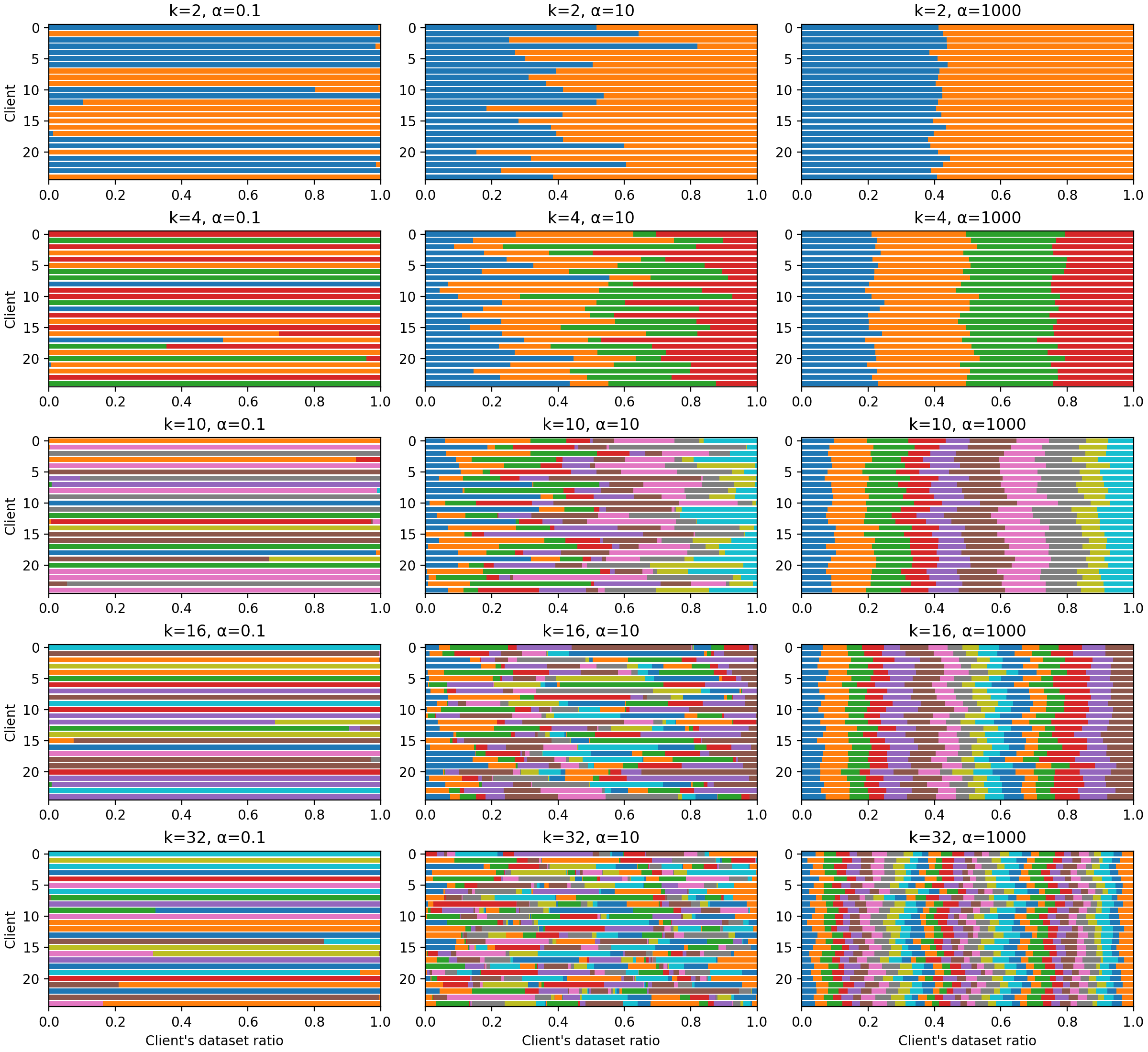}
    \caption{Client-level ratios of data originating from each embedding cluster across varying $K$ and Dirichlet concentration parameters $\alpha$ for the Euclidean Depth Estimation task.}
    \label{fig:client-data-distributions}
\end{figure*}

We next analyze the effective client-level data composition after Dirichlet allocation. Fig.~\ref{fig:client-data-distributions} visualizes, for each $(k,\alpha)$ pair, the ratio of each client’s dataset originating from each embedding cluster.

Across all tested values of $k$, we observe consistent qualitative behavior. Under high data heterogeneity ($\alpha=0.1$), client datasets are dominated by a very small number of clusters, whereas under low data heterogeneity ($\alpha=1000$), client datasets approach near-uniform mixtures over clusters. Also, intermediate heterogeneity ($\alpha=10$) exhibits partially mixed but still skewed compositions. These structural patterns are preserved across all values of $k$.

The stability of both cluster size distributions and effective client sample compositions across $k$ explains why the monotonic relationship between $\alpha$ and performance, as well as the relative ranking of performance under different data heterogeneity levels reported in Sec.~\ref{sec:results}, remain consistent when varying $k$. 
% In other words, changing $k$ primarily controls the resolution of grouping but does not qualitatively alter the induced heterogeneity structure or bake in a particular difficulty level. Similar qualitative behavior is observed across the remaining tasks, and we therefore report results for a representative task to avoid redundancy.

\end{document}